\documentclass{article}

\usepackage{arxiv}

\usepackage[utf8]{inputenc} 
\usepackage[T1]{fontenc}    
\usepackage{hyperref}       
\usepackage{url}            
\usepackage{booktabs}       
\usepackage{caption}        
\usepackage{amsmath}        
\usepackage{amsfonts}       
\usepackage{amssymb}        
\usepackage{nicefrac}       
\usepackage{microtype}      
\usepackage{cleveref}       
\usepackage{lipsum}         
\usepackage{graphicx}
\usepackage{float}          
\usepackage{placeins}       
\usepackage{tikz}
\usetikzlibrary{arrows.meta, positioning, fit}
\tikzset{
  box/.style={draw, rectangle, rounded corners,
    minimum width=2cm, minimum height=0.8cm,
  align=center, font=\normalsize},
  arrow/.style={-{Stealth}, thick},
  llm/.style={box, fill=blue!10},
  cp/.style={box, fill=red!10}
}
\usepackage{natbib}
\usepackage{doi}
\usepackage{tcolorbox}
\tcbuselibrary{breakable, listings}
\newtcolorbox{promptbox}{
  colback=gray!5,
  colframe=black!60,
  boxrule=0.5pt,
  arc=3pt,
  left=6pt,
  right=6pt,
  top=6pt,
  bottom=6pt,
  breakable,
  listing only,
  listing options={
    basicstyle=\ttfamily\small,
    breaklines=true
  }
}
\usepackage{fvextra}
\usepackage{enumitem}
\newcommand{\tx}{\tilde{x}}

\title{Is Conformal Factuality for RAG-based LLMs Robust? Novel Metrics and Systematic Insights}

\date{March 17, 2026}

\author{
  \href{https://orcid.org/0000-0002-7936-1575}{\includegraphics[scale=0.06]{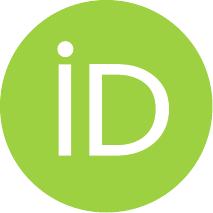} \hspace{1mm} Yi Chen} \\
  University of Wisconsin-Madison \\
  \texttt{yi.chen@wisc.edu} \\
  \And
  Daiwei Chen \\
  University of Wisconsin-Madison \\
  \texttt{daiwei.chen@wisc.edu} \\
  \And
  Sukrut Madhav Chikodikar \\
  University of Wisconsin-Madison \\
  \texttt{chikodikar@wisc.edu} \\
  \AND
  Caitlyn Heqi Yin \\
  University of Wisconsin-Madison \\
  \texttt{hyin66@wisc.edu} \\
  \And
  Ramya Korlakai Vinayak \\
  University of Wisconsin-Madison \\
  \texttt{ramya@ece.wisc.edu} \\
}

\renewcommand{\shorttitle}{Is Conformal Factuality for RAG-based LLMs Robust?}

\hypersetup{
  pdftitle={Is Conformal Factuality for RAG-based LLMs Robust? Novel Metrics and Systematic Insights},
  pdfsubject={cs.AI, stat.ML},
  pdfauthor={Yi Chen, Daiwei Chen, Sukrut Madhav Chikodikar, Caitlyn Heqi Yin, Ramya Korlakai Vinayak},
  pdfkeywords={RAG, LLM, hallucination mitigation, conformal prediction, factuality guarantees},
}

\begin{document}
\maketitle

\begin{abstract}
Large language models (LLMs) frequently hallucinate, limiting their reliability in knowledge-intensive applications. Retrieval-augmented generation (RAG) and conformal factuality have emerged as potential ways to address this limitation. While RAG aims to ground responses in retrieved evidence, it provides no statistical guarantee that the final output is correct. Conformal factuality filtering offers distribution-free statistical reliability by scoring and filtering atomic claims using a threshold calibrated on held-out data, however, the informativeness of the final output is not guaranteed. We \textbf{\emph{systematically analyze the reliability and usefulness of conformal factuality for RAG-based LLMs}} across generation, scoring, calibration, robustness, and efficiency. We propose \emph{\textbf{novel informativeness-aware metrics}} that better reflect task utility under conformal filtering. Across three benchmarks and multiple model families, we find that (i) conformal filtering suffers from \emph{low usefulness at high factuality levels} due to vacuous outputs, (ii) conformal factuality guarantee is \emph {not robust to distribution shifts and distractors}, highlighting the limitation that requires calibration data to closely match deployment conditions, and (iii) lightweight entailment-based verifiers match or outperform LLM-based model confidence scorers while requiring over $100\times$ fewer FLOPs. Overall, \emph{\textbf{our results expose factuality--informativeness trade-offs and fragility of conformal filtering framework under distribution shifts and distractors}}, highlighting the need for new approaches for reliability with robustness and usefulness as key metrics, and provide actionable guidance for building RAG pipelines that are both reliable and computationally efficient.
\end{abstract}

\section{Introduction}
Large language models (LLMs) have demonstrated remarkable capabilities across open-domain question answering, reasoning, and scientific discovery~\citep{brown2020language, guo2025deepseek, zhang2025exploring}. Yet, a persistent barrier to their reliable deployment is the phenomenon of \emph{hallucinations}: outputs that are fluent and confident but factually incorrect~\citep{ji2023survey, nadeau2024benchmarking, huang2025survey}. Such errors are not merely cosmetic. In safety-critical settings, such as medicine, law, or finance, a single fabricated claim can erode trust, propagate misinformation, and incur high societal or financial costs. This makes hallucination mitigation one of the central challenges in advancing trustworthy LLMs. A rich body of work has emerged to address this challenge along two main directions: (1) retrieval-augmented generation (RAG) and (2) conformal methods. RAG aims to reduce hallucinations by grounding responses in trusted external knowledge sources, typically by conditioning generation on retrieved passages~\citep{lewis2020retrieval, gao2023retrieval, siriwardhana2023improving}. While RAG reduces the likelihood of unsupported claims, it does not offer statistical guarantees on the factuality of the final response. Even with reference, LLMs can produce hallucinations in the generated response~\citep{huang2025survey}. The conformal prediction (CP) framework, on the other hand, aims to provide a statistical guarantee on the final output, often via post-processing of the initial LLM response. CP frameworks usually first decompose the initial LLM output into atomic claims, score each claim with a factuality scoring function, and filter those falling below a threshold determined using a calibration dataset~\citep{mohri2024language, cherian2024large}. This procedure provides formal coverage guarantees but often at the expense of informativeness, since aggressive filtering may yield empty or vacuous outputs. Furthermore, CP filtering cannot improve the usefulness or accuracy of the LLM response; it can only remove hallucinations.

Despite their complementary strengths, it is unclear whether conformal prediction (CP) can improve reliability of RAG-based LLMs. While several recent works integrate RAG with CP methods~\citep{li2023traq, rouzrokh2024conflare, feng2025response}, they fall short of a systematic analysis that disentangles where gains come from and when guarantees break down. A comprehensive understanding of the strengths and limitations of this combination requires not only integrating the two frameworks, but also developing principled evaluation. In particular, standard metrics such as empirical factuality can exhibit multiple failure modes and obscure real utility—for example, an empty answer is trivially ``factually correct'' under such measures. 
Additionally, it remains unclear whether improved factuality \emph{necessarily} requires larger, more computationally expensive verifiers, or whether lightweight alternatives can achieve comparable or superior performance. This raises a scaling-law-style question for factuality filtering: how do reliability and utility scale with verifier capacity and inference cost, and where are the diminishing returns? Understanding the relationship between factuality, model scale, and computational cost is critical for deploying reliable LLM systems in practice, where both latency and compute budgets constrain end-to-end RAG pipelines. We address these issues in this work.

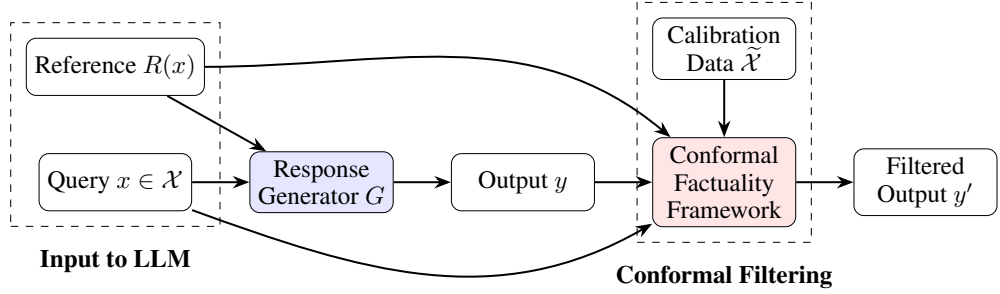
\begin{figure}[t]
  \centering
  \resizebox{0.8\textwidth}{!}{%
    \begin{tikzpicture}[node distance=0.8cm, transform shape]

      \node[box] (input) {Query $x \in \mathcal{X}$};
      \node[box, above=of input] (reference) {Reference $R(x)$};
      \node[llm, right=of input] (generator) {Response\\Generator $G$};
      \node[box, right=of generator] (output) {Output $y$};
      \node[cp, right=of output] (cp) {Conformal\\Factuality\\Framework};
      \node[box, above=of cp] (calib) {Calibration\\Data $\widetilde{\mathcal{X}}$};
      \node[box, right=of cp] (foutput) {Filtered\\Output $y'$};

      \draw[arrow] (input) -- (generator);
      \draw[arrow] (reference) -- (generator);
      \draw[arrow] (generator) -- (output);
      \draw[arrow] (output) -- (cp);
      \draw[arrow] (calib) -- (cp);
      \draw[arrow] (reference) edge[out=0, in=140] (cp);
      \draw[arrow] (input) edge[out=-20, in=-150] (cp);
      \draw[arrow] (cp) -- (foutput);

      \node[draw,dashed,fit=(input)(reference), inner sep=6pt,label={[yshift=-0.2cm]below:\textbf{Input to LLM}}] (pregroup) {};

      \node[draw,dashed,fit=(calib)(cp), inner sep=6pt,label={[yshift=-0.2cm]below:\textbf{Conformal Filtering}}] (pregroup) {};

    \end{tikzpicture}
  }
  \caption{Overview of our framework. Given a query $x$ and retrieved references $R(x)$, the Response Generator $G$ produces an output $y$. The conformal factuality framework utilizes a separate calibration data to determine a threshold used to filter out information from the output $y$ and yield $y'$. (See Figure~\ref{diag:pipeline} for the details of different stages involved in conformal filtering.)}
  \label{diag:pipeline-simplified}
\end{figure}

\textbf{Our Contributions}: We systematically investigate the conformal factuality filtering framework for RAG-based LLMs (Figure~\ref{diag:pipeline-simplified}) making the following contributions:
\begin{itemize} [leftmargin=*, topsep=-2pt]
    \item We propose \emph{novel metrics} to capture informativeness component that is often missed by traditional metrics: \emph{non-empty rate} and \emph{non-vacuous empirical factuality}, which jointly capture both correctness and information retention, and \emph{sufficient correctness} and \emph{conditional sufficient correctness}, which measure whether an output contains enough correct information to infer the final answer to the query. These novel metrics capture the trade-off between factual correctness and informativeness, providing practical insights and tools for future work on hallucination mitigation.
    \item We conduct comprehensive evaluation that spans diverse datasets (questions with free form answers, math and natural question answering), various open-source model families and sizes (with and without reasoning), and scoring functions (entailment-based and LLM-based scorers). Furthermore, we evaluate robustness against distribution shifts and distractors, shedding light on both the capabilities and limitations of this approach. Our evaluation highlights the key issues -- \emph{limited usefulness at high factuality levels} and \emph{non-robustness to distribution shifts and distractors}. This raises the need for new approaches for guaranteeing factuality with usefulness and robustness as important metrics.
    \item  In addition, we analyze the trade-offs between verification accuracy and computational efficiency, demonstrating that lightweight verifiers can outperform larger LLM-based scorers while requiring orders-of-magnitude fewer FLOPs. 
\end{itemize}
 

\section{Problem Setting, Datasets, Models, and Metrics}
In this section, we describe the conformal filtering framework for RAG-based LLMs (Figure~\ref{diag:pipeline-simplified}), introduce the scoring functions, datasets and evaluation metrics used in this work.

Let $x \in \mathcal{X}$ denote an input query. Let $R(x)$ denote the reference material sufficient to answer $x$. We assume the existence of an oracle retriever that can retrieve $R(x)$ such that the true answer $y^\star$ is either in or can be deduced from $R(x)$. This enables us to focus on evaluating the effectiveness of conformal filtering of answers generated by RAG-based LLMs by decoupling the effects of specific methods used for retrieval. A response generator $G$, instantiated as a large language model (LLM), is prompted with both $x$ and $R(x)$ to produce an output $y = G(x, R(x))$. 
The goal of the conformal filtering is to create a final output $y'$ such that each statement in $y'$ is factually correct at a user's expected level, e.g., 85\%, while being useful in answering the query. Conformal filtering method as applied to this system is described below. 

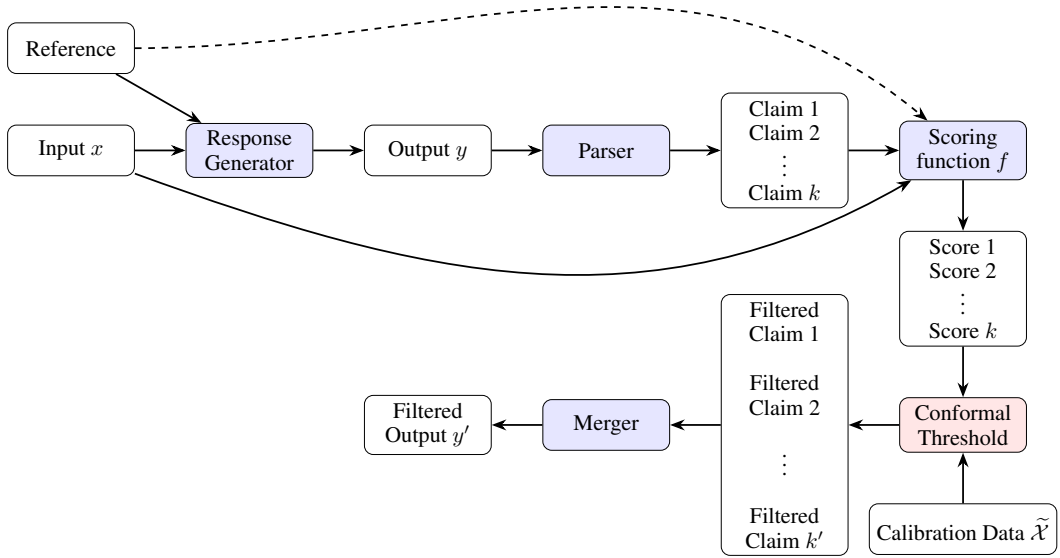
\begin{figure}[ht]
  \centering
  \resizebox{0.85\textwidth}{!}{%
    \begin{tikzpicture}[node distance=0.8cm, transform shape]
      \node[box] (input) {Input $x$};
      \node[box, above=of input] (reference) {Reference};
      \node[llm, right=of input] (generator) {Response\\Generator};
      \node[box, right=of generator] (output) {Output $y$};
      \node[llm, right=of output] (parser) {Parser};
      \node[box, right=of parser] (claims) {Claim 1 \\ Claim 2 \\ $\vdots$ \\ Claim $k$};
      \node[llm, right=of claims] (scorer) {Scoring\\ function $f$};

      \node[box, below=of scorer] (scores) {Score 1 \\ Score 2 \\ $\vdots$ \\ Score $k$};
      \node[cp, below=of scores] (cp) {Conformal\\Threshold};
      \node[box, below=of cp] (calibration) {Calibration Data $\widetilde{\mathcal{X}}$};
      \node[box, left=of cp] (fclaims) {Filtered\\Claim 1 \\ \\ Filtered\\Claim 2 \\ \\ $\vdots$ \\ \\ Filtered\\Claim $k'$};
      \node[llm, left=of fclaims] (merger) {Merger};
      \node[box, left=of merger] (foutput) {Filtered\\Output $y'$};

      \draw[arrow] (input) -- (generator);
      \draw[arrow] (reference) -- (generator);
      \draw[arrow] (generator) -- (output);
      \draw[arrow] (output) -- (parser);
      \draw[arrow] (parser) -- (claims);
      \draw[arrow, dashed] (reference) edge[out=0, in=140] (scorer);
      \draw[arrow] (input) edge[out=-20, in=-150] (scorer);
      \draw[arrow] (claims) -- (scorer);
      \draw[arrow] (scorer) -- (scores);
      \draw[arrow] (scores) -- (cp);
      \draw[arrow] (calibration) -- (cp);
      \draw[arrow] (cp) -- (fclaims);
      \draw[arrow] (fclaims) -- (merger);
      \draw[arrow] (merger) -- (foutput);

    \end{tikzpicture}
  }
  \caption{Given an input $x$ and a reference text related to $x$, the Response Generator produces an output $y$, which is then parsed by the Parser into a list of claims. Each individual claim is subsequently scored by the Scorer, conditioned on the input $x$ and, optionally, the reference text. These scores are passed to the conformal prediction algorithm, which filters out claims whose scores fall below a learned threshold. Finally, the remaining claims are merged into a single paragraph and returned to the user.}
  \label{diag:pipeline}
\end{figure}

\textbf{Conformal Filtering.}  Let $\widetilde{\mathcal{X}}$ denote a calibration set that is exchangeable with $\mathcal{X}$. For each query $\tx \in \widetilde{\mathcal{X}}$, a set of claims $\{ c_i \}$ is obtained by parsing the corresponding $y$ using a parser $P$. These claims are then scored by a factuality scoring function $f$. For each calibration query $\tx$, the candidate threshold is defined as the smallest value of $\tau$ such that all claims with score above $\tau$ are factual: $\inf \{ \tau \mid \forall c \in F(\tau),\; c \text{ is factual} \},$ where $F(\tau) = \{ c \mid f(c) > \tau \}$ denotes the set of claims scoring above $\tau$. Given a specified error level $\alpha \in (0,1)$, the conformal filtering framework determines a threshold $\tau_\alpha$ from the calibration set. Specifically, $\tau_\alpha$ is chosen as the $\tfrac{\lceil(n+1)(1-\alpha)\rceil}{n}$ quantile of a set of candidate thresholds, where $n = |\widetilde{\mathcal{X}}|$ is the size of calibration data.  The resulting guarantee is that $ \mathbb{P}\big( \forall c \in F(\tau_\alpha),\; c \text{ is factual} \big) \;\ge\; 1 - \alpha$. At inference time, for each query $x$, the generated output $y$ is decomposed into atomic claims by a parser $P$, yielding $C(y) = \{ c_i \}_{i=1}^k$.
Each claim $c_i$ is scored with $f$, and those exceeding the threshold are retained:
$ C'(y) = \{\, c_i \;\mid\; f(c_i) > \tau_\alpha,\; i=1,\dots,k \,\}$.
Finally, the retained claims $C'(y)$ are merged by a merger $M$ into a single filtered response: $y' = M(C'(y))$. Figure~\ref{diag:pipeline} illustrates the entire conformal filtering pipeline. 

We use \texttt{gpt-5-nano} \citep{gpt5} in our experiments for tasks such as claim parsing, claim merging and factuality labeling\footnote{We validate the factuality labeling quality of using a \texttt{gpt-5-nano} as a judge by comparing with human labeling (See Appendix \ref{app:human_evaluation} for details of the human evaluation).}.


\subsection{Scoring Functions}\label{sec:scoring-functions}

Scoring function $f$ is a core component of the conformal factuality framework, the determination of which claims are retained or filtered depends on the score. We study two main families of scoring functions:

\textbf{(a) \underline{Entailment-based scorers}} are natural language inference (NLI) models to assess whether the reference text supports a claim. We use both document-level and sentence-level variants. 
\begin{enumerate}[leftmargin=*, topsep=-2pt, noitemsep]
    \item For \textbf{document-level entailment}, the entailment score is computed directly between the entire reference $R(x)$ and the claim $c_i$. We use the entailment model from \citep{laurer2022less} trained on the DocNLI dataset \citep{yin2021docnli}. 
    \item In the \textbf{sentence-level entailment} setting, entailment scores are computed between the claim $c_i$ and each sentence in $R(x)$, and then aggregated in two different ways: 
    \begin{enumerate}[leftmargin=*, topsep=-2pt, noitemsep]
        \item \textbf{Conservative entailment}, where a claim is marked as contradictory if any sentence in the reference contradicts it. It is marked as entailed only if there are no contradictions and there is at least one sentence supporting it. 
        \item \textbf{Average entailment}, which averages the scores of all non-neutral sentence-level comparisons.  For sentence-level entailment, we use \texttt{roberta-large-mnli} as the entailment model \citep{liu2019roberta}.
    \end{enumerate}
\end{enumerate}

\textbf{(b) \underline{LLM-based model confidence scorers.}} In this family, we prompt a language model to assign a factuality score to each claim. We explore the design space of the prompt by varying five dimensions: (i) the inclusion of retrieved references $R(x)$, (ii) evidence highlighting within $R(x)$, (iii) the use of Chain-of-Thought (CoT) reasoning, (iv) output granularity for verbalizing (continuous $[0,1]$ vs. Boolean) , and (v) evaluation consistency (single generation vs. averaging over five independent generations). We refer to this scoring function as \textbf{model confidence score}.

\subsection{Datasets}\label{sec:datasets}
We perform evaluations on three datasets that span open-ended summarization, mathematical reasoning, and question answering tasks. This diversity allows us to assess both the factual reliability and the task-level utility of our approach.
\begin{itemize} [leftmargin=*, topsep=-2pt]

  \item \textbf{FActScore} dataset~\citep{min2023factscore} consists of 601 individuals, each paired with a Wikipedia page. Queries are: ``Tell me a paragraph biography about \texttt{[person]},'' where the reference $R(x)$ is the Wikipedia page of the person. Because no canonical ground-truth answers are provided, this dataset is well-suited for evaluating factuality in \emph{open-ended generation} and for testing whether models can produce faithful summaries grounded in external references. We additionally consider \textbf{FActScore Rare}, a subset of 198 queries focusing on less well-known individuals. This subset probes the robustness of models when the model's parametric knowledge is not enough to answer the question, a regime where hallucinations are more likely, if a reference is not given.

  \item \textbf{MATH} dataset~\citep{hendrycks2021measuring} contains 12,446 competition-style mathematics problems spanning five difficulty levels and seven categories. Each problem provides a question $x$ and a ground-truth answer $y^\star$. To construct reference materials, we prompted \texttt{gpt-5-nano} to generate prerequisite knowledge relevant to solving each problem, which serves as $R(x)$.

  \item \textbf{Natural Questions (NQ)} dataset~\citep{kwiatkowski2019natural} consists of 10K real-world queries collected from search engines. Each query is annotated with both a long answer and a short answer; we use the long answer as the reference $R(x)$ and the short answer as the ground-truth.
\end{itemize}

Together, these datasets provide coverage over distinct capabilities: factual summarization (FActScore), mathematical reasoning (MATH), and reference-based question answering (NQ).

\subsection{Language Models} \label{sec:models}

We evaluate our framework over several open-source language models in order to systematically evaluate different components of the factuality and RAG pipeline under varying architectures, reasoning modes, and parameter scales.

Our open-source suite includes multiple families. The \texttt{Qwen3} models \citep{yang2025qwen3} are evaluated both in their base form and in a reasoning-enabled variant, \texttt{Qwen3-Think}, where reasoning with the \texttt{<think></think>} tag is enabled. This contrast allows us to study whether reasoning-oriented training improves factuality scoring and filtering. To broaden architectural diversity, we also include \texttt{Llama-3.x-Instruct} \citep{dubey2024llama}, \texttt{SmolLM2-Instruct} \citep{allal2025smollm2}, and \texttt{gpt-oss} \citep{agarwal2025gpt}. Table~\ref{tab:models} summarizes the model families, parameter counts, and architectures.

These models are chosen to probe three orthogonal dimensions: (i) \emph{model architecture}, by comparing across families; (ii) \emph{reasoning capability}, by contrasting \texttt{Qwen3} with \texttt{Qwen3-Think}; and (iii) \emph{model scale}. This diversity allows us to assess how each factor influences factuality scoring, and to highlight regimes where smaller, more efficient models suffice.

\begin{table}[t]
  \centering
  \small
  \setlength{\tabcolsep}{6pt}
  \begin{tabular}{llccc}
    \toprule
    \textbf{Family} & \textbf{Model} & \textbf{Params} & \textbf{Arch.} & \textbf{Activated} \\
    \midrule
    Qwen3 & \texttt{Qwen3-0.6B} & 0.6B & Dense & 0.6B \\
    Qwen3 & \texttt{Qwen3-4B} & 4B & Dense & 4B \\
    Qwen3 & \texttt{Qwen3-8B} & 8.2B & Dense & 8.2B \\
    Qwen3 & \texttt{Qwen3-30B-A3B} & 30.5B & MoE & 3.3B \\
    Qwen3 & \texttt{Qwen3-32B} & 32.8B & Dense & 32.8B \\
    \midrule
    Llama-3.x & \texttt{Llama-3.2-1B-Instruct} & 1.2B & Dense & 1.2B \\
    Llama-3.x & \texttt{Llama-3.2-3B-Instruct} & 3.2B & Dense & 3.2B \\
    Llama-3.x & \texttt{Llama-3.1-8B-Instruct} & 8B & Dense & 8B \\
    \midrule
    SmolLM2 & \texttt{SmolLM2-135M-Instruct} & 135M & Dense & 135M \\
    SmolLM2 & \texttt{SmolLM2-360M-Instruct} & 360M & Dense & 360M \\
    SmolLM2 & \texttt{SmolLM2-1.7B-Instruct} & 1.7B & Dense & 1.7B \\
    \midrule
    gpt-oss & \texttt{gpt-oss-20b} & 21B & MoE & 3.6B \\
    gpt-oss & \texttt{gpt-oss-120b} & 117B & MoE & 5.1B \\
    \bottomrule
  \end{tabular}
  \captionsetup{skip=6pt}
  \caption{Summary of evaluated open-source language models. For dense models, the number of activated parameters equals the total parameter count. For MoE models, activated parameter counts are shown when known.}
  \label{tab:models}
\end{table}

\subsection{Evaluation Metrics}

We evaluate the performance using both traditional factuality measures and proposed novel metrics that are designed to capture aspects of informativeness that existing metrics overlook. We use the following commonly used criteria:
\begin{itemize} [leftmargin=*, topsep=-2pt, noitemsep]

  \item \textbf{Empirical Factuality (EF)} measures the fraction of outputs $y'$ in which all retained claims $C'(y)$ are factual. By convention, an empty claim set $C'(y)=\varnothing$ is treated as factual, which can artificially inflate EF when filtering is aggressive. \emph{Higher EF is better}, as it indicates stronger factual reliability.

  \item \textbf{Power} quantifies the average proportion of true claims retained. \emph{Higher Power is better}, since it means fewer correct claims are lost.

  \item \textbf{False Positive Rate (FPR)} measures the fraction of non-factual claims that survive. \emph{Lower FPR is better}, as it reflects stronger suppression of hallucinations.

  \item \textbf{Correctness} measures the fraction of outputs equivalent to the ground-truth answer $y^\star$.
\emph{Higher Correctness is better}, though this metric is intentionally strict and is most applicable on datasets with unambiguous ground-truth answers.
\end{itemize}

These metrics only provide limited insight into the overall usefulness of the final answer. Each of the statements in an LLM response could be factually correct while still not being informative enough to answer the input query. Furthermore, a vacuous or empty output is factually correct by definition, but is not useful. When the input reference does contain information to provide a correct answer to the query, an empty final answer is an indication of failure. While EF, Power, and FPR capture factuality and error rates, they fail to penalize vacuous but “factual” outputs. To address this, \textbf{\emph{we propose the following novel evaluation metrics}}.

\begin{itemize} [leftmargin=*, topsep=-2pt]

  \item \textbf{Non-empty Rate (NR)}, the fraction of outputs that preserve at least one claim. \emph{Higher NR is better}, rewarding informative responses rather than empty ones.

  \item \textbf{Non-vacuous Empirical Factuality (NvEF)}, which computes EF only over non-empty outputs. \emph{Higher NvEF is better}, reflecting factuality conditional on informativeness.

  \item \textbf{Sufficient Correctness (SC)} evaluates whether an output to a given query $x$ (initial output $y$ or filtered output $y'$) contains enough correct information---relative to a reference $R(x)$---to recover the correct answer.
\emph{Higher SC indicates better end-task utility.}
Unlike \emph{Correctness}, which can be overly strict (e.g., penalizing partially correct but still useful responses, or being inapplicable when there is no single canonical $y^\star$, e.g., open-ended summarization task in FActScore), SC explicitly measures whether the content in the output is sufficient to answer the query. 

  \item \textbf{Conditional Sufficient Correctness (CSC)} restricts evaluation of the filtered outputs $y'$ whose unfiltered counterparts $y$ already satisfy SC.
\emph{A higher CSC reflects stronger fidelity of the filtering process.}
CSC isolates the effect of filtering from generation quality. SC on the filtered output can drop either because the base model failed to include sufficient information in the initial response \emph{or} because filtering removed it. Since CSC is conditioned on cases where the unfiltered output has sufficient information and asks whether the filtered output preserves that sufficiency, it provides a direct measurement of whether filtering maintains useful content rather than unnecessarily deleting it.

\end{itemize}

NR, NvEF are claim-level, while correctness and (conditional) sufficient correctness are at the final task-level outcome. \textbf{\emph{Together, these metrics balance factual reliability, informativeness, and task-level utility, ensuring that evaluation reflects not only safety (removing hallucinations) but also usefulness (retaining an adequate signal for the end task).}}
We now design a series of experiments to systematically analyze: (i) the impact of references, (ii) the design of scoring functions, (iii) robustness to distributional shifts and adversarial inputs and (iv) end-to-end evaluation with a focus on overall computation involved.

\section{Impact of References} \label{sec:impact_of_reference}
We begin by isolating the role of references, asking: \emph{How much do retrieved references improve generation quality before filtering is even applied?} To study this, we evaluate outputs $y$ produced by the response generator $G$ under two conditions: query-only generation $y = G(x)$ and query-plus-reference generation $y = G(x, R(x))$. We then measure sufficient correctness of the initial LLM response $y$ with respect to the reference $R(x)$ using \texttt{gpt-5-nano} \footnote{more details in Appendix \ref{app:impact_of_reference}} for different datasets: FActScore, FActScore-Rare, \textbf{MATH-200} (a 200-example subset of MATH), and \textbf{NQ-200} (a 200-example subset of Natural Questions). These datasets help provide insights into whether conditioning on $R(x)$ improves different types of tasks: factual summarization, reasoning, and question answering.

\begin{figure}[h]
  \centering
  \includegraphics[width=\linewidth]{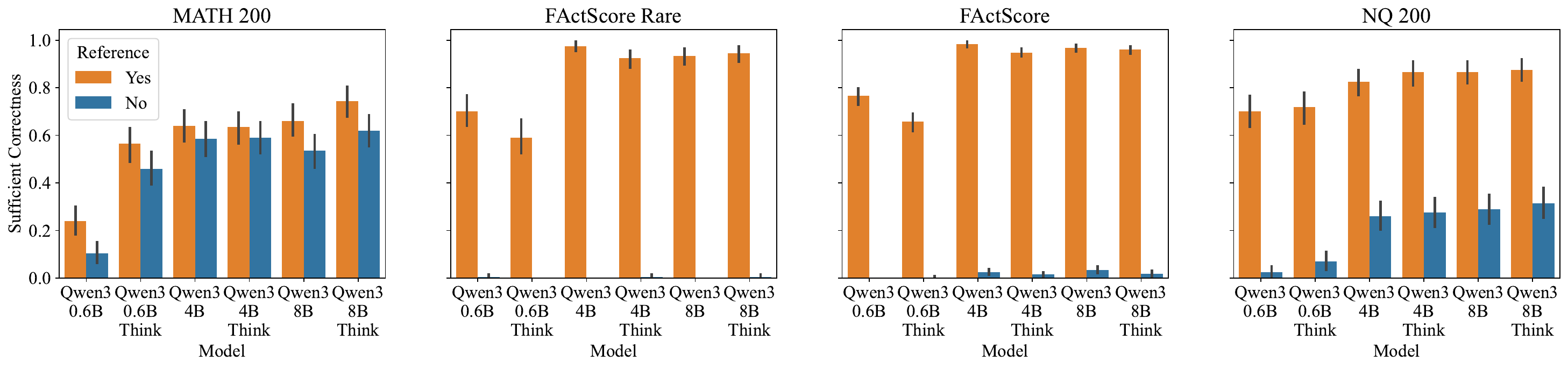}
  \caption{Sufficient correctness (SC) of \texttt{Qwen3} models (0.6B, 4B, 8B) on four datasets (MATH-200, FActScore-Rare, FActScore, NQ-200), with and without access to references. Across model sizes and datasets, providing references consistently improves generation quality.}
  \label{fig:impact_of_references_qwen3}
\end{figure}

\textbf{Results.} Figure~\ref{fig:impact_of_references_qwen3} compares sufficient correctness (SC) with and without references for the \texttt{Qwen3} family. On FActScore and NQ, references play a major role, highlighting their importance when the model may not have memorized all relevant information. On MATH, while the gap between the two settings decreases as model size increases, the SC improves with the reference, suggesting that while larger models possess stronger reasoning abilities the reference material still plays a role. For \texttt{Qwen3-0.6B}, enabling \emph{think} reasoning produces a large jump in SC for MATH dataset, highlighting the role of reasoning capacity for mathematical questions. In contrast, for FActScore and NQ, where reasoning plays a smaller role, enabling \emph{think} has little effect. Lastly, we observe only marginal gains as model size increases within the \texttt{Qwen3} family: the improvement in SC from 4B to 8B is much smaller than the improvement from 0.6B to 4B. Overall, providing references consistently improves generation quality. We observe similar trends for \texttt{Llama-3.x}, \texttt{SmolLM2}, and some frontier models (see Figures~\ref{fig:impact_of_references_llama3}, \ref{fig:impact_of_references_smollm2}, and \ref{fig:impact_of_references_frontier} in Appendix~\ref{app:impact_of_reference}). On FActScore-Rare, the performance of \texttt{Qwen3} 4B is comparable to that of Gemini 2.5 Pro and GPT-5.1 when reference is provided (Figure~\ref{fig:impact_of_references_frontier}). This demonstrates that having good reference enables even small and medium sized LLMs to generate good outputs. The references are provided to the generator $G$ in all subsequent experiments which focus on effectiveness of conformal filtering.

\section{Design Choices for Factuality Scoring in Conformal Filtering}

We now turn to the systematic evaluation of conformal filtering which provides statistical guarantees on factuality of the final output. 
A key component of conformal filtering (Figure~\ref{diag:pipeline}) is the factuality scoring function, which assigns a score indicating how well a generated output is supported by the reference. The effectiveness of conformal filtering therefore depends critically on these scores. In this section, we systematically study several design choices for factuality scoring functions, including prompting strategies for LLM-based model confidence score, the role of references during scoring, the choice of scorer model, and different families of scoring functions (entailment based scores compared with LLM-based scores). Through experiments across multiple datasets and model families, we identify practical configurations that improve filtering performance and calibration.

\subsection{Prompting Strategies for LLM Model Confidence Scores} \label{sec:prompting_strategy}

We begin by examining the effect of prompting variations for LLM-based scoring functions (model confidence score), including: (i) highlighting supporting evidence, (ii) enabling chain-of-thought reasoning, (iii) scalar vs.~Boolean scoring for the verbalized scores, and (iv) consistency averaging (majority of multiple responses). We run the experiments with different configurations of the above mentioned variations on FActScore, MATH-1K (a 1,050-example subset of MATH), and NQ-1K (a 1,000-example subset of Natural Questions). References are always provided to both the scoring function $f$ and the LLM generator $G$ that produces the initial response $y$ (Figure~\ref{diag:pipeline}) in these experiments\footnote{ Note that the sufficient correctness (SC) in this section is calculated on the filtered output $y'$ by prompting \texttt{gpt-5-nano} with the prompts in Appendix \ref{sec:prompt_sc}.} . 

\begin{figure}[h]
  \centering
  \includegraphics[width=\linewidth]{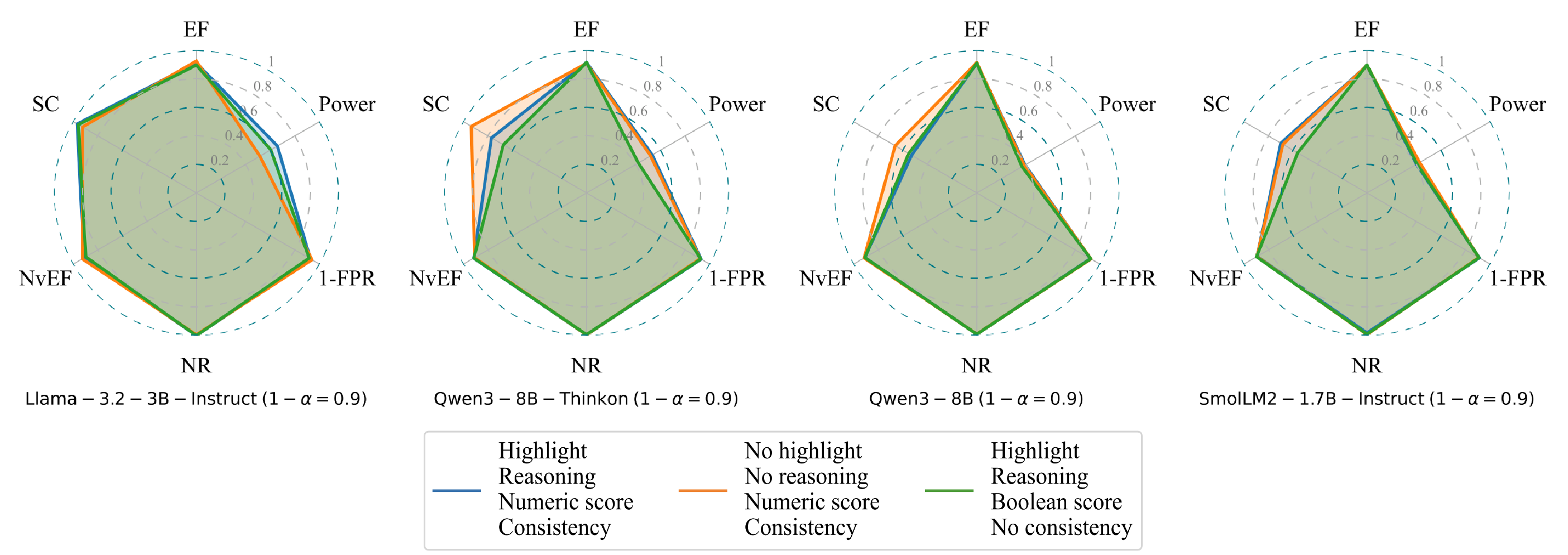}
  \caption{Evaluation of various prompting strategies across different LLMs on the FActScore dataset (Section~\ref{sec:prompting_strategy}) at level $1-\alpha = 0.9$. Results demonstrate that: (i) prompting models to generate numeric scores consistently outperforms Boolean scoring; (ii) sampling multiple responses uniformly improves performance; however, (iii) incorporating chain-of-thought reasoning or evidence highlighting do not yield reliable performance gains across models.} 
  \label{fig:prompting_strategy}
\end{figure}

\textbf{Results.} Figure~\ref{fig:prompting_strategy} summarizes the effect of prompting strategies on model confidence scores across three LLM families for a subset of variations on FActScore dataset at level $1- \alpha = 0.9$. 
While no single strategy emerges as universally optimal, several consistent patterns are observed. 
First, instructing models to output numeric scores reliably outperforms Boolean values. 
Second, sampling multiple responses provides consistent gains, indicating that aggregation reduces variance and stabilizes confidence estimates. 

\subsection{Role of References on Model Confidence Scores} \label{sec:role_of_reference_scoring}

In this section, we evaluate how providing references to the LLM-based model confidence scoring function improves scoring. Outputs $y = G(x, R(x))$ are generated using \texttt{gpt-5-nano}, while open-source models serve as scorers. This is because API inference is faster than running models on our own GPUs. When varying reference access, we hold all other prompt settings fixed, including chain-of-thought prompting, scalar scoring, and consistency averaging. Because reference access changes across conditions, we accordingly adjust whether the model is instructed to highlight supporting evidence from the reference. Experiments are conducted on FActScore, MATH-1K, and NQ-1K.

\begin{figure}[h]
  \centering
  \includegraphics[width=\linewidth]
  {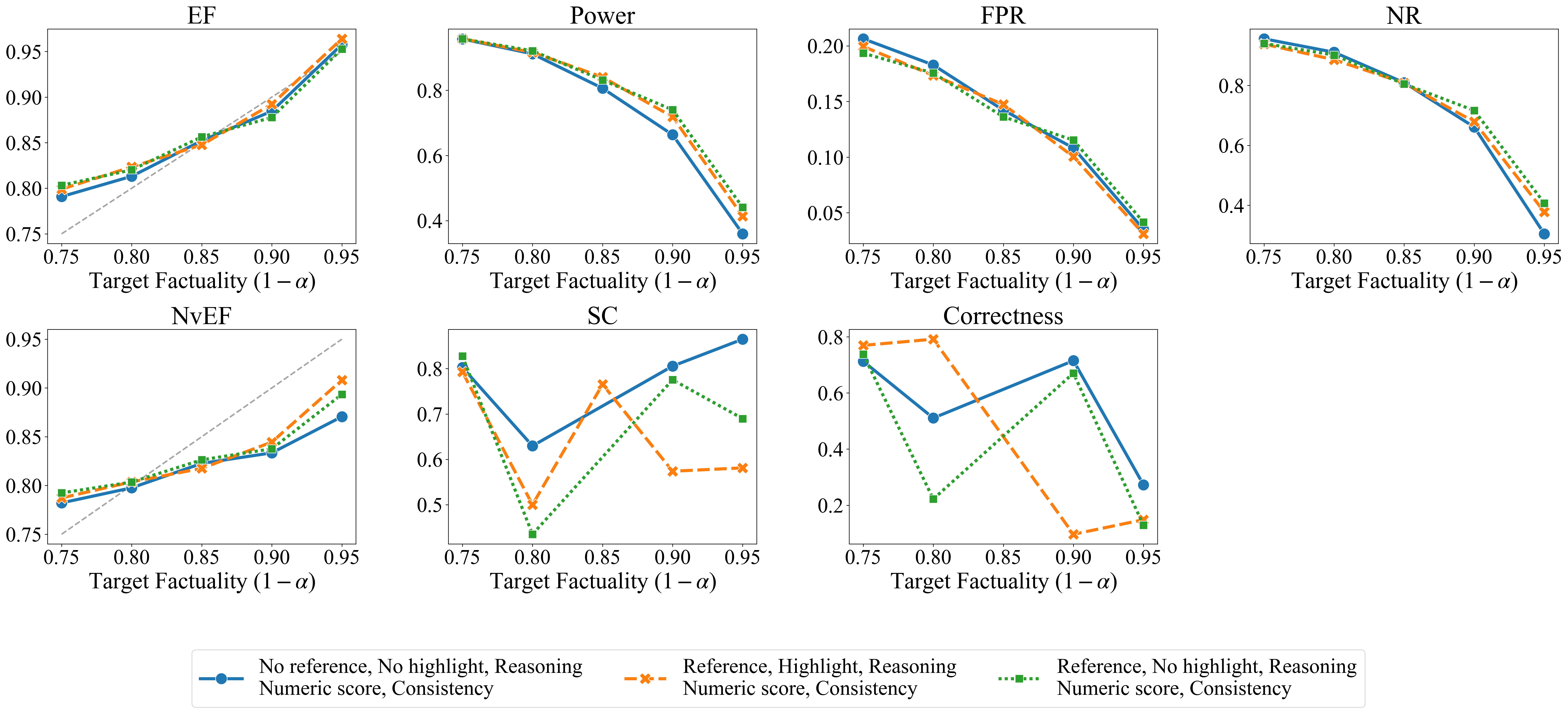}
  \caption{Performance of model confidence score on MATH-1K dataset with and without reference provided to scoring functions using \texttt{Qwen3-4B} as the LLM-based scorer.}
  \label{fig:role_of_reference_scoring_qwen3-4b_math}
\end{figure}

\textbf{Results.} Figure~\ref{fig:role_of_reference_scoring_qwen3-4b_math} shows the performance under various metrics for the model confidence scores with and without reference provided to the model confidence scoring function for the MATH-1K dataset using \texttt{Qwen3-4B} as the scorer. We observe that when a reference is introduced to the model confidence scoring function, the power is consistently higher than in the case where no reference is given to the scoring function. Similarly, for the non-empty rate (NR), the scoring functions with reference have an advantage when the target factuality is of a larger size. These results show the benefit of feeding the reference to the LLM-based model confidence scoring function. Therefore, in the subsequent experiments, we provide a reference to model confidence scoring function. We defer the results for FActScore and NQ-1K dataset to Appendix \ref{app:role_of_reference_scoring}.

\subsection{Model Choice for LLM-based Scorers} \label{sec:model_choice_scorers}
In this section, we study the usage of different open-source models in model confidence scoring functions $f$, using \texttt{gpt-5-nano} as the generator $G$ for initial response. All prompts include references, require evidence highlighting and chain-of-thought reasoning, produce scalar scores, and apply consistency averaging. This experiment assesses how the scorer model family and scale affect factuality filtering.

\begin{figure}[h]
  \centering
  \includegraphics[width=0.8\linewidth]{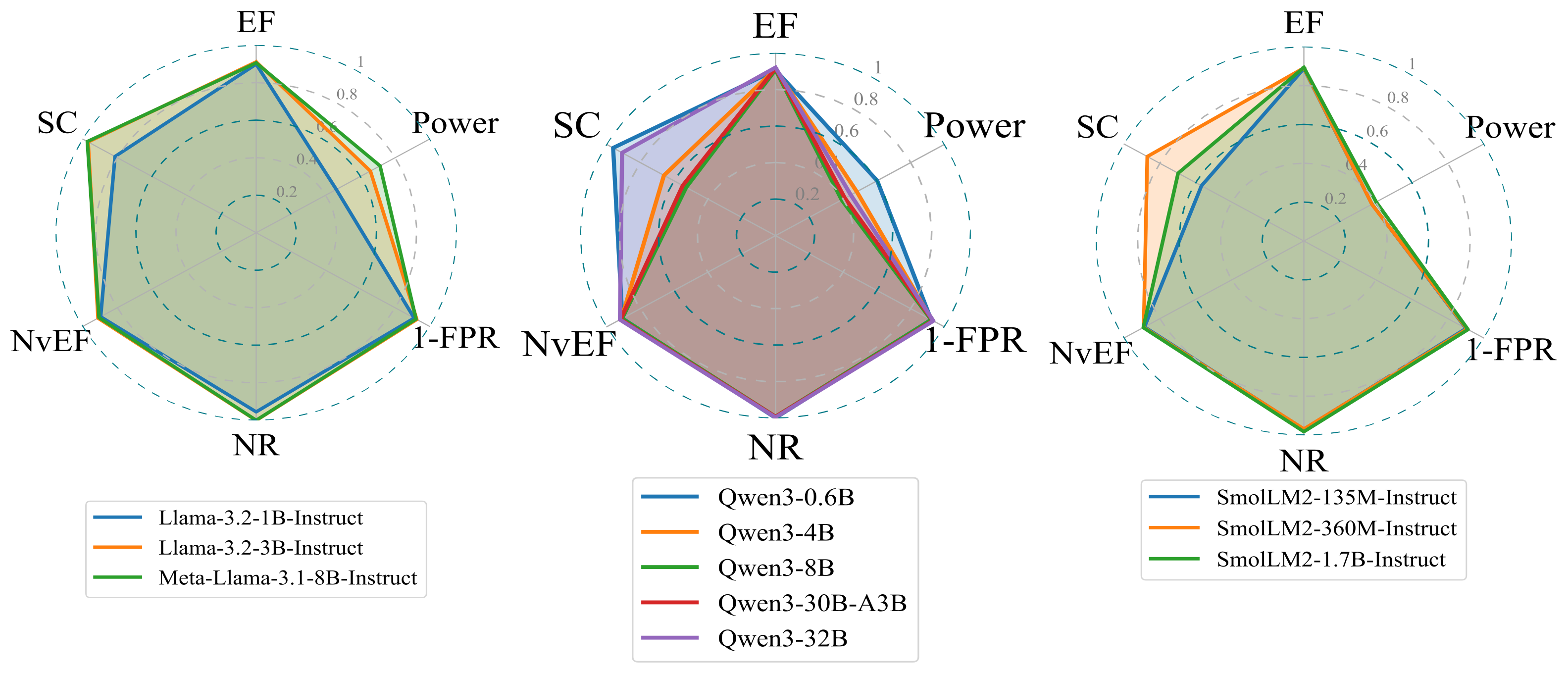}
  \caption{Experimental evaluation across three model families of various scales on the FActScore dataset at level $1-\alpha = 0.9$. Results demonstrate that increasing model size does not consistently improve performance of conformal factuality.} 
  \label{fig:model_choice_scorers}
\end{figure}

\textbf{Results.} Our experiments reveal that scaling LLMs used in model confidence scorer does not guarantee improved confidence calibration in conformal factuality, as shown in Figure~\ref{fig:model_choice_scorers}. While among the \texttt{Llama-3.x} models, there is an improvement in terms of power and sufficient correctness with larger models, this trend breaks with other model families. The \texttt{SmolLM2} models show no systematic benefit as we increase model size; and in the case of \texttt{Qwen3}, scaling even degrades performance. These experiments highlight that smaller models are competitive for scoring in model confidence scores. This observation is practically useful especially since the model used has to be repeatedly called for scoring each claim (for each query). 
We defer the results of different factuality level $\alpha$ to Appendix \ref{app:model_choice_scorers}.

\subsection{Comparison between Entailment-based and LLM-based Scoring Functions} \label{sec:family_choice_scorers}

\begin{figure}[h]
  \centering
  \includegraphics[width=\linewidth]{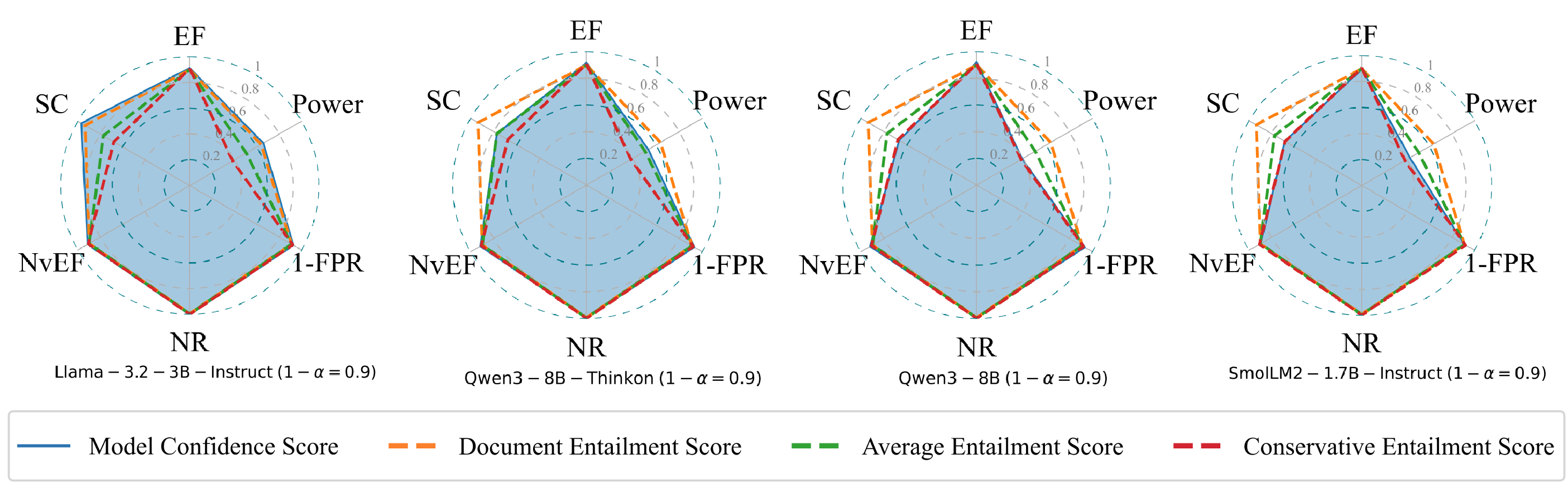}
  \caption{Comparison between entailment-based scores against the model confidence score on the FActScore dataset. Across our evaluated settings, the entailment-based scores match or exceed the model-confidence baseline, despite the entailment model being substantially smaller than the target LLMs.}
  \label{fig:family_choice_scorers}
\end{figure}

Beyond model confidence scores, where we utilize an LLM to generate a score, we also examine entailment-based scoring functions. We consider three different entailment-based scoring functions (Section~\ref{sec:scoring-functions})): document entailment score, average entailment score and conservative entailment score. We compare conformal filtering with entailment-based filtering on FActScore and MATH-1K datasets to assess whether entailment signals offer additional advantages in factuality filtering. 

\textbf{Results.} Figure~\ref{fig:family_choice_scorers} presents a direct comparison between model confidence scores and entailment-based scoring functions. Notably, entailment-based scores, in particular, \textbf{\emph{document entailment score consistently match or exceeds the performance of model confidence score}}, despite the entailment model being substantially smaller than the target LLMs\footnote{We employ \texttt{DeBERTa} and \texttt{RoBERTa} to compute entailment scores, achieving computational efficiency gains of more than two orders of magnitude compared to LLM-based confidence scoring methods. For detailed comparisons of model parameters and computational complexity, see Table~\ref{tab:flops}.}. Together with our analysis in Section~\ref{sec:model_choice_scorers}, provide practical insight that targeted, lightweight scorers (or verifiers) can deliver both computational efficiency and superior performance. 
We also note that, \emph{\textbf{at higher factuality levels, the power is quite limited across all scoring functions.}} We defer the results on the MATH-1K dataset to Appendix \ref{app:model_and_entialment}.

\begin{figure}[h]
  \centering
  \includegraphics[width=\linewidth]{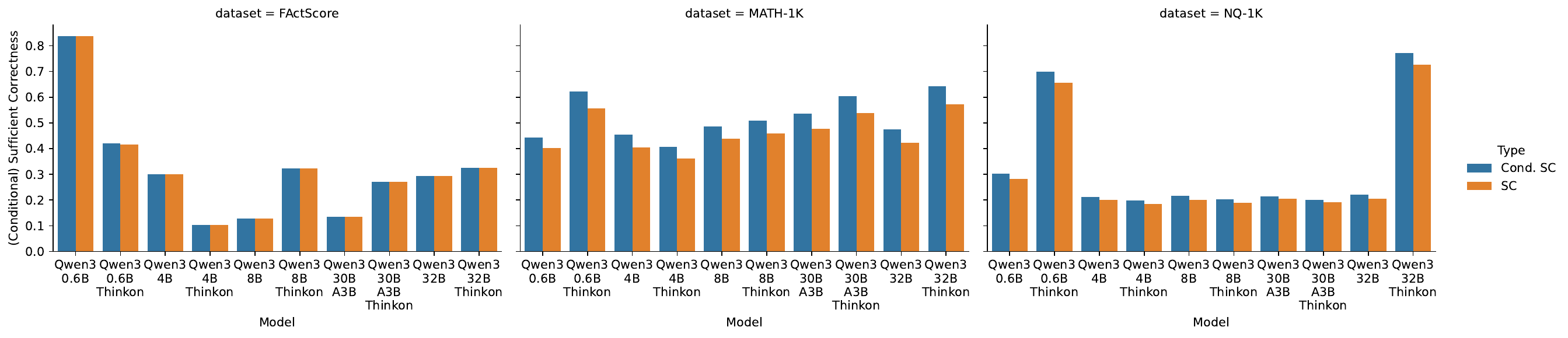}
  \caption{Sufficient Correctness (SC) and Conditional Sufficient Correctness (CSC) for the \texttt{Qwen3} model family at $\alpha=0.05$ across FActScore, MATH-1K, and NQ-1K dataset.}
  \label{fig:qwen3_csc}
\end{figure}

\subsection{Conditional Sufficient Correctness of the Filtered Output} \label{sec:sc_and_csc}

\emph{Sufficient Correctness} (SC) measures whether the output contains sufficient correct information, relative to a reference $R(x)$, to recover the correct answer, with higher values indicating stronger end-task utility. When we measure SC on the final filtered output, it can potentially conflate roles of conformal filtering and the quality of the LLM generator $G$ providing the initial unfiltered output $y$ (see Figure~\ref{diag:pipeline-simplified}). Note that conformal filtering can only remove claims in the initial output but not add to them. So, when the filtered output $y'$ fails to satisfy SC, it does not distinguish whether this failure is caused by the conformal filtering process or by the fact that the original output $y$ already lacked sufficient information. To decouple these effects, we propose \emph{Conditional Sufficient Correctness} (CSC), which evaluates sufficient correctness of the final output $y'$ conditioned on the initial output $y$ being sufficiently correct. This is done by evaluating sufficient correctness only on instances where the unfiltered output already satisfies SC, thereby isolating how well filtering preserves valid answers. For more detailed description on how SC and CSC is measured in Appendix \ref{app:sc_and_csc}.

\textbf{Results.} Figure~\ref{fig:qwen3_csc} reports SC and CSC for the \texttt{Qwen3} family (used in the model confidence scorer) at $\alpha=0.05$ across FActScore, MATH-1K, and NQ-1K datasets. CSC consistently matches or exceeds SC across model scales and reasoning variants. The gap between CSC and SC widens on MATH-1K and NQ-1K. Moreover, we observe that larger models do not automatically yield better SC/CSC within the \texttt{Qwen3} family: the smaller \texttt{Qwen3-0.6B} performs as well as its 32B counterpart on all three datasets. This aligns with our findings in Section~\ref{sec:model_choice_scorers}. A similar scaling trend is observed for the \texttt{gpt-oss} family; results are deferred to Appendix~\ref{app:sc_and_csc}.

\section{Robustness of Factuality Scoring Functions}

Having examined the design and performance of different factuality scoring functions in Section~\ref{sec:family_choice_scorers}, we now study their robustness under distribution shifts. While conformal filtering provides the guarantee of factuality at the required level $1-\alpha$, the practical usefulness depends on whether it remains reliable when key assumptions are violated. In particular, conformal filtering guarantee relies on the exchangeability between the claims in the calibration set and the test data. 
In real-world deployments, the exchangeability assumption may fail due to distribution shifts (which can occur due to various reasons, e.g., the phrasing of the input query, change in the LLM used to generate the initial answer or parser) or adversarial perturbations. To evaluate the robustness of conformal factuality under such conditions, we perform a series of stress tests that examine how scoring functions behave under calibration distribution shifts and distractor injection.

\subsection{Robustness to Calibration Distribution Shift} \label{sec:distribution_shift}

We first study robustness to violations of the exchangeability assumption due to distribution shift in the test data when compared to distribution of claims in the calibration data. Conformal factuality assumes that calibration and test data are exchangeable. We evaluate robustness when this assumption is violated by using mismatched calibration data. To achieve this, we use open-source models as the generator to produce $y$ and parse them into claims. Then, we compare the empirical factuality under two different calibration datasets:
\begin{itemize}
    \item 50 human-annotated claims from~\citet{mohri2024language} (\texttt{gpt-4}-generated claims, denoted as MH). Since these claims are generated by \texttt{gpt-4}, they come from a \emph{different distribution} than claims generated by the open-source model.
    \item 50 randomly selected queries from the held-out test half for which the claims associated are generated by the open-source model, and thus the claims follow the \emph{same distribution} as the test data.
\end{itemize}

We evaluate empirical factuality using \texttt{Qwen3-4B}, \texttt{SmolLM2-360M-Instruct}, and \texttt{Llama-3.2-3B-Instruct} across FActScore, MATH-1K, and NQ-1K datasets.

\begin{figure}[h]
  \centering
  \includegraphics[width=\linewidth]{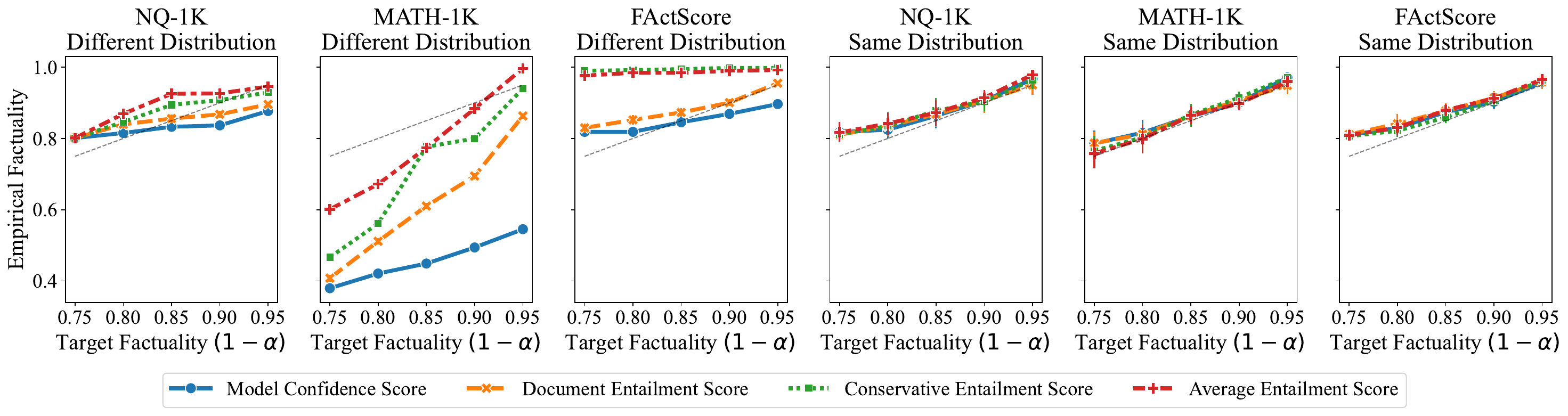}
  \caption{Empirical factuality (EF) on FActScore, MATH-1K, and NQ-1K, under two calibration settings: (i) calibration claims generated in \cite{mohri2024language}, which comes from a different distribution (ii) calibration claims drawn from the same distribution as the test data. We use \texttt{Qwen3-4B} as the scoring function. The results show how the distribution shift in the calibration set affects conformal factuality guarantees.}
  \label{fig:distribution_shift_qwen3}
\end{figure}

\begin{figure}[h]
  \centering
  \includegraphics[width=\linewidth]{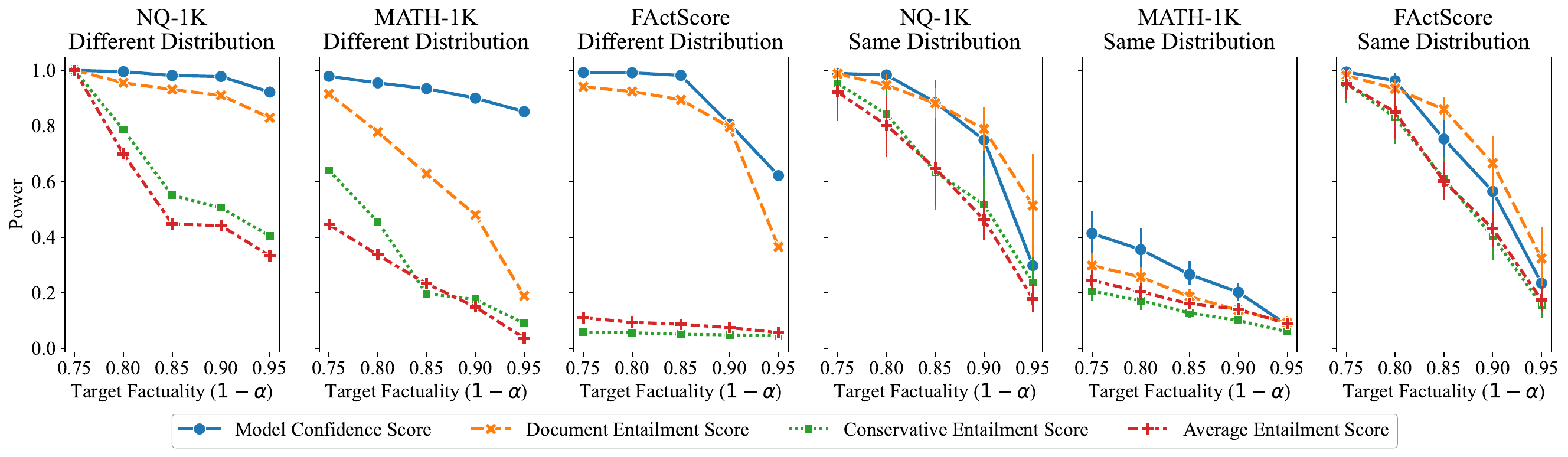}
  \caption{Power on FActScore, MATH-1K, and NQ-1K, under two calibration settings: (i) calibration claims generated in \cite{mohri2024language}, which comes from a different distribution (ii) calibration claims drawn from the same distribution as the test data. We use \texttt{Qwen3-4B} as the scoring function.}
  \label{fig:distribution_shift_qwen3_power}
\end{figure}

\textbf{Results.} Figure~\ref{fig:distribution_shift_qwen3} compares empirical factuality (EF) under different sources of calibration data, with test outputs generated and scored by \texttt{Qwen3-4B}. Figure~\ref{fig:distribution_shift_qwen3_power} compares the power. When calibration data come from a different distribution (generated by \texttt{gpt-4}), EF can fall below the target level (gray dashed line), especially at higher factuality levels. The drop is significant for MATH-1K. While some entailment-based scorers seem to be robust to this distribution shift, it comes at a steep cost on the power as shown in Figure~\ref{fig:distribution_shift_qwen3_power}. Furthermore, this robustness is not consistent across language models: switching $G$ and $f$ to \texttt{Llama-3.2-3B-Instruct} or \texttt{SmolLM2-360M-Instruct} yields different behaviors for the scoring functions that seem robust in the \texttt{Qwen3-4B} setting (see Figures~\ref{fig:distribution_shift_llama3} and~\ref{fig:distribution_shift_smollm2} in Appendix~\ref{app:distribution_shift}). This shows that the conformal filtering is not robust to distribution shifts between the claims in the calibration set and the test set. 

\subsection{Robustness to Distractors} \label{sec:distractors}

Distribution shifts can also occur when LLM outputs contain irrelevant, misleading or hallucinated content. LLM outputs may include claims that appear plausible yet are factually incorrect. This can arise because LLMs are susceptible to being distracted by irrelevant information in the input \citep{shi2023large}. To simulate such conditions, we replace a proportion of factual claims in each test query with distractor statements generated by \texttt{gpt-5-nano}.

Specifically, for each query $x$, its associated reference text $R(x)$, and the set of claims $\{c_i\}$, we prompt the LLM to modify each $c_i$, conditioning on $x$ and $R(x)$, so that the result reflects a type of hallucination the model could plausibly produce. The prompt design used to generate these hallucinated claims is documented in Appendix \ref{generating-hallucination-claims}. Our goal is to create hallucinated claims that are sufficiently convincing to the model, such that it would judge them as potentially originating from itself. To ensure this, after we generate a hallucinated claim, we ask the LLM to check if it thinks that the claim might be generated or hallucinated by itself (prompt in Appendix \ref{hallucination-verification}), given the same $x, R(x)$. If the hallucination claim can cause the model to think that it is the one who generates it, given $x$ and $R(x)$, then we keep this hallucination claim. Otherwise, we repeat this process and generate a new hallucination claim.

We evaluate whether the different scoring functions used in the conformal factuality framework can reliably distinguish correct claims from these plausible but incorrect claims. Experiments are performed on the FActScore, MATH-1K, and NQ-1K datasets.

\begin{figure}[h]
  \centering
  \includegraphics[width=\linewidth]{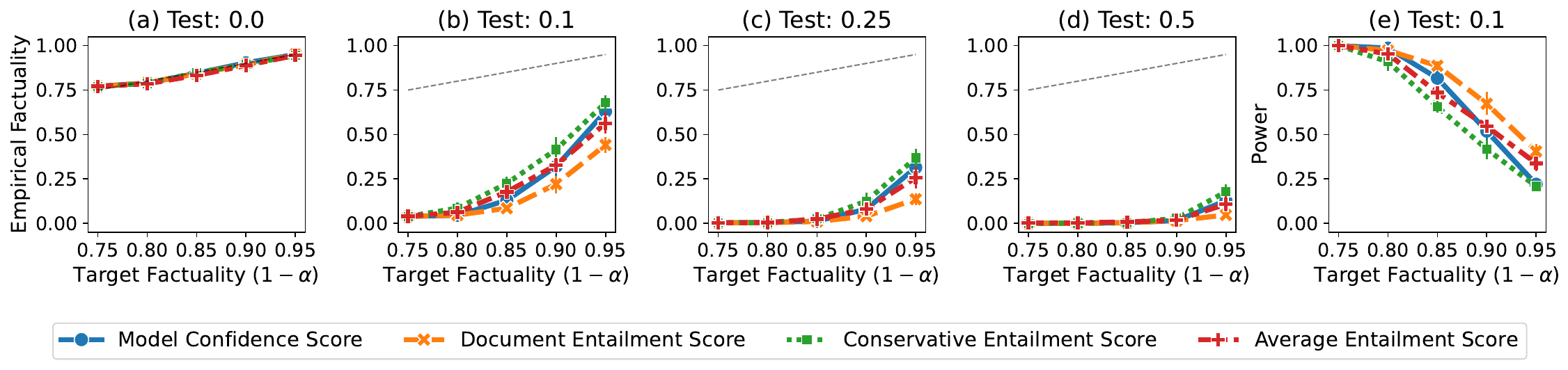}
  \caption{(a) - (d) Empirical factuality versus target factuality (1- $\alpha$) for \texttt{Qwen3-4B} on the FActScore dataset. Each panel corresponds to a different distractor injection rate in the test set (0.0, 0.1, 0.25, 0.5). The gray dashed diagonal represents perfect calibration ($y=x$), where empirical factuality matches the target. (e) Power versus target factuality (1 - $\alpha$) when the injection rate is 0.1.}
  \label{fig:distractors_qwen3_factscore}
\end{figure}

\textbf{Results.} Figure~\ref{fig:distractors_qwen3_factscore} (a)-(d) show that as the distractor rate increases, empirical factuality (EF) drops sharply. This degradation occurs because adding distractors to the test set violates the exchangeability assumption underlying the conformal factuality framework, causing EF to fall below the target level. Although EF can increase when the target factuality is set very high, this comes at the cost of a substantial loss in power, as shown in Figure \ref{fig:distractors_qwen3_factscore} (e). Our results indicate that the conformal filtering with current scoring functions is not robust to distractors, underscoring the need for improved scoring functions that maintain robustness in their presence. 

\subsection{Can Distraction-Aware Threshold Help?} \label{sec:distraction_aware}

A natural attempt to mitigate the effect of distractors is to anticipate their presence by using distraction-aware calibration. To study the efficacy of this approach, we extend the previous setting by introducing distractors not only into the test set but also into the calibration set. This models potential distribution shifts caused by distractor content and evaluates whether conformal filtering remains reliable when calibration data include such claims. Since the true level of distractors in practice is unknown, we vary the amount of distractors in calibration set keeping the fraction of distractors in the test sets to $0.25$. This setup enables us to assess both under-estimation and over-estimation of distractor prevalence. Experiments are conducted on FActScore, MATH-1K, and NQ-1K datasets.

\begin{figure}[h]
  \centering
  \includegraphics[width=\linewidth]{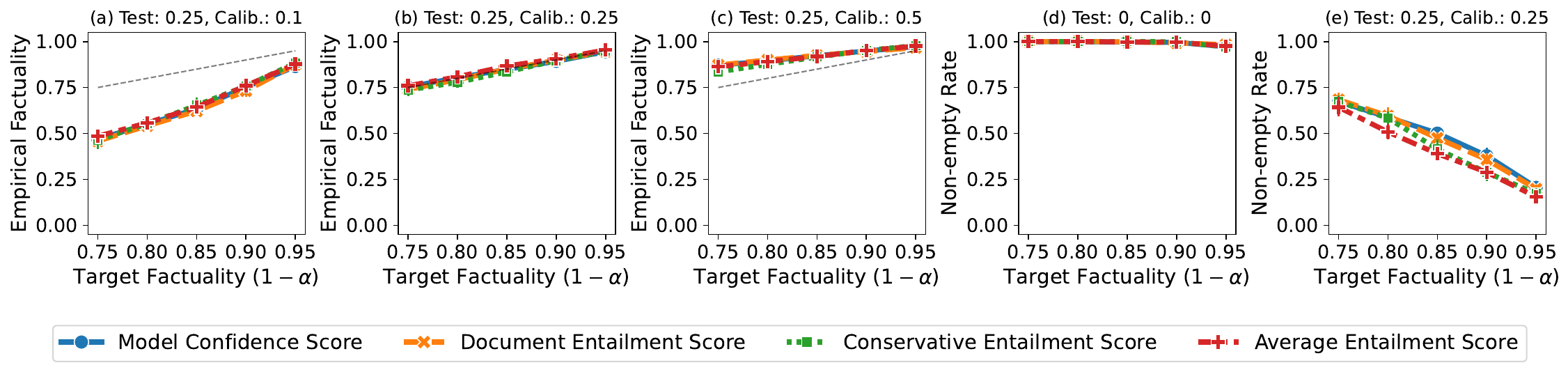}
  \caption{(a) - (c) Empirical factuality versus target factuality ($1 - \alpha$) for \texttt{Qwen3-4B} on the FActScore dataset when the test set is injected with 25\% distractors. We vary the proportion of distractors in the calibration set from (0.1, 0.25, 0.5). As the proportion matches, we can see that the empirical factuality rises to the $y=x$ line. (d)-(e) Comparison of non-empty rate when both the test and calibration sets contain no distractors and contain 25\% of the distractors. Although introducing distractors to the calibration set can achieve target factuality, the non-empty rate suffers.}
  \label{fig:distraction_aware_qwen3_factscore}
\end{figure}

\textbf{Results.} In Figure~\ref{fig:distraction_aware_qwen3_factscore}, we show the result for the setting with a test set with a distractor proportion of 0.25 and varying the levels of distractors in the calibration set. When the fraction of distractors is underestimated, the EF is still far below the target EF (Figure \ref{fig:distraction_aware_qwen3_factscore} (a)). From Figure \ref{fig:distraction_aware_qwen3_factscore} (b)-(c), we can see that introducing a large enough fraction of distractors to the calibration set can bring up the empirical factuality. However, we note that this incurs a high cost on the non-empty rate. As we see in Figure \ref{fig:distraction_aware_qwen3_factscore} (d)-(e), the non-empty rate drops significantly when distractors are introduced to the calibration set. When both calibration and test contain no distractors, we have a much higher non-empty rate compared to the case where we inject 25\% distractors into both the calibration and test sets. This happens likely due to the thresholds found by the conformal filtering framework becoming more stringent. Therefore, the scoring functions cannot distinguish the distractors that are factually incorrect from the factually correct claims.

Overall, \textbf{\emph{the experiments on robustness show that the conformal filtering framework is not robust to distribution shifts and distractors.}} This calls attention to the need for re-thinking factuality guarantees for LLM outputs with robustness as an important criteria.

\section{Efficiency Evaluation of the Conformal Factuality Pipeline End-to-End} \label{sec:end_to_end}


Providing factuality guarantees on the output of an LLM requires additional inference and therefore is necessarily increases the expense of the final response. In this section, we evaluate the complete end-to-end pipeline, jointly considering generation, scoring, and conformal filtering with a particular focus on efficiency measured by FLOPs. 
We use \texttt{gpt-oss-20b}, a mixture-of-experts (MoE) model with 3.6B active parameters, as the response generator $G$ and consider two options for the scoring function $f$: (i) using \texttt{gpt-oss-20b} itself as the scorer, reflecting the scenario in which the same model is available for both generation and scoring; and (ii) using \texttt{Qwen3-8B} as the scorer, allowing us to examine whether a dense model can serve as an alternative to a FLOPs-efficient MoE model. Finally, we compare these LLM-based model confidence scorers with entailment-based scoring while analyzing the associated computational costs. All experiments in this section are conducted on the FActScore dataset.

\begin{table}[htbp]
  \centering
  \begin{tabular}{lrr}
    \toprule
    \textbf{Model} & \textbf{Total Tokens} & \textbf{Total FLOPs} \\
    \midrule
    \texttt{gpt-oss-20b} (3.6B active) \citep{agarwal2025gpt}  & 2000 & $1.44 \times 10^{13}$ \\
    \texttt{Qwen3-8B}   (8.19B active) \citep{yang2025qwen3}  & 2000 & $3.28 \times 10^{13}$ \\
    \texttt{DeepSeek-R1} (37B active) \citep{deepseekr1} & 2000 & $1.5 \times 10^{14}$\\
    \texttt{DeBERTa} (184M active) \citep{he2020deberta} & 2000 & $4.9 \times 10^{11}$ \\
    \texttt{RoBERTa} (356M active) \citep{liu2019roberta} & 2000 & $1.6 \times 10^{12}$ \\
    \bottomrule
  \end{tabular}
  \vspace{4pt}
  \caption{Estimated FLOPs for generating 1000 tokens with a 1000-token prompt (assuming KV caching).}
  \label{tab:flops}
\end{table}

\begin{figure}[h]
  \centering
  \includegraphics[width=\linewidth]{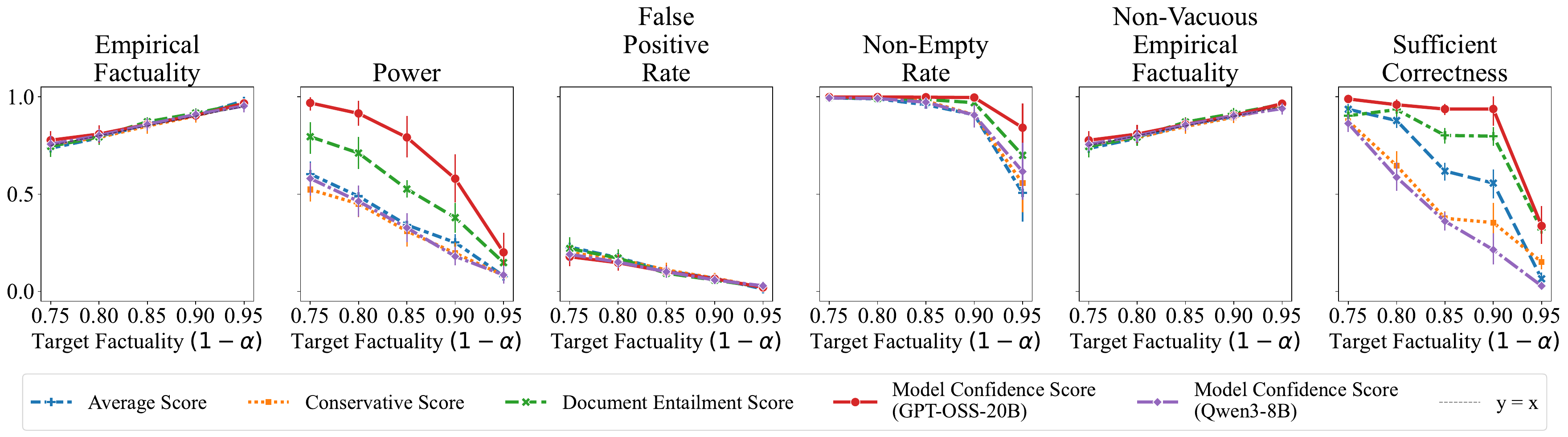}
  \caption{Performance of entailment- and confidence-based scorers using \texttt{gpt-oss-20b} and \texttt{Qwen3-8B} on FActScore.}
  \label{fig:end_to_end}
\end{figure}

\textbf{Results.} Figure~\ref{fig:end_to_end} shows that confidence-based scoring with \texttt{gpt-oss-20b} achieves higher power and sufficient correctness than \texttt{Qwen3-8B}, particularly at moderate target factuality levels. Despite activating substantially fewer parameters, \texttt{gpt-oss-20b} consistently outperforms the dense \texttt{Qwen3-8B}, highlighting the advantages of the MoE architecture for scalable factuality filtering. Table~\ref{tab:flops} further underscores this efficiency–performance trade-off. 
Notably, the document entailment score based on \texttt{DeBERTa} lies at an even more favorable point on this frontier. It is over $100\times$ more computationally efficient than \texttt{gpt-oss-20b}, yet achieves comparable non-empty rates and sufficient correctness. This suggests that \emph{lightweight entailment models can serve as strong surrogates for large LLM-based scorers when computational budgets are constrained, without substantial losses in factuality performance.}
Together, these results demonstrate that both parameter-efficient MoE models and compact entailment-based scorers offer compelling alternatives to dense large models, enabling end-to-end factuality evaluation that is not only more effective but also dramatically more economical.
\section{Related Works}

Many studies show that LLMs are prone to hallucinations~\citep{nadeau2024benchmarking, huang2025survey} despite their impressive capabilities in summarization, dialogue, and coding~\citep{achiam2023gpt, zhang2024comprehensive, nam2024using}. RAG and conformal prediction based filtering methods have emerged as prominent methods to mitigate hallucinations and provide factuality guarantees. 

\subsection{Retrieval-Augmented Generation}

Retrieval-augmented generation (RAG) improves LLM performance on knowledge-intensive tasks by grounding responses in retrieved external context~\citep{lewis2020retrieval, joren2025sufficient, gao2023retrieval}. Formally, given a query $x$, a retriever $R$ returns context $R(x)$ that supplements the model's parametric knowledge, guiding generation toward more factually correct outputs. In this work, we assume access to an oracle retriever that always provides relevant, accurate references. While RAG is powerful, it does not provide statistical guarantees on the factuality of its outputs, and generated responses can still contain hallucinations~\citep{huang2025survey}.

In addition, LLMs may not utilize references effectively, especially information appearing in the middle of the context---a phenomenon known as \emph{lost-in-the-middle}~\citep{liu2023lost, ravaut2023context, chen2023understanding, tang2023found}. This may be exacerbated by positional encoding mechanisms such as rotary positional embeddings (RoPE)~\citep{su2024roformer, huang2025survey}. Another contributing factor is that, in pretraining data, salient information is often located near the beginning or the end of a document rather than in the middle~\citep{ravaut2023context, huang2025survey}. The metric sufficient context is proposed in \citep{joren2025sufficient} to study the usefulness of the retrieved reference in a RAG-based LLMs. In contrast to this metric which focuses on the quality of retrieval, sufficient correctness and conditional sufficient correctness proposed in our work focuses on the quality of output from the RAG-based LLM measured with respect to the reference.

\subsection{Conformal Prediction}

Conformal prediction provides another promising mitigation strategy by filtering non-factual content from LLM outputs~\citep{vovk2005algorithmic, angelopoulos2021gentle, mohri2024language, cherian2024large}. These methods not only improve factuality but also offer a statistical guarantee, e.g., $\mathbb{P}(\text{output is factual}) \geq 1 - \alpha$. For instance, \citet{mohri2024language} introduced \emph{conformal factuality}, which scores claims in the output and removes those below a threshold calibrated on held-out data. The choice of scoring function $f$ is therefore critical to the effectiveness of the framework. 

\subsection{Conformal Prediction and RAG}

Recent work has begun integrating conformal prediction into retrieval-augmented generation (RAG) to provide statistical reliability guarantees. TRAQ \citep{li2023traq} applies conformal prediction at both the retriever and generator stages, ensuring with high probability that a semantically correct answer is included in the output set. Conformal-RAG \citep{feng2025response} instead operates at the sub-claim level, filtering unreliable statements to guarantee factuality across domains. Conflare \citep{rouzrokh2024conflare} focuses on the retrieval stage, calibrating similarity thresholds so that retrieved contexts contain the true answer with user-specified confidence. While these approaches provide important coverage guarantees at different stages of the pipeline, they largely assess correctness in isolation. In contrast, our work provides a systematic analysis of the performance of conformal filtering for guaranteeing factuality of RAG-based LLMs by introducing new metrics—non-empty rate, non-vacuous empirical factuality, sufficient correctness and conditional sufficient correctness—that explicitly capture the trade-off between correctness and informativeness, which existing frameworks do not address. Furthermore, we perform robustness analysis that reveals the fragility of conformal filtering under distribution shifts and in the presence of distractors.  
\section{Conclusion}

We systematically investigate conformal filtering framework used in guaranteeing factuality of RAG-based LLM outputs. Our experiments extensively study the importance of references for generation and scoring, various scoring functions, sensitivity to calibration data, robustness to distractor-induced hallucinations and the efficiency of the end-to-end pipeline.
A key limitation we uncover is that standard factuality measures can overstate practical progress: because they primarily reward \emph{absence of incorrect content}, they can be optimized by filtering systems that abstain (returning empty answers) or produce generic, non-committal responses that are technically ``factual'' yet unhelpful even when the input reference has the information to answer the question.
As a result, high empirical factuality may coincide with low end-task utility, obscuring the correctness--informativeness trade-off that real deployments must navigate.
To address these limitations, we introduced novel metrics---\emph{non-empty rate}, \emph{non-vacuous empirical factuality}, and \emph{(conditional) sufficient correctness}---that explicitly measure whether filtered outputs remain informative and sufficient for answering the query along with empirical factuality.

Our experiments span three datasets for diverse tasks -- FActScore for open-ended summarization,  MATH for mathematical queries, and Natural Questions and multiple model families, revealing several insights.
In particular, we show that stronger factuality does not require larger or more expensive verifiers: lightweight entailment-based verifiers consistently outperform LLM-based confidence scorers while requiring orders of magnitude fewer FLOPs. Our comprehensive analysis of robustness reveals that the current conformal filtering approaches are not robust under distribution shift and distractors,  which is an important limitation for practical usage in safety critical settings. 


Overall, our findings provide actionable guidance for building reliable and efficient RAG systems and underscore the need to rethink how factuality in LLMs is measured, enforced, and optimized under realistic deployment constraints with a particular focus on robustness and usefulness.


\bibliographystyle{unsrtnat}
\bibliography{references}

\appendix
\section{Appendix}
 
 
This appendix provides supplementary material that supports and extends the analyses presented in the main paper. It is organized as follows.
 
\textbf{Appendix~\ref{app:extended_results}} presents extended experimental results that complement the main findings. Section~\ref{app:impact_of_reference} broadens the analysis of how retrieved references improve generation quality (Section~\ref{sec:impact_of_reference}) to additional model families (\texttt{Llama-3.x}, \texttt{SmolLM2}) and frontier models. Section~\ref{app:prompting_strategies} provides the full set of prompting-strategy comparisons across all target factuality levels, extending the summary in Section~\ref{sec:prompting_strategy}. Section~\ref{app:role_of_reference_scoring} expands the reference-in-scoring analysis (Section~\ref{sec:role_of_reference_scoring}) across four model families and three datasets. Section~\ref{app:model_choice_scorers} supplements the scorer scaling study (Section~\ref{sec:model_choice_scorers}) with detailed radar plots at multiple factuality targets. Section~\ref{app:sc_and_csc} reports sufficient correctness and conditional sufficient correctness for model families beyond \texttt{Qwen3} (Section~\ref{sec:sc_and_csc}). Section~\ref{app:distribution_shift} and Section~\ref{app:adversarial_distractors} extend the robustness analyses of Sections~\ref{sec:distribution_shift} and \ref{sec:distractors} to additional scoring functions and model families. Section~\ref{app:adversarial_calibration} complements Section~\ref{sec:distraction_aware} with full results on distraction-aware calibration.
 
\textbf{Appendix~\ref{app:distractors}} details how adversarial distractor claims are generated and verified, supporting the experimental protocol in Section~\ref{sec:distractors}.
 
\textbf{Appendix~\ref{app:human_evaluation}} reports the human evaluation used to validate \texttt{gpt-5-nano} as a factuality judge.
 
\textbf{Appendix~\ref{app:prompts}} collects all prompts used throughout our experiments, including those for generation, parsing, labeling, scoring, merging, correctness evaluation, and distractor creation.
 
\subsection{Extended Results} \label{app:extended_results}
 
The following subsections present extended experimental results that complement the main-paper analyses. Each subsection identifies the corresponding main-paper section and provides additional figures covering model families, datasets, or configurations not shown in the main text.
 
\subsubsection{Impact of References} \label{app:impact_of_reference}
 
Section~\ref{sec:impact_of_reference} of the main paper demonstrates that providing retrieved references to the response generator consistently improves sufficient correctness, using the \texttt{Qwen3} family as the primary illustration. Here we first describe how sufficient correctness is measured and then verify that this finding generalizes across model families and scales.

\paragraph{Measuring sufficient correctness}
For each original generated output $y$, we use \texttt{gpt-5-nano} with the prompt provided in Appendix~\ref{sec:prompt_sc} to assess whether $y$ is sufficient correct. Concretely, we replace the \texttt{\{response\}} placeholder in the prompt with the string representation of $y$. At the same time, we replace \texttt{\{query\}} and \texttt{\{reference\}} with the corresponding query and reference, respectively.
 
\begin{figure}[htbp]
  \centering
  \includegraphics[width=\linewidth]{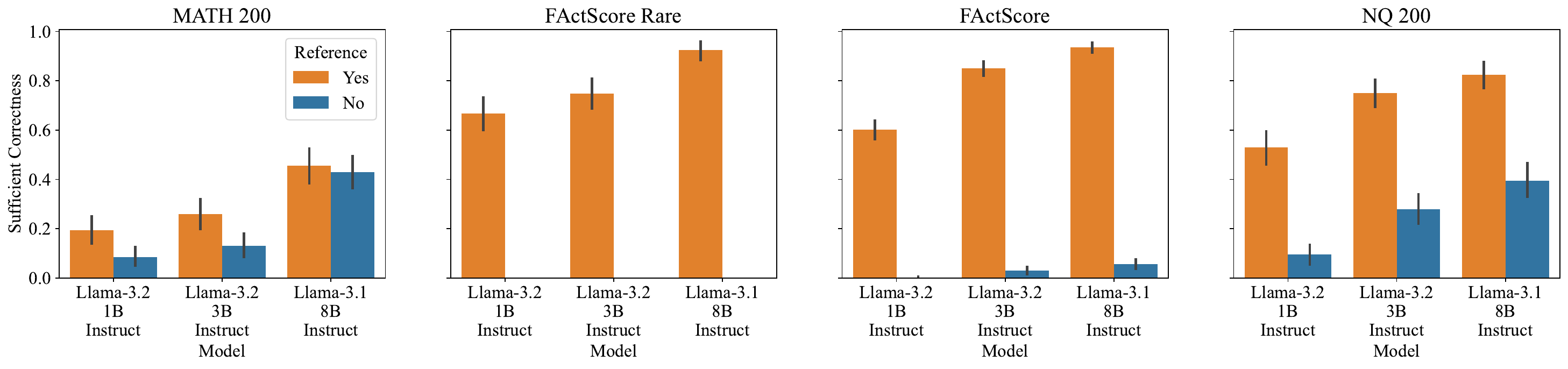}
  \caption{Sufficient correctness (SC) of \texttt{Llama-3.x} models on four datasets (MATH-200, FActScore-Rare, FActScore, NQ-200), with and without access to references. Across model sizes and datasets, providing references consistently improves generation quality.}
  \label{fig:impact_of_references_llama3}
\end{figure}

\paragraph{Llama-3.x family.}
Figure~\ref{fig:impact_of_references_llama3} extends the reference-impact analysis to the \texttt{Llama-3.x} family. Consistent with the \texttt{Qwen3} results, all three model sizes---1B, 3B, and 8B---show clear gains in sufficient correctness when references are provided, across all four datasets. The improvement is especially pronounced on FActScore and NQ-200, where parametric knowledge alone is insufficient.
 
\begin{figure}[htbp]
  \centering
  \includegraphics[width=\linewidth]{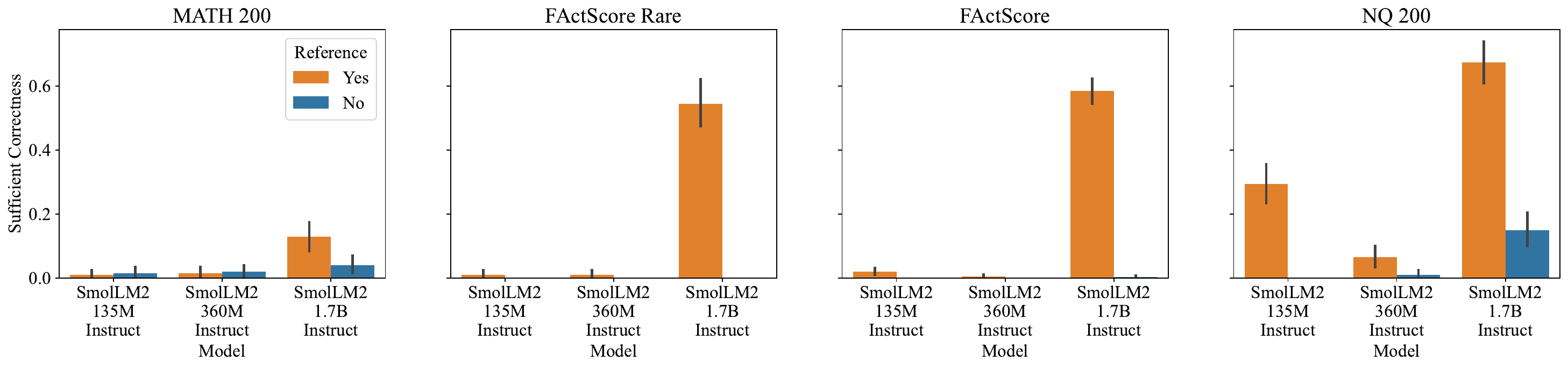}
  \caption{Sufficient correctness (SC) of \texttt{SmolLM2} models on four datasets (MATH-200, FActScore-Rare, FActScore, NQ-200), with and without access to references. Across model sizes and datasets, providing references consistently improves generation quality.}
  \label{fig:impact_of_references_smollm2}
\end{figure}

\paragraph{SmolLM2 family.}
Figure~\ref{fig:impact_of_references_smollm2} presents the same comparison for the \texttt{SmolLM2} family. Despite their substantially smaller parameter counts (135M--1.7B), these models also benefit from references, although the absolute level of sufficient correctness remains lower than that of the larger families. This confirms that reference grounding is beneficial even at very small scales.
 
\begin{figure}[htbp]
  \centering
  \includegraphics[width=0.5\linewidth]{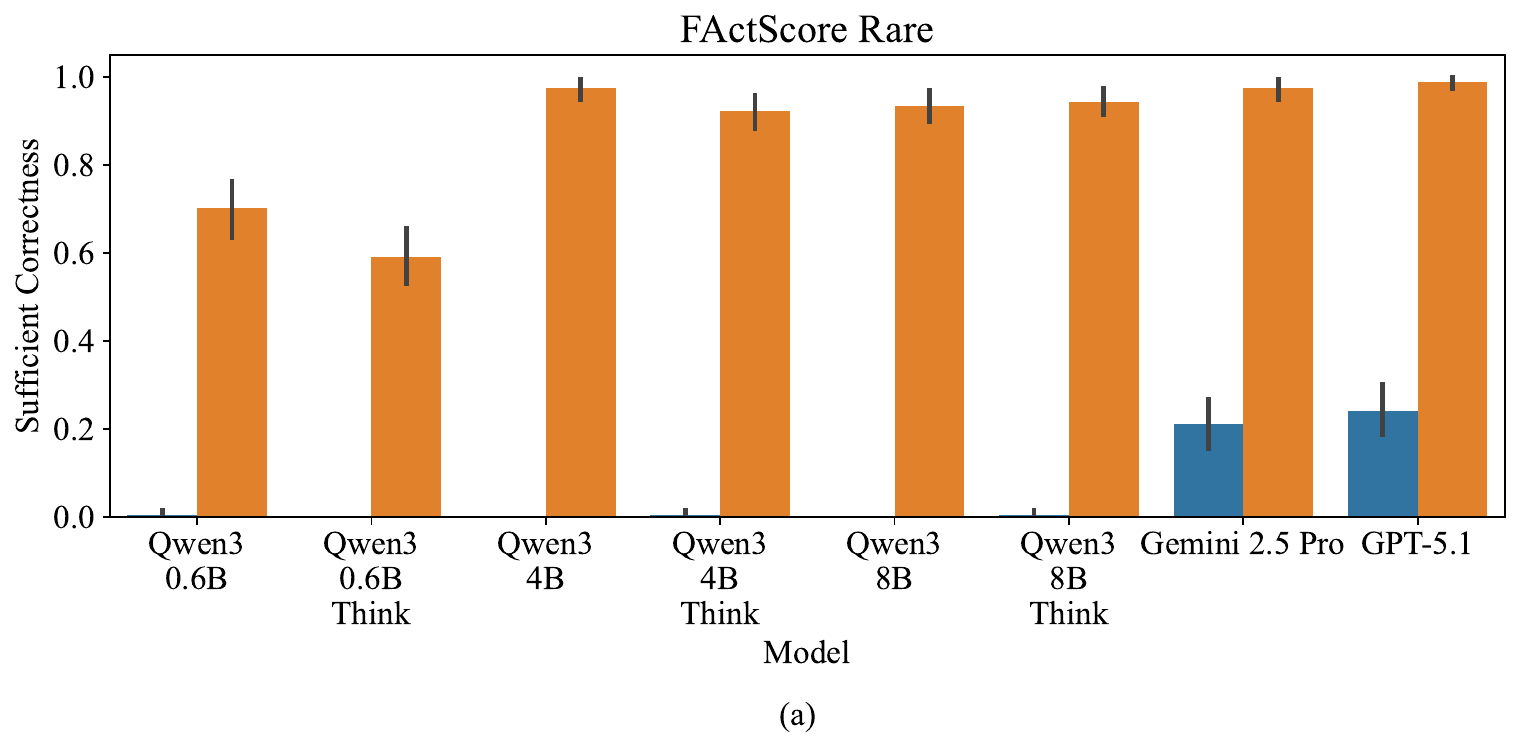}
  \caption{Sufficient correctness (SC) of \texttt{Qwen3} and frontier models on FActScore-Rare, with and without access to references. Providing references consistently improves generation quality; notably, \texttt{Qwen3-4B} with references is comparable to frontier models.}
  \label{fig:impact_of_references_frontier}
\end{figure}

\paragraph{Frontier model comparison.}
Finally, Figure~\ref{fig:impact_of_references_frontier} compares the \texttt{Qwen3} family against two frontier models---\texttt{gemini-2.5-pro}~\citep{team2023gemini} and \texttt{gpt-5.1}~\citep{gpt5}---on the FActScore-Rare dataset. A noteworthy finding is that \texttt{Qwen3-4B}, when given a reference, achieves sufficient correctness comparable to these frontier models. This underscores that, with proper retrieval augmentation, even moderately sized models can match frontier-level factual accuracy on knowledge-intensive tasks.
 
\clearpage
\subsubsection{Prompting Strategies for Scorers} \label{app:prompting_strategies}
 
Figure~\ref{fig:prompting_strategy} in Section~\ref{sec:prompting_strategy} summarizes the effect of prompting strategies on LLM-based scoring functions using three representative configurations per model. Here we present the full set of comparisons at five target factuality levels ($1-\alpha \in \{0.75, 0.8, 0.85, 0.9, 0.95\}$), enabling a more granular assessment of how each prompting dimension---evidence highlighting, chain-of-thought reasoning, scalar vs.\ Boolean scoring, and consistency averaging---interacts with the target factuality.
 
\begin{figure}[htbp]
  \centering
  \includegraphics[width=\linewidth]{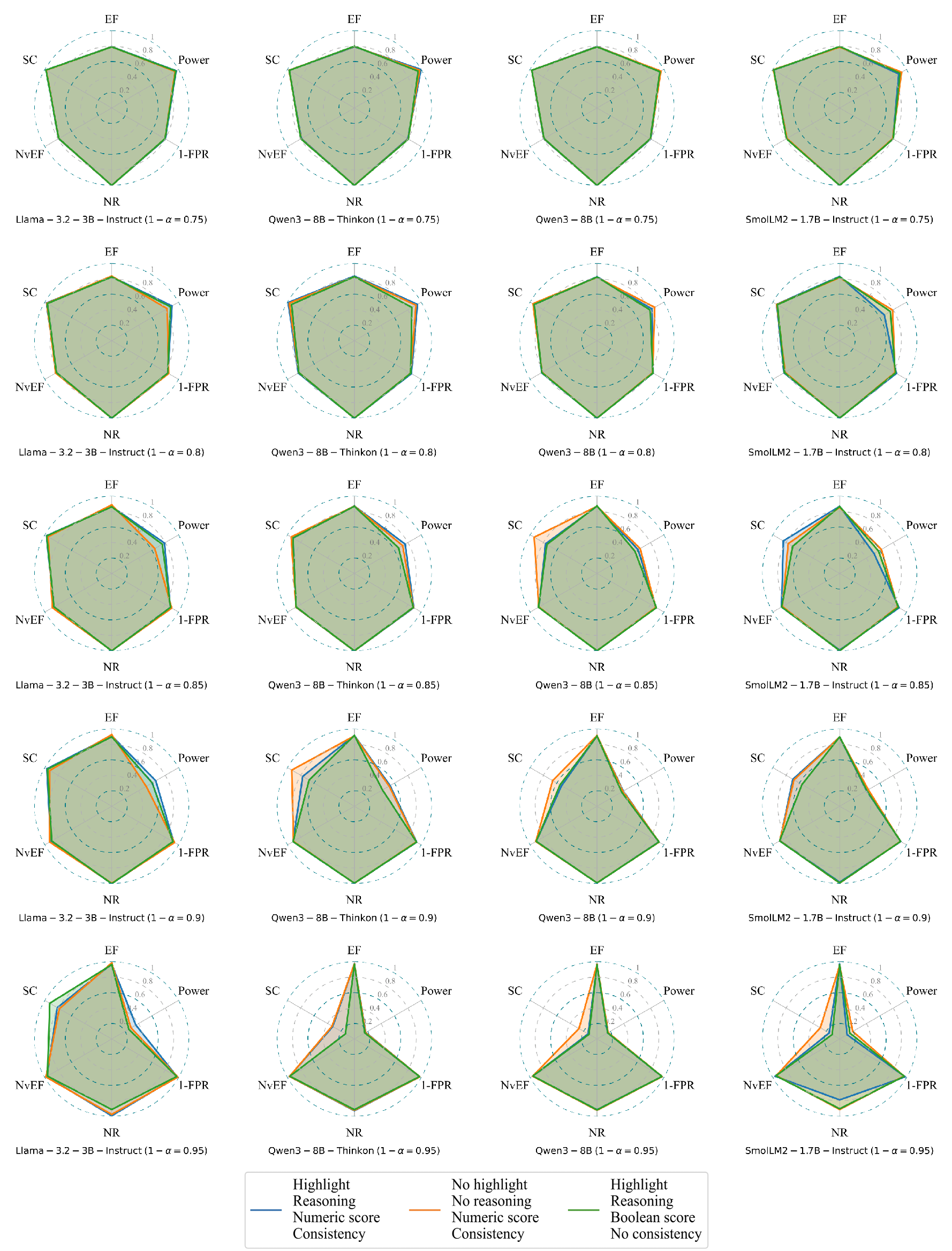}
  \caption{Overall performance comparison of model confidence scores across 16 prompting strategies at five target factuality levels ($1-\alpha$) on the FActScore dataset. Each column corresponds to a different LLM scorer; each row corresponds to a different target factuality. Numeric scoring and consistency averaging yield the most robust improvements.}
  \label{fig:16prompting}
\end{figure}

\paragraph{FActScore dataset.}
Figure~\ref{fig:16prompting} shows radar plots for all 16 prompting strategies across four LLMs on FActScore. The plots confirm the main-paper finding that numeric scoring and consistency averaging are the most reliably beneficial dimensions, while chain-of-thought and highlighting provide inconsistent gains that vary by model and factuality target.
 
\begin{figure}[htbp]
  \centering
  \includegraphics[width=\linewidth]{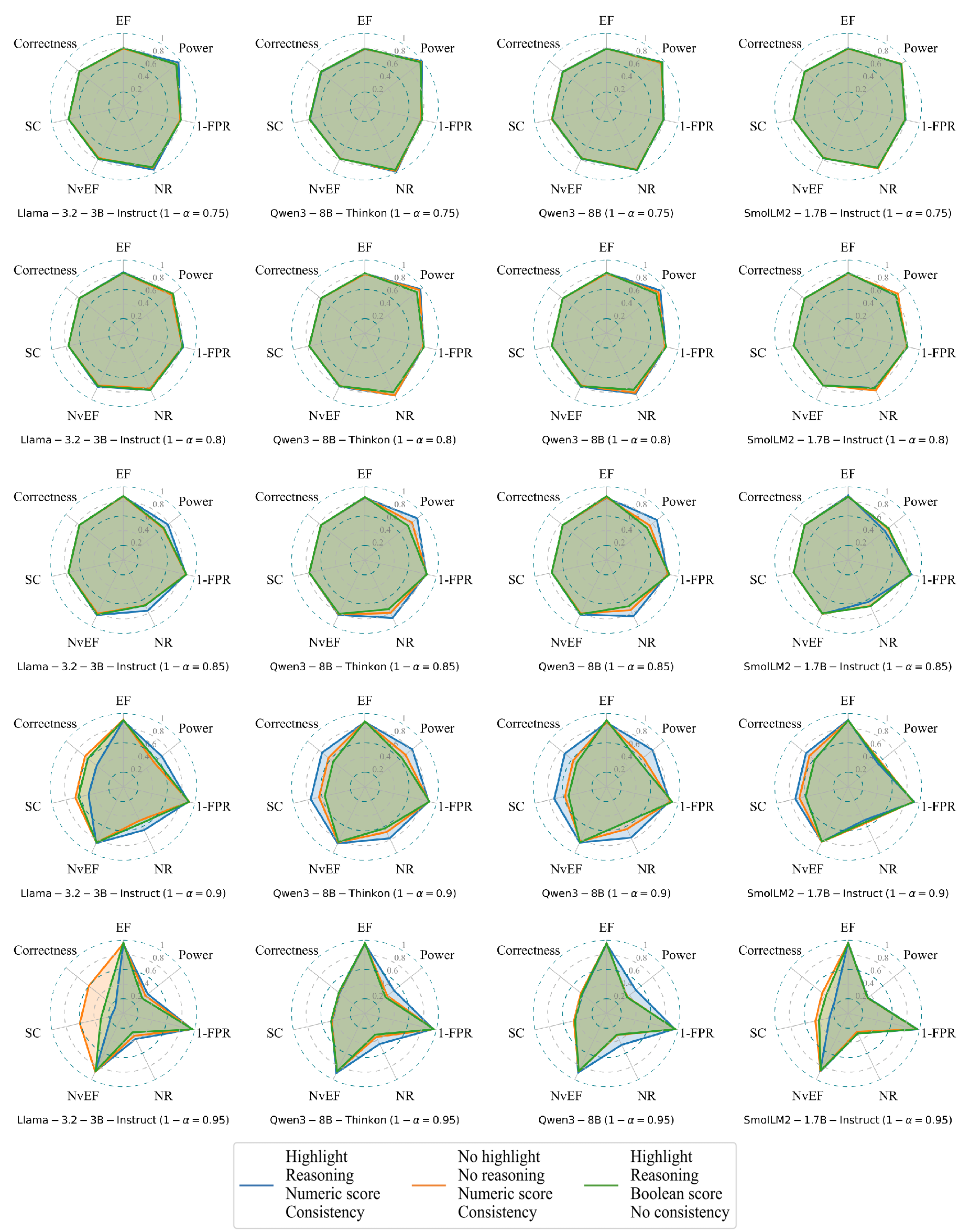}
  \caption{Overall performance comparison of model confidence scores across various prompting strategies at five target factuality levels ($1-\alpha$) on the MATH-1K dataset. The advantage of numeric over Boolean scoring is especially clear for mathematical reasoning.}
  \label{fig:16prompting_math}
\end{figure}

\paragraph{MATH-1K dataset.}
Figure~\ref{fig:16prompting_math} repeats this analysis on MATH-1K. On mathematical reasoning tasks, the relative advantage of numeric scoring is even more pronounced, likely because scalar confidence values better capture the degree of certainty in multi-step derivations than binary labels.
 
\clearpage
\subsubsection{Role of References in Scoring} \label{app:role_of_reference_scoring}
 
Section~\ref{sec:role_of_reference_scoring} demonstrates that providing references to the scoring function improves power and non-empty rate, using \texttt{Qwen3-4B} on MATH-1K as the primary example. Here we extend this analysis to additional scorer models and datasets to assess how broadly the benefit holds.
 
\begin{figure}[htbp]
  \centering
  \includegraphics[width=\linewidth]
  {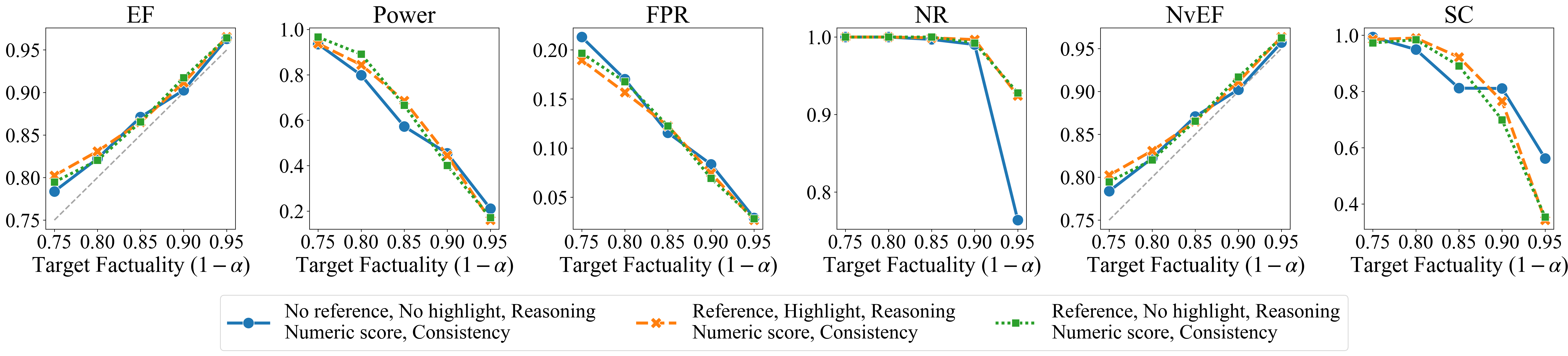}
  \caption{Performance of model confidence score on FActScore with and without reference provided to scoring functions, using \texttt{gpt-5-nano} as generator and \texttt{Qwen3-4B} as scorer. Reference access improves power and non-empty rate.}
  \label{fig:F1_gpt_5_nano_Qwen3_4B_Factscore}
\end{figure}

\begin{figure}[htbp]
  \centering
  \includegraphics[width=\linewidth]
  {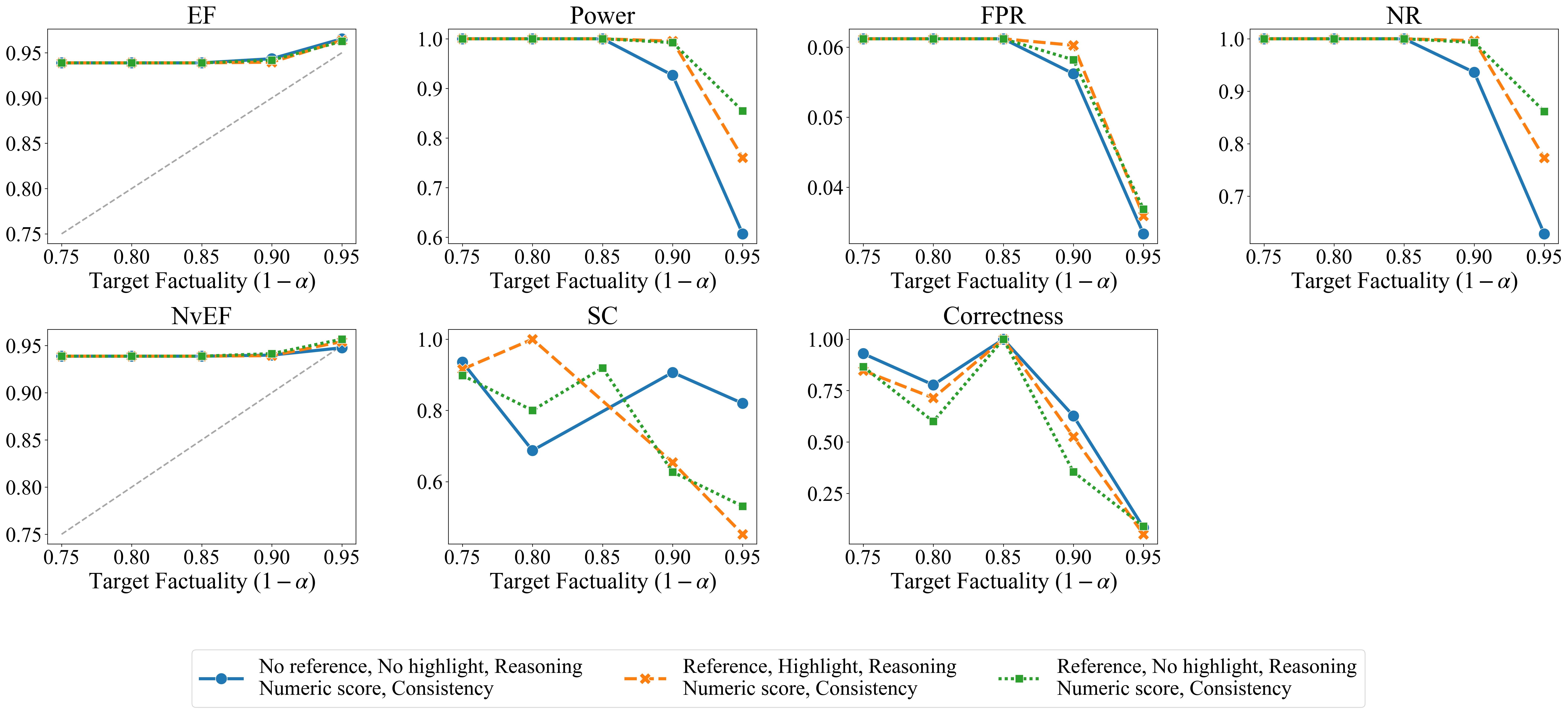}
  \caption{Performance of model confidence score on NQ-1K with and without reference provided to scoring functions, using \texttt{gpt-5-nano} as generator and \texttt{Qwen3-4B} as scorer. The benefit of reference access is consistent with the MATH-1K and FActScore results.}
  \label{fig:F1_gpt_5_nano_Qwen3_4B_NQ_1K}
\end{figure}

\paragraph{Qwen3-4B across datasets.}
Figures~\ref{fig:F1_gpt_5_nano_Qwen3_4B_Factscore} and~\ref{fig:F1_gpt_5_nano_Qwen3_4B_NQ_1K} show the reference effect for \texttt{Qwen3-4B} on FActScore and NQ-1K, respectively. On both datasets, providing the reference to the scorer consistently improves power and non-empty rate, mirroring the MATH-1K results in the main paper. The gains are largest at high target factuality, where the scoring function must be most discriminating.
 
\begin{figure}[htbp]
  \centering
  \includegraphics[width=\linewidth]
  {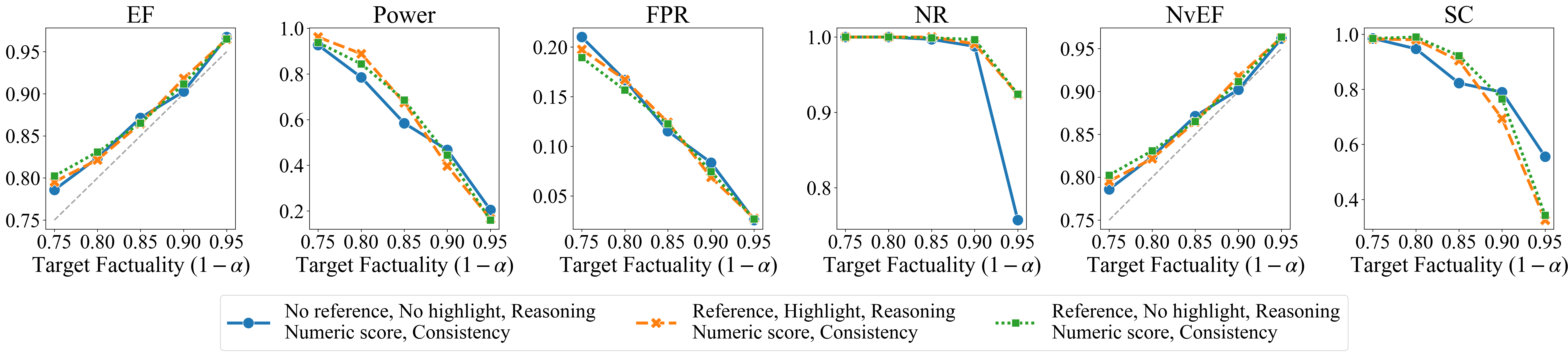}
  \caption{Performance of model confidence score on FActScore with and without reference provided to scoring functions, using \texttt{gpt-5-nano} as generator and \texttt{Qwen3-8B} as scorer.}
  \label{fig:F1_gpt_5_nano_Qwen3_8B_Factscore}
\end{figure}

\begin{figure}[htbp]
  \centering
  \includegraphics[width=\linewidth]
  {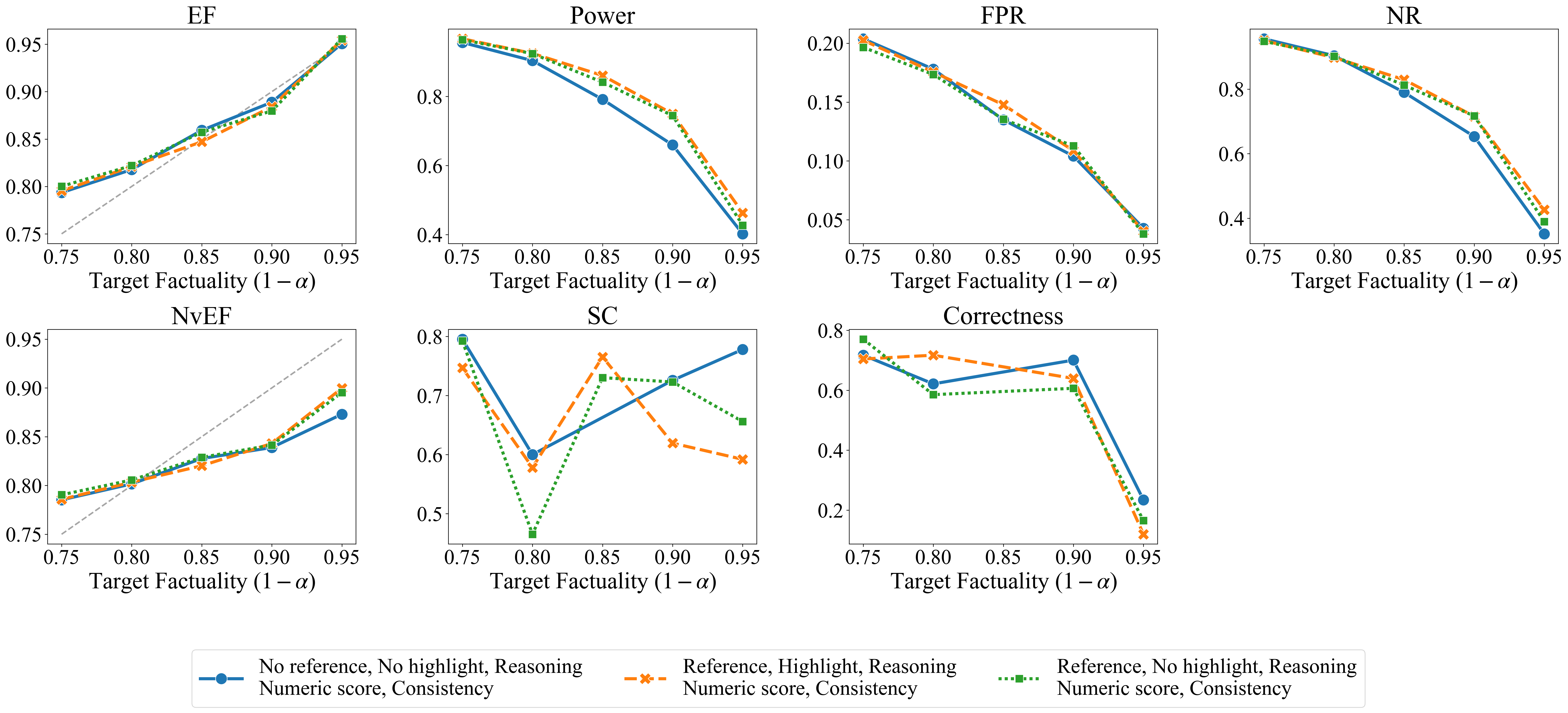}
  \caption{Performance of model confidence score on MATH-1K with and without reference provided to scoring functions, using \texttt{gpt-5-nano} as generator and \texttt{Qwen3-8B} as scorer.}
  \label{fig:F1_gpt_5_nano_Qwen3_8B_MATH_1K}
\end{figure}

\begin{figure}[htbp]
  \centering
  \includegraphics[width=\linewidth]
  {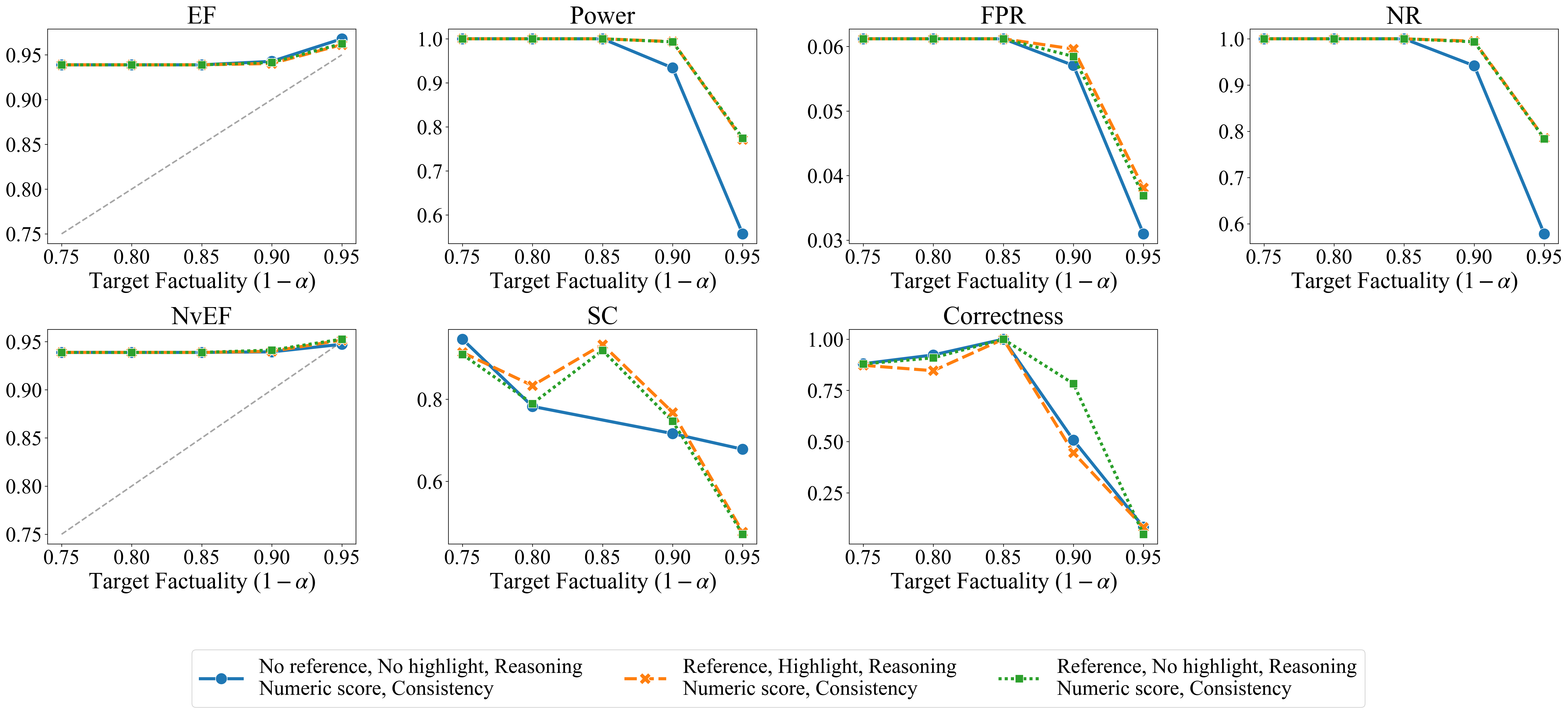}
  \caption{Performance of model confidence score on NQ-1K with and without reference provided to scoring functions, using \texttt{gpt-5-nano} as generator and \texttt{Qwen3-8B} as scorer.}
  \label{fig:F1_gpt_5_nano_Qwen3_8B_NQ_1K}
\end{figure}

\clearpage
\paragraph{Qwen3-8B across datasets.}
Figures~\ref{fig:F1_gpt_5_nano_Qwen3_8B_Factscore}--\ref{fig:F1_gpt_5_nano_Qwen3_8B_NQ_1K} repeat the analysis with \texttt{Qwen3-8B}. The trends are qualitatively similar to \texttt{Qwen3-4B}: reference access yields consistent improvements, although the magnitude of the gain varies across datasets. This suggests that the benefit of feeding references to the scorer is not an artifact of a particular model scale.
 
\begin{figure}[htbp]
  \centering
  \includegraphics[width=\linewidth]
  {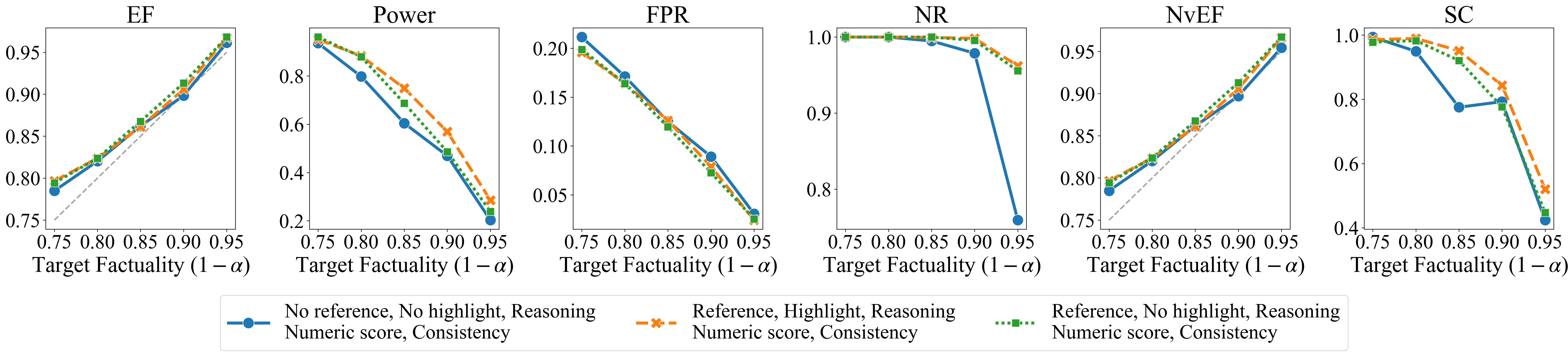}
  \caption{Performance of model confidence score on FActScore with and without reference provided to scoring functions, using \texttt{gpt-5-nano} as generator and \texttt{Llama-3.2-3B-Instruct} as scorer.}
  \label{fig:F1_gpt_5_nano_Llama_3_2_3B_Factscore}
\end{figure}

\begin{figure}[htbp]
  \centering
  \includegraphics[width=\linewidth]
  {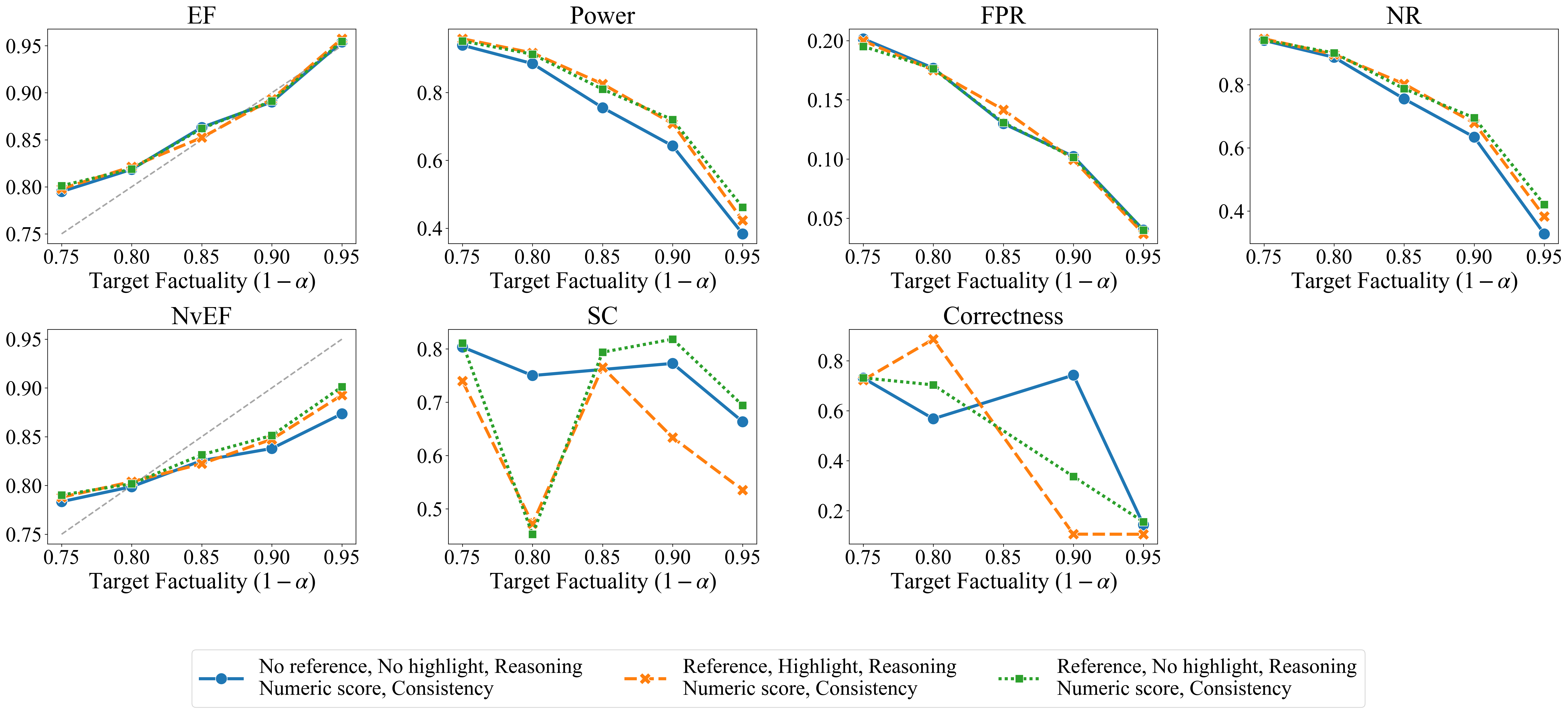}
  \caption{Performance of model confidence score on MATH-1K with and without reference provided to scoring functions, using \texttt{gpt-5-nano} as generator and \texttt{Llama-3.2-3B-Instruct} as scorer.}
  \label{fig:F1_gpt_5_nano_Llama_3_2_3B_MATH}
\end{figure}

\begin{figure}[htbp]
  \centering
  \includegraphics[width=\linewidth]
  {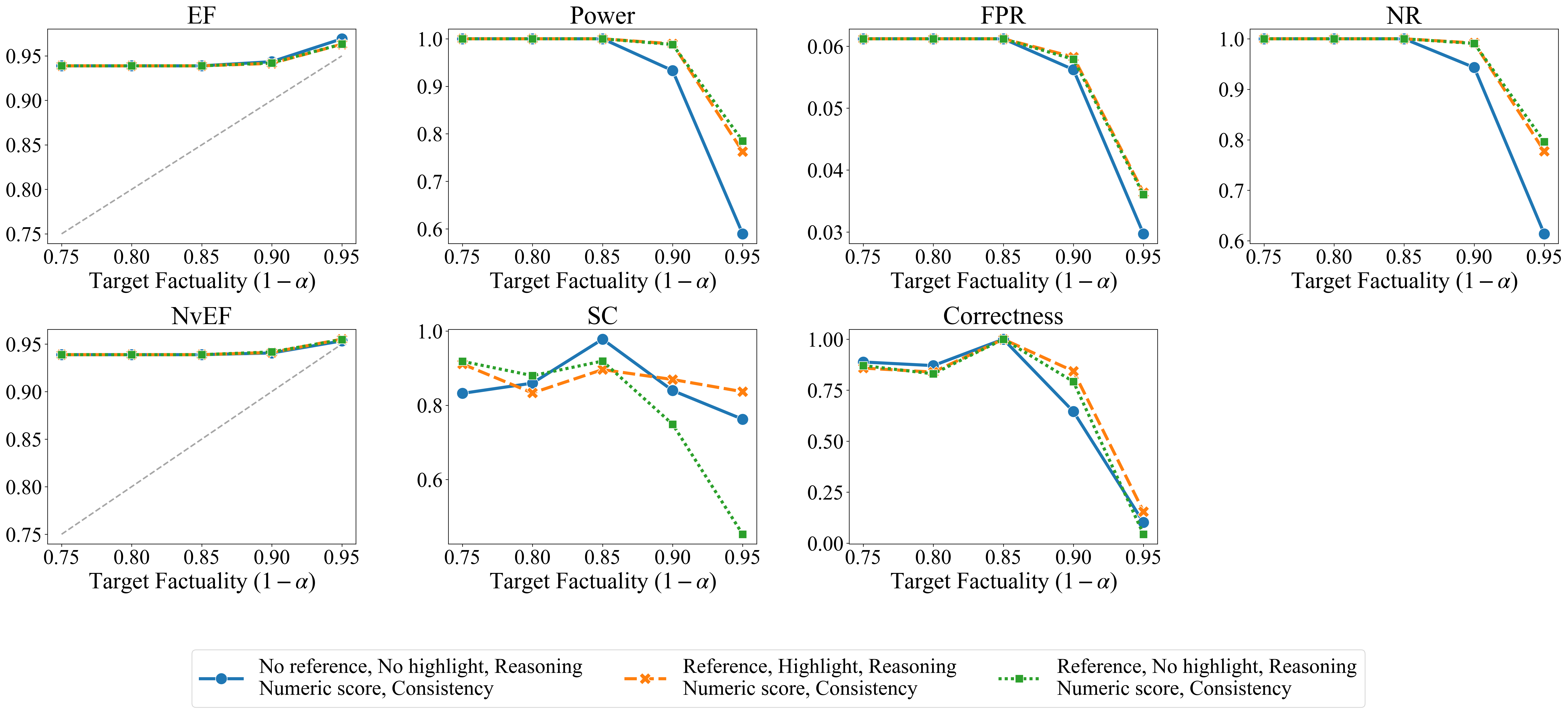}
  \caption{Performance of model confidence score on NQ-1K with and without reference provided to scoring functions, using \texttt{gpt-5-nano} as generator and \texttt{Llama-3.2-3B-Instruct} as scorer.}
  \label{fig:F1_gpt_5_nano_Llama_3_2_3B_NQ}
\end{figure}

\clearpage
\paragraph{Llama-3.2-3B across datasets.}
Figures~\ref{fig:F1_gpt_5_nano_Llama_3_2_3B_Factscore}--\ref{fig:F1_gpt_5_nano_Llama_3_2_3B_NQ} present the same comparison for \texttt{Llama-3.2-3B-Instruct}. The reference benefit is again observed, confirming that it is not specific to the \texttt{Qwen3} family. Notably, on FActScore, the improvement in sufficient correctness from adding a reference is among the largest we observe across all model--dataset pairs.
 
\begin{figure}[htbp]
  \centering
  \includegraphics[width=\linewidth]
  {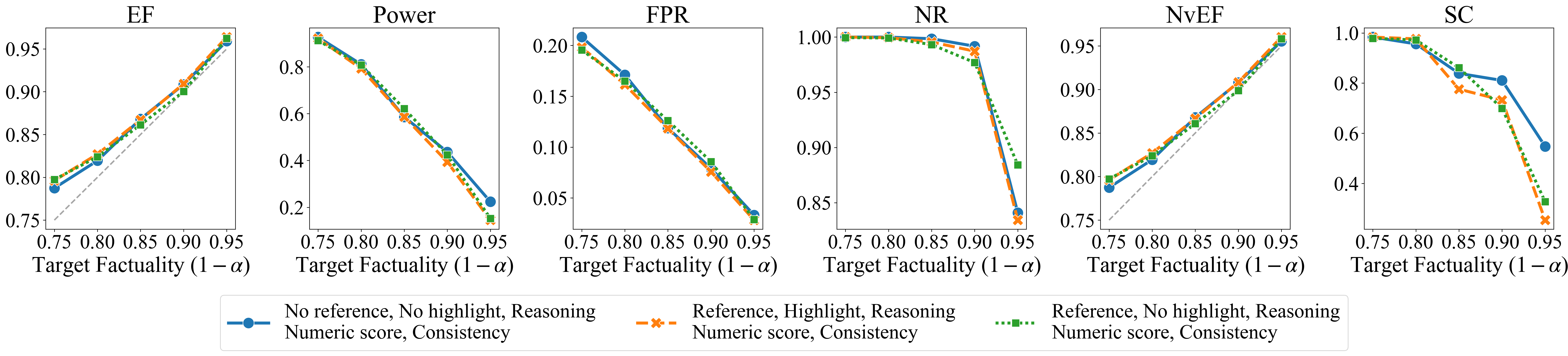}
  \caption{Performance of model confidence score on FActScore with and without reference provided to scoring functions, using \texttt{gpt-5-nano} as generator and \texttt{SmolLM2-1.7B-Instruct} as scorer.}
  \label{fig:F1_gpt_5_nano_SmolLM2_1_7_B_Factscore}
\end{figure}

\begin{figure}[htbp]
  \centering
  \includegraphics[width=\linewidth]
  {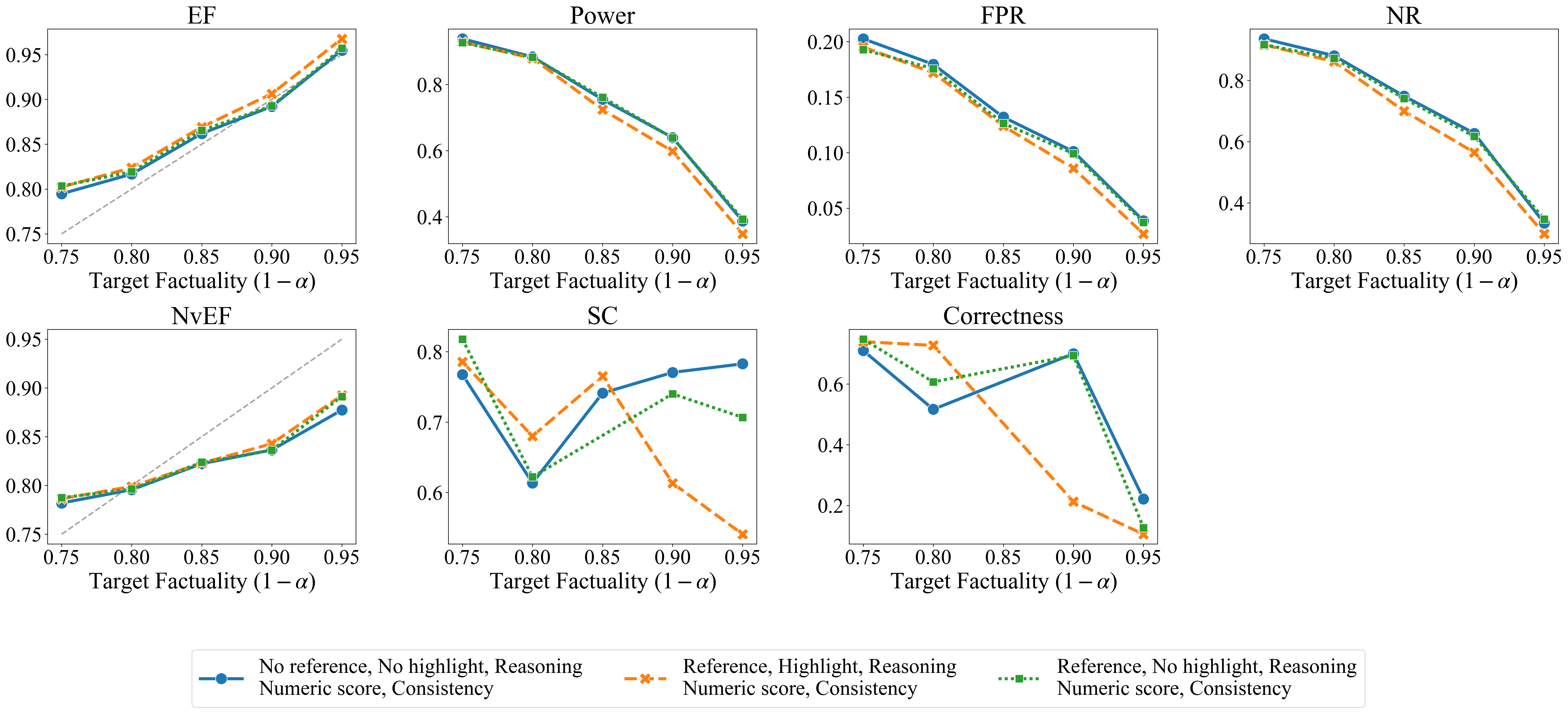}
  \caption{Performance of model confidence score on MATH-1K with and without reference provided to scoring functions, using \texttt{gpt-5-nano} as generator and \texttt{SmolLM2-1.7B-Instruct} as scorer.}
  \label{fig:F1_gpt_5_nano_SmolLM2_1_7_B_MATH}
\end{figure}

\begin{figure}[htbp]
  \centering
  \includegraphics[width=\linewidth]
  {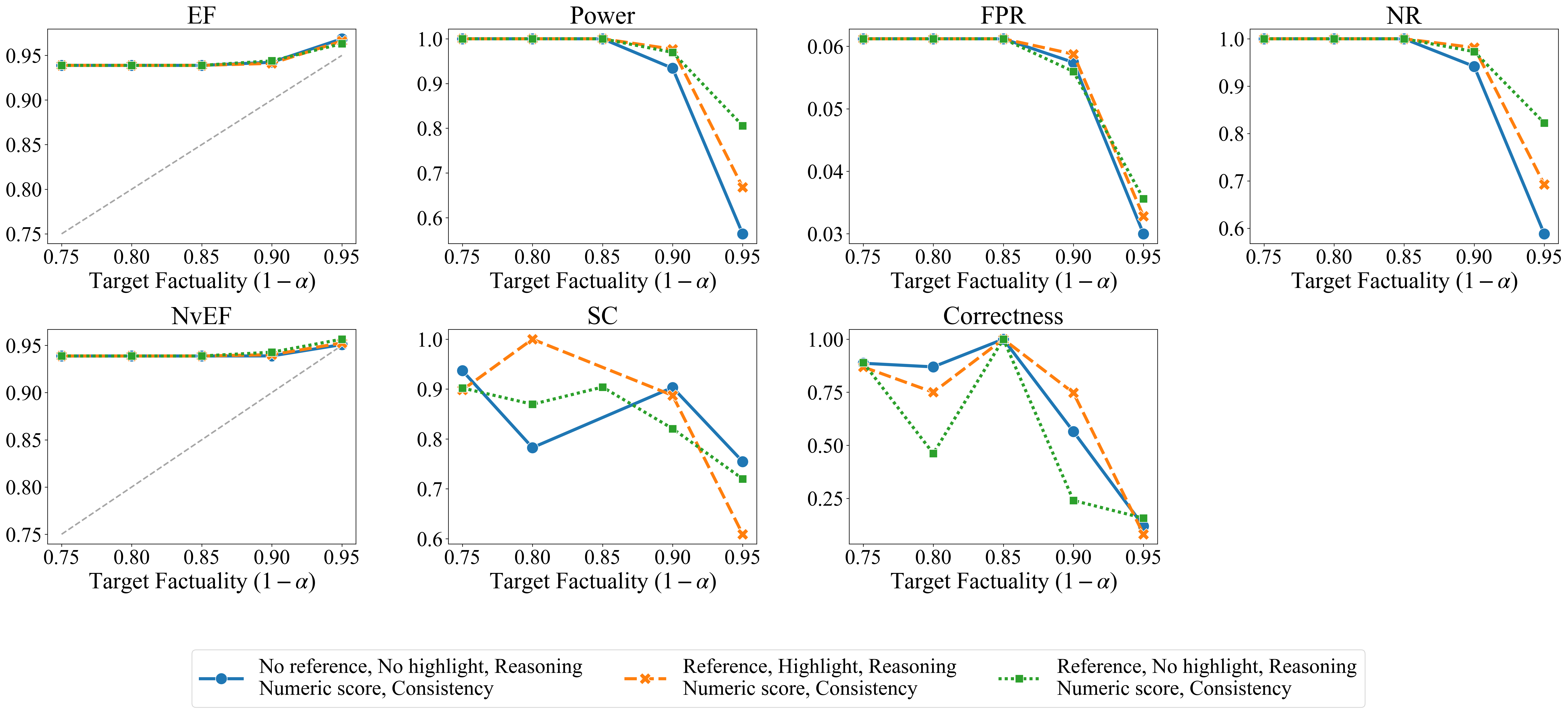}
  \caption{Performance of model confidence score on NQ-1K with and without reference provided to scoring functions, using \texttt{gpt-5-nano} as generator and \texttt{SmolLM2-1.7B-Instruct} as scorer.}
  \label{fig:F1_gpt_5_nano_SmolLM2_1_7_B_NQ}
\end{figure}

\clearpage
\paragraph{SmolLM2-1.7B across datasets.}
Finally, Figures~\ref{fig:F1_gpt_5_nano_SmolLM2_1_7_B_Factscore}--\ref{fig:F1_gpt_5_nano_SmolLM2_1_7_B_NQ} show results for \texttt{SmolLM2-1.7B-Instruct}, the smallest LLM-based scorer in our study. Even at this scale, reference access improves scoring performance, although the absolute gains are more modest. This is consistent with the hypothesis that smaller models have less capacity to leverage long reference contexts, but still benefit from the additional grounding signal.
 
\clearpage
\subsubsection{Model Choice for Scorers} \label{app:model_choice_scorers}
 
Section~\ref{sec:model_choice_scorers} examines how the scorer model family and scale affect factuality filtering, using \texttt{gpt-5-nano} as the generator. The main paper presents a condensed summary; here we provide the full set of radar plots at each target factuality level ($1-\alpha \in \{0.75, 0.8, 0.85, 0.9, 0.95\}$) for all three model families (\texttt{Llama-3.x}, \texttt{Qwen3}, and \texttt{SmolLM2}).
 
\begin{figure}[htbp]
  \centering
  \includegraphics[width=\linewidth]{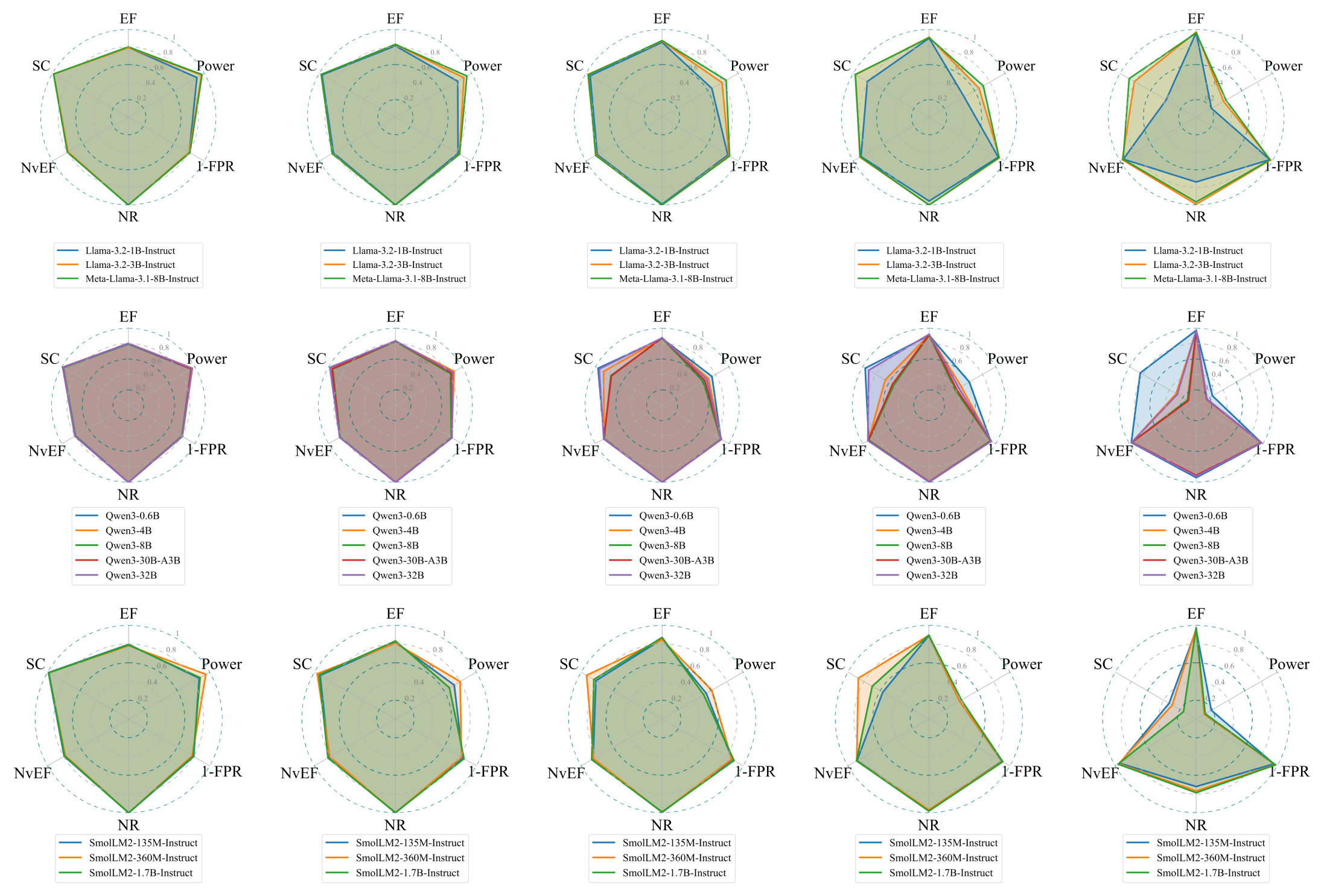}
  \caption{Overall performance comparison of model confidence scores across different model scales and families at five target factuality levels ($1-\alpha$) on the FActScore dataset. While \texttt{Llama-3.x} models show consistent gains with scale, \texttt{Qwen3} and \texttt{SmolLM2} do not, highlighting that scaling alone is insufficient for improving conformal factuality.}
  \label{fig:16prompting_scaling}
\end{figure}

Figure~\ref{fig:16prompting_scaling} reveals the heterogeneous scaling behaviors discussed in Section~\ref{sec:model_choice_scorers} in greater detail. While the \texttt{Llama-3.x} family shows monotonic improvement with model size across most metrics and factuality targets, this pattern does not hold for \texttt{Qwen3} or \texttt{SmolLM2}. For \texttt{Qwen3}, the smallest model (0.6B) sometimes outperforms larger variants, and for \texttt{SmolLM2}, there is no systematic ordering by parameter count. These results reinforce the main-paper conclusion that scaling alone does not guarantee improved conformal factuality.
 
\clearpage
\subsubsection{Comparison between Entailment-based and LLM-based Scoring Functions} \label{app:model_and_entialment}

Section \ref{sec:family_choice_scorers} compares model confidence score with entailment-based scoring functions on the FActScore dataset. Here, we extend the results to the MATH-1K dataset.

\begin{figure}[htbp]
  \centering
  \includegraphics[width=\linewidth]{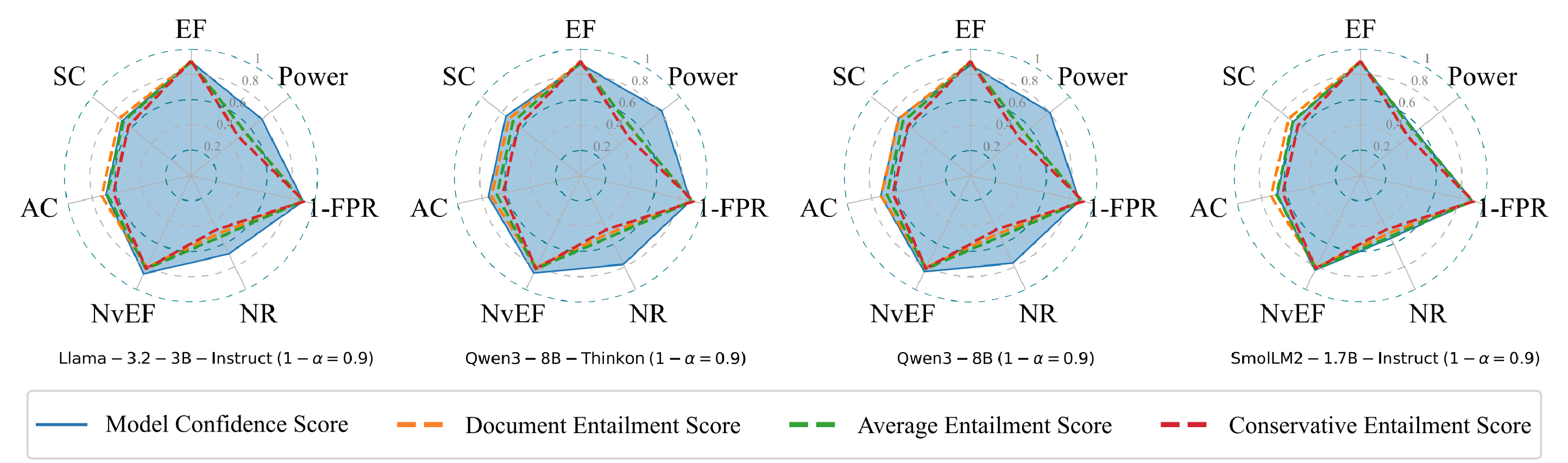}
  \caption{Comparison between entailment-based scores against the model confidence score on the MATH-1K dataset.}
  \label{fig:family_choice_scorers_math}
\end{figure}

As we can see in Figure \ref{fig:family_choice_scorers_math}, model confidence score yields a better power and non-empty rate. However, in terms of empirical factuality and non-vacuous empirical factuality, the gap between the different scoring functions is close. Moreover, with some models, the document entailment score has a better sufficient correctness and and accuracy comparing to model confidence score.

\clearpage
\subsubsection{Sufficient Correctness and Conditional Sufficient Correctness} \label{app:sc_and_csc}
 
Section~\ref{sec:sc_and_csc} introduces Conditional Sufficient Correctness (CSC) and reports results for the \texttt{Qwen3} family. Here, we first describe how SC and CSC is measured. Then, we extend this analysis to three additional model families---\texttt{Llama-3.x}, \texttt{SmolLM2}, and \texttt{gpt-oss}---to assess whether the observed patterns generalize.

\paragraph{Measuring (conditional) sufficient correctness}.
Let $y$ and $y'$ denote the original and filtered outputs for the same input, respectively. In this section, sufficient correctness measures whether the filtered output $y'$ contains enough correct information to recover the answer. To evaluate this, we use \texttt{gpt-5-nano} with the prompt in Appendix~\ref{sec:prompt_sc}, replacing the placeholder \texttt{\{response\}} with the string representation of $y'$. Let $n$ denote the number of examples in the dataset. We then define sufficient correctness as
$$
\mathrm{SC} = \frac{1}{n}\sum_{i=1}^n \mathbb{I}[y_i' \text{ is sufficient correct}].
$$

Conditional sufficient correctness is defined by restricting attention to examples for which the original output $y$ is already sufficient correct. Formally,
$$
\mathrm{CSC} = \frac{\sum_{i=1}^n \mathbb{I}[y_i' \text{ is sufficient correct}]}{\sum_{i=1}^n \mathbb{I}[y_i \text{ is sufficient correct}]}.
$$
 
\begin{figure}[htbp]
  \centering
  \includegraphics[width=\linewidth]{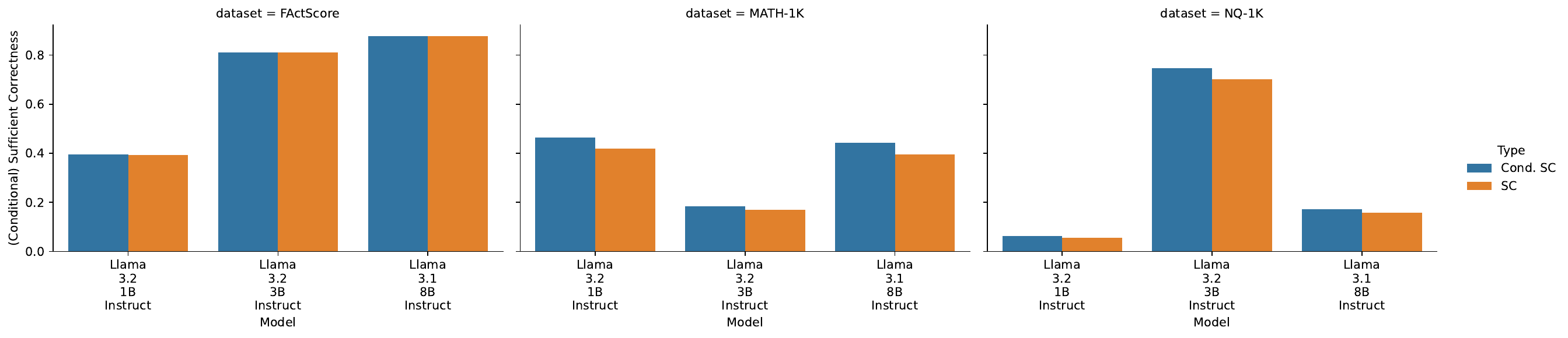}
  \caption{Sufficient Correctness (SC) and Conditional Sufficient Correctness (CSC) for the \texttt{Llama-3.x} family at $\alpha=0.05$ across FActScore, MATH-1K, and NQ-1K. CSC consistently exceeds SC, indicating that filtering largely preserves useful content when the generator provides it.}
  \label{fig:llama3_csc}
\end{figure}

\paragraph{Llama-3.x family.}
Figure~\ref{fig:llama3_csc} shows SC and CSC for \texttt{Llama-3.x} at $\alpha = 0.05$. As with \texttt{Qwen3}, CSC consistently exceeds SC, confirming that a substantial portion of SC failures are attributable to the generator rather than the filter. The gap is especially wide on MATH-1K, where the smaller models struggle to produce sufficiently correct unfiltered outputs.
 
\begin{figure}[htbp]
  \centering
  \includegraphics[width=\linewidth]{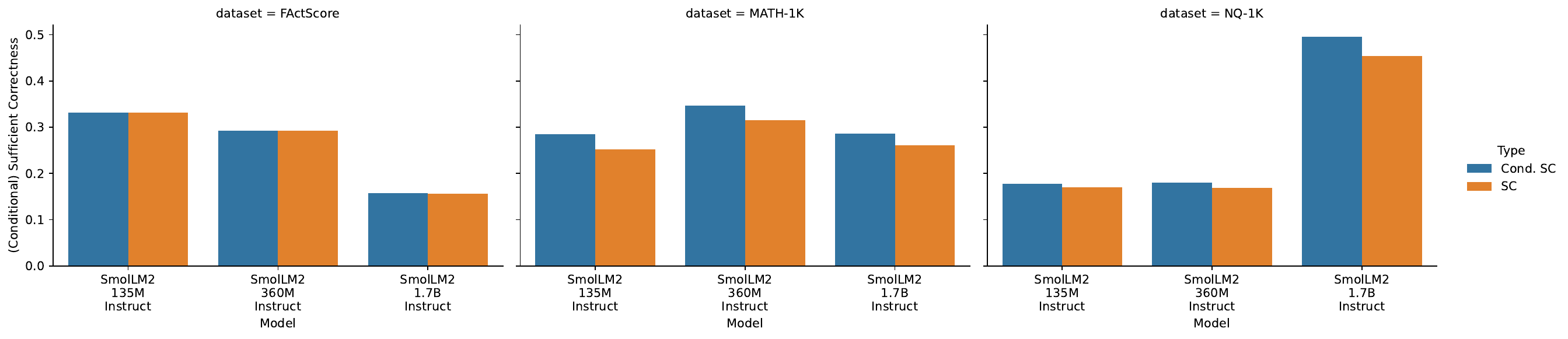}
  \caption{Sufficient Correctness (SC) and Conditional Sufficient Correctness (CSC) for the \texttt{SmolLM2} family at $\alpha=0.05$ across FActScore, MATH-1K, and NQ-1K. No consistent scaling benefit is observed within this family.}
  \label{fig:smollm2_csc}
\end{figure}

\paragraph{SmolLM2 family.}
Figure~\ref{fig:smollm2_csc} presents the same comparison for \texttt{SmolLM2}. The absolute SC values are lower due to the smaller model sizes, but the CSC--SC gap is qualitatively similar. Notably, there is no consistent improvement in SC or CSC with model size within this family, echoing the main-paper finding that scaling the scorer does not reliably improve conformal factuality.
 
\begin{figure}[htbp]
  \centering
  \includegraphics[width=\linewidth]{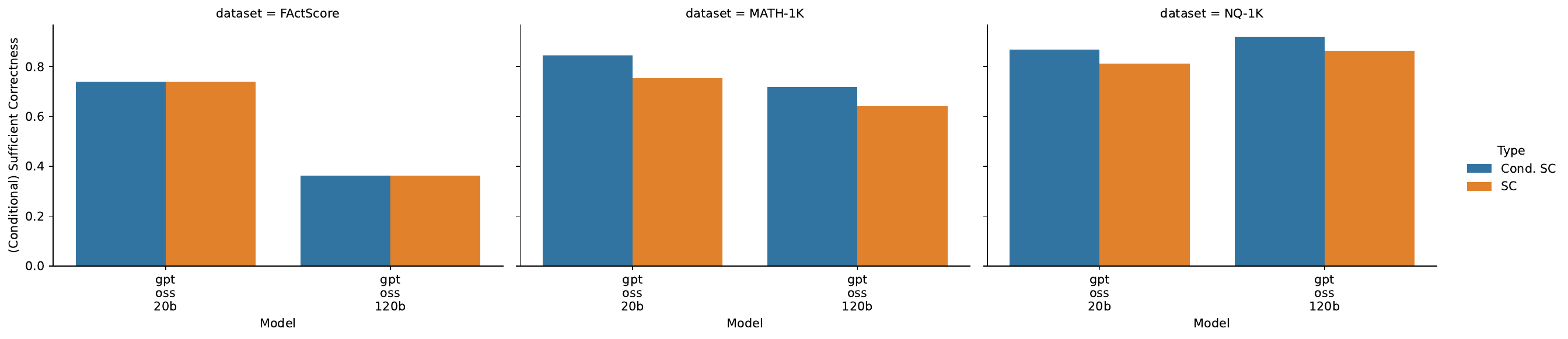}
  \caption{Sufficient Correctness (SC) and Conditional Sufficient Correctness (CSC) for the \texttt{gpt-oss} family at $\alpha=0.05$ across FActScore, MATH-1K, and NQ-1K. The larger \texttt{gpt-oss-120b} does not consistently outperform \texttt{gpt-oss-20b}.}
  \label{fig:gptoss_csc}
\end{figure}

\paragraph{gpt-oss family.}
Figure~\ref{fig:gptoss_csc} compares the two \texttt{gpt-oss} models (20B and 120B). Despite a roughly $6\times$ difference in total parameters, the larger model does not consistently outperform the smaller one in SC or CSC, further supporting the conclusion that parameter count is not the primary driver of conformal factuality performance.
 
\clearpage
\subsubsection{Robustness to Calibration Distribution Shift} \label{app:distribution_shift}
 
Section~\ref{sec:distribution_shift} studies the sensitivity of conformal factuality guarantees to distribution shift between calibration and test data, using \texttt{Qwen3-4B} as the scorer. Here we extend this analysis to \texttt{Llama-3.2-3B-Instruct} and \texttt{SmolLM2-360M-Instruct} to assess whether the robustness (or lack thereof) of different scoring functions is consistent across model families.
 
\begin{figure}[htbp]
  \centering
  \includegraphics[width=\linewidth]{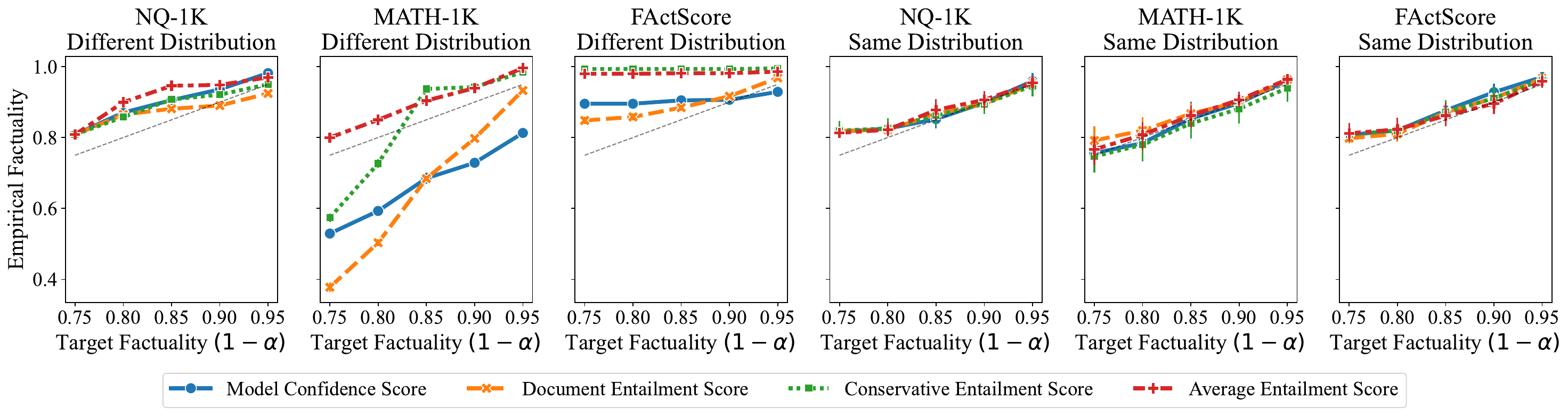}
  \caption{Empirical factuality (EF) on FActScore, MATH-1K, and NQ-1K under same-distribution and different-distribution calibration, using \texttt{Llama-3.2-3B-Instruct} as the scorer. Distribution shift causes the factuality guarantee to break for several scoring functions.}
  \label{fig:distribution_shift_llama3}
\end{figure}

\paragraph{Llama-3.2-3B-Instruct.}
Figure~\ref{fig:distribution_shift_llama3} shows empirical factuality under same-distribution and different-distribution calibration for \texttt{Llama-3.2-3B-Instruct}. When calibration data come from a different distribution (left panels), the factuality guarantee frequently fails across all three datasets. Interestingly, the entailment-based scorers, which appeared robust to distribution shift under \texttt{Qwen3-4B}, show degraded performance here, indicating that robustness to distribution shift is model-dependent rather than a universal property of any particular scoring function family.
 
\begin{figure}[htbp]
  \centering
  \includegraphics[width=\linewidth]{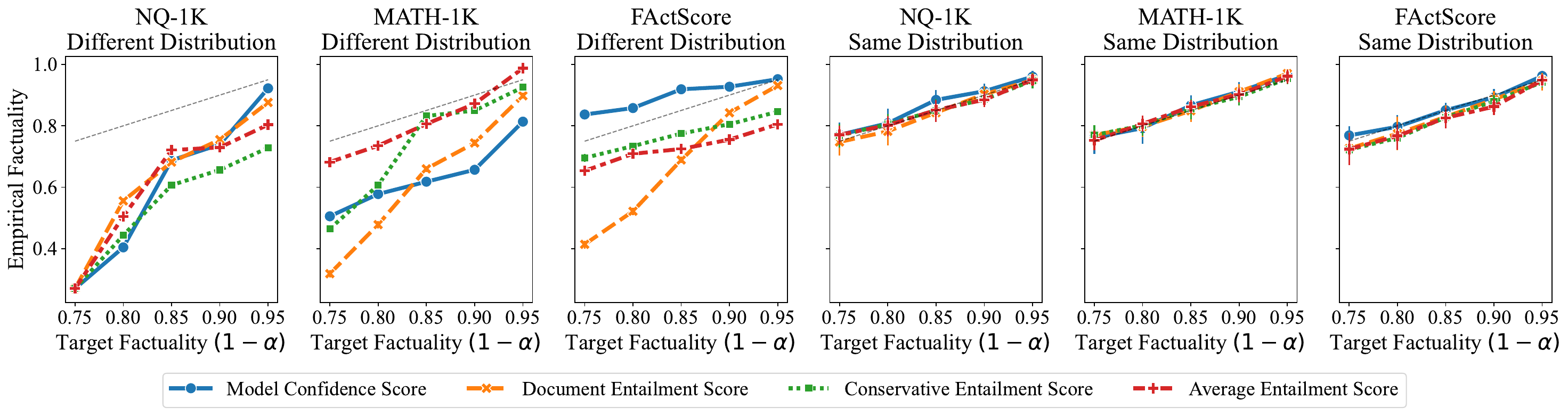}
  \caption{Empirical factuality (EF) on FActScore, MATH-1K, and NQ-1K under same-distribution and different-distribution calibration, using \texttt{SmolLM2-360M-Instruct} as the scorer. The results confirm that distribution shift in the calibration set undermines conformal factuality guarantees across model families.}
  \label{fig:distribution_shift_smollm2}
\end{figure}

\paragraph{SmolLM2-360M-Instruct.}
Figure~\ref{fig:distribution_shift_smollm2} presents the same analysis for \texttt{SmolLM2-360M-Instruct}. The pattern is consistent: different-distribution calibration leads to violations of the target factuality level, particularly on NQ-1K and MATH-1K. Together with the \texttt{Qwen3-4B} and \texttt{Llama-3.2-3B} results, these findings underscore that calibration data must be collected using the same generator and scorer that will be deployed, as even switching the underlying LLM for the same dataset can break the exchangeability assumption.
 
\clearpage
\subsubsection{Robustness to Adversarial Distractors} \label{app:adversarial_distractors}
 
Section~\ref{sec:distractors} examines how injecting adversarial distractor claims into the test set degrades empirical factuality, using \texttt{Qwen3-4B} on FActScore as the primary example. Here we extend this analysis to additional scorers (\texttt{Llama-3.2-3B-Instruct} and \texttt{SmolLM2-1.7B-Instruct}) and datasets (MATH-1K and NQ-1K) to assess the generality of this vulnerability.
 
\begin{figure}[htbp]
  \centering
  \includegraphics[width=\linewidth]{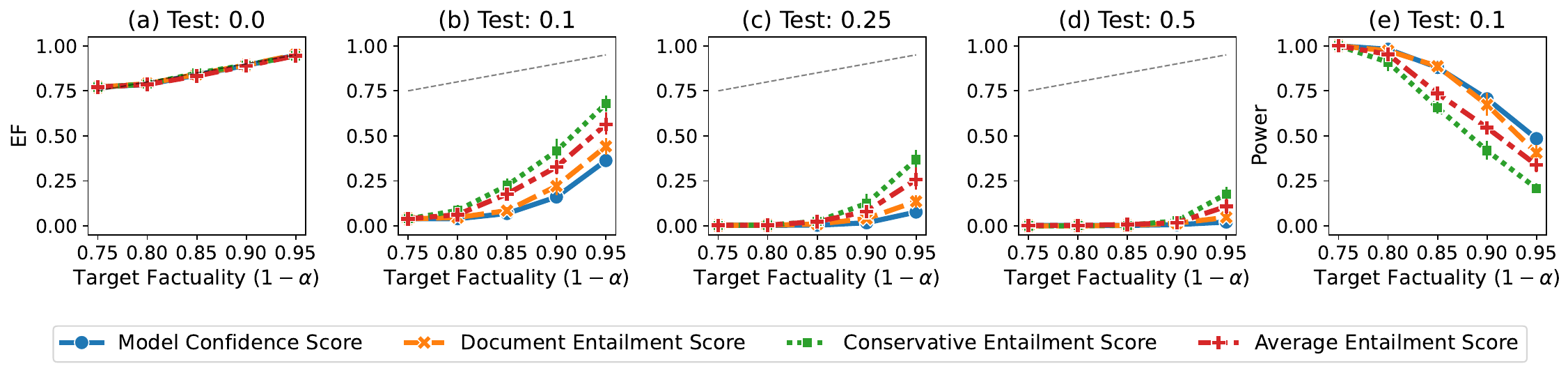}
  \caption{Empirical factuality under varying test distractor proportions ($0.0$ to $0.5$) with fixed clean calibration data on FActScore, using \texttt{Llama-3.2-3B-Instruct}. Factuality degrades sharply as distractor rate increases.}
  \label{fig:factscore_llama}
\end{figure}

\begin{figure}[htbp]
  \centering
  \includegraphics[width=\linewidth]{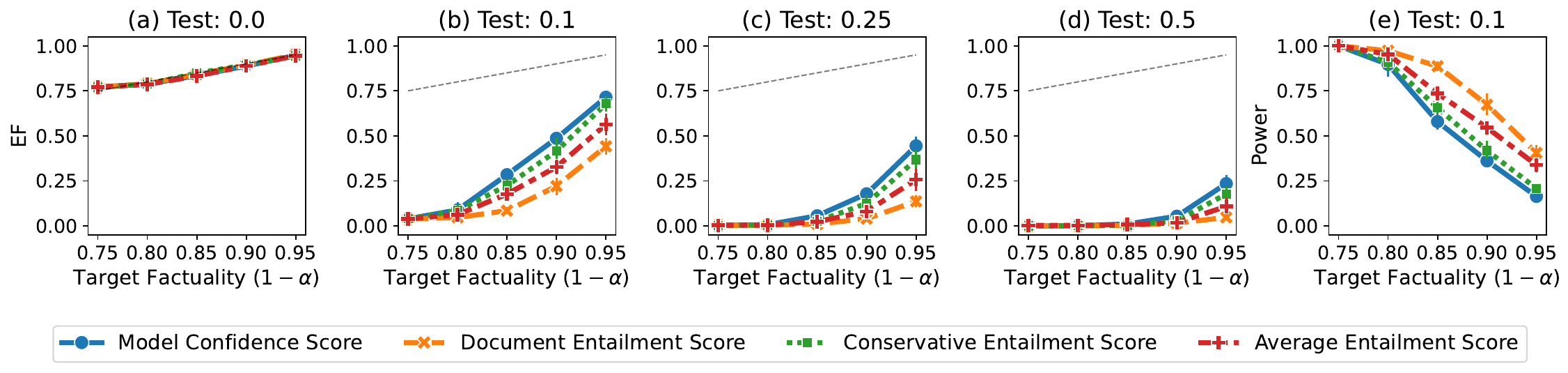}
  \caption{Empirical factuality under varying test distractor proportions ($0.0$ to $0.5$) with fixed clean calibration data on FActScore, using \texttt{SmolLM2-1.7B-Instruct}. The degradation pattern is consistent with other model families.}
  \label{fig:factscore_smol}
\end{figure}

\paragraph{FActScore dataset.}
Figures~\ref{fig:factscore_llama} and~\ref{fig:factscore_smol} show that both \texttt{Llama-3.2-3B-Instruct} and \texttt{SmolLM2-1.7B-Instruct} exhibit the same sharp degradation in empirical factuality as the distractor rate increases, consistent with the \texttt{Qwen3-4B} results. This confirms that the vulnerability to adversarial distractors is not model-specific but a systemic issue with the current conformal filtering framework.
 
\begin{figure}[htbp]
  \centering
  \includegraphics[width=\linewidth]{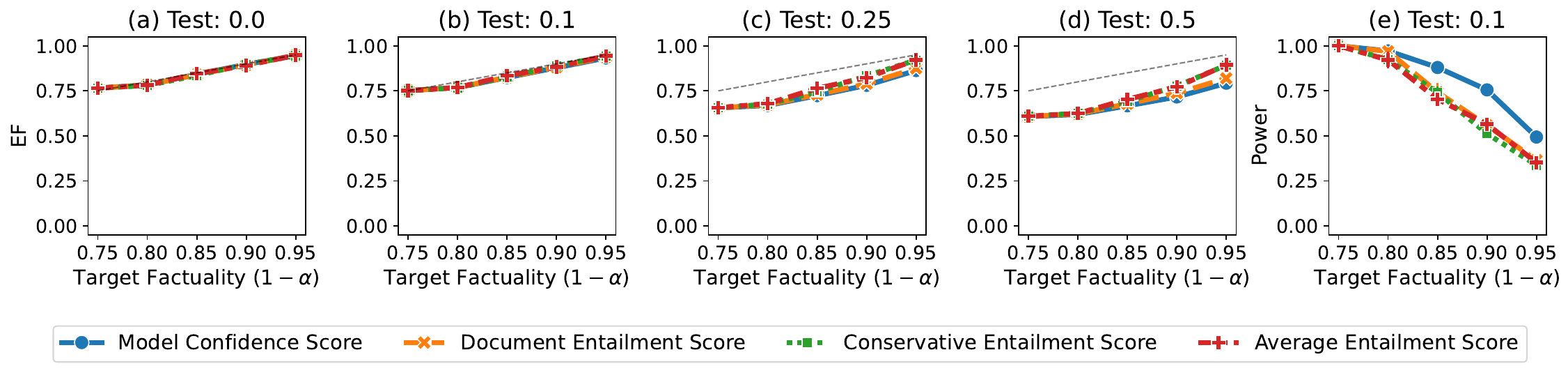}
  \caption{Empirical factuality under varying test distractor proportions on MATH-1K with \texttt{Qwen3-4B}. Mathematical claims are especially vulnerable to subtle numerical perturbations.}
  \label{fig:math_qwen}
\end{figure}

\begin{figure}[htbp]
  \centering
  \includegraphics[width=\linewidth]{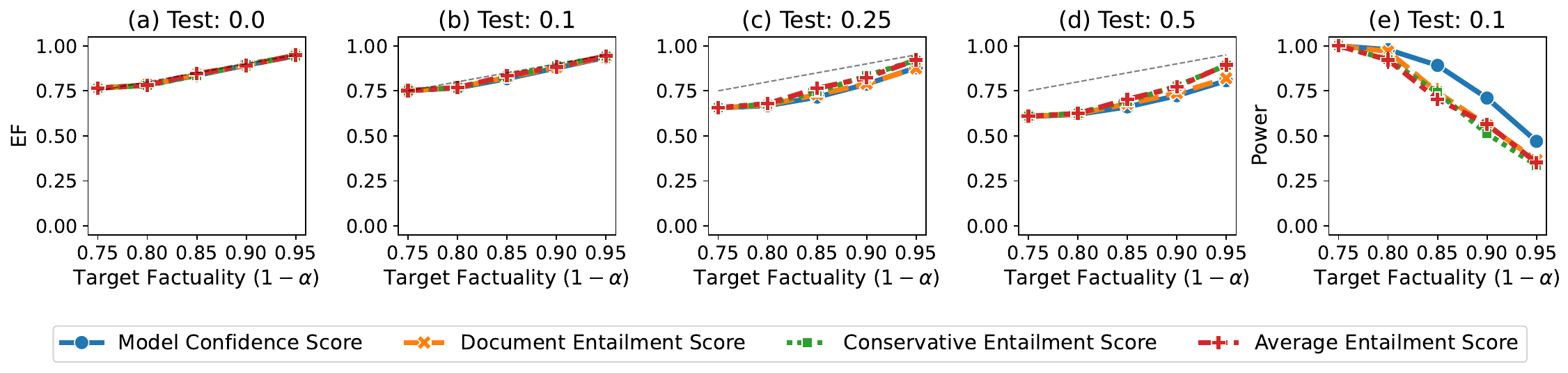}
  \caption{Empirical factuality under varying test distractor proportions on MATH-1K with \texttt{Llama-3.2-3B-Instruct}.}
  \label{fig:math_llama}
\end{figure}

\begin{figure}[htbp]
  \centering
  \includegraphics[width=\linewidth]{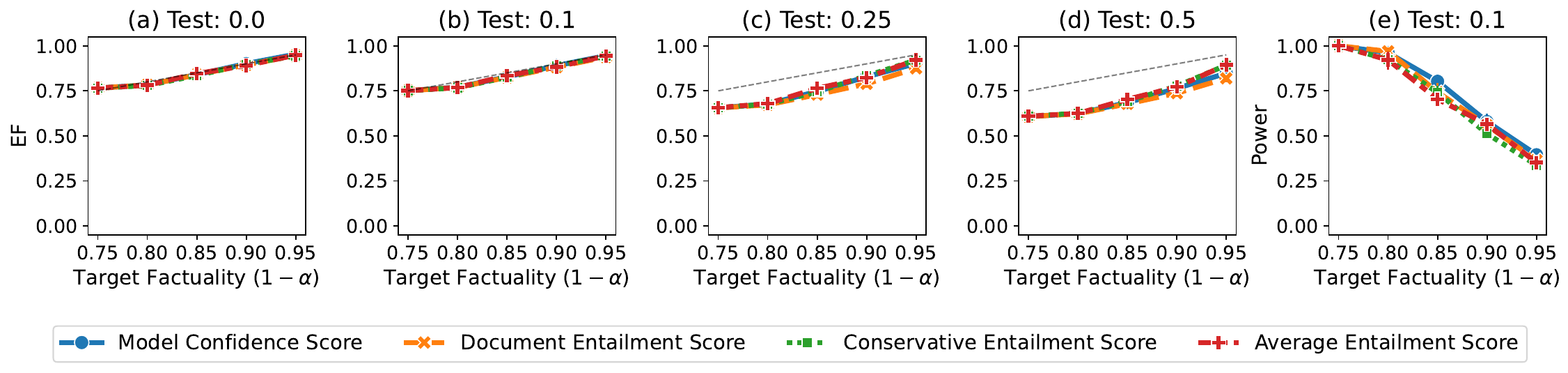}
  \caption{Empirical factuality under varying test distractor proportions on MATH-1K with \texttt{SmolLM2-1.7B-Instruct}.}
  \label{fig:math_smol}
\end{figure}

\clearpage
\paragraph{MATH-1K dataset.}
Figures~\ref{fig:math_qwen}--\ref{fig:math_smol} extend the distractor robustness analysis to MATH-1K. Mathematical claims are particularly susceptible to subtle numerical perturbations, and indeed we observe that empirical factuality drops rapidly even at low distractor rates (e.g., 10\%). This suggests that distractor robustness is an especially critical concern for reasoning-intensive tasks.
 
\begin{figure}[htbp]
  \centering
  \includegraphics[width=\linewidth]{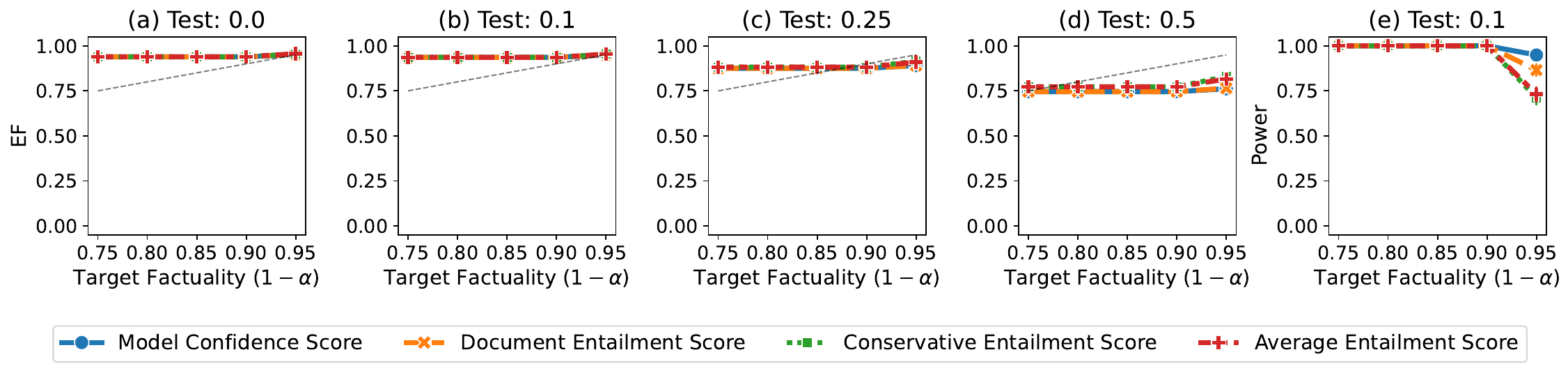}
  \caption{Empirical factuality under varying test distractor proportions on NQ-1K with \texttt{Qwen3-4B}.}
  \label{fig:nq_qwen}
\end{figure}

\begin{figure}[htbp]
  \centering
  \includegraphics[width=\linewidth]{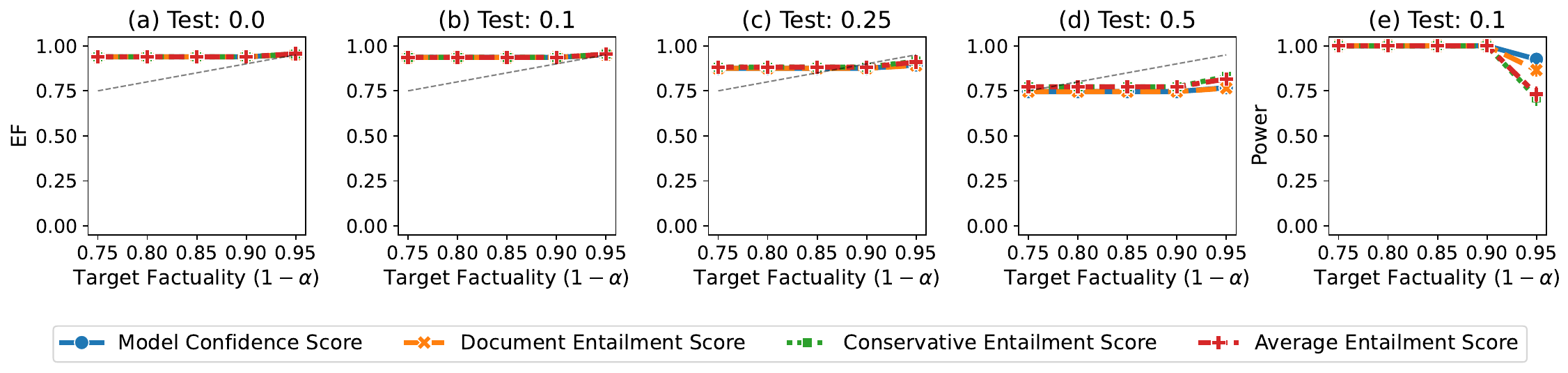}
  \caption{Empirical factuality under varying test distractor proportions on NQ-1K with \texttt{Llama-3.2-3B-Instruct}.}
  \label{fig:nq_llama}
\end{figure}

\begin{figure}[htbp]
  \centering
  \includegraphics[width=\linewidth]{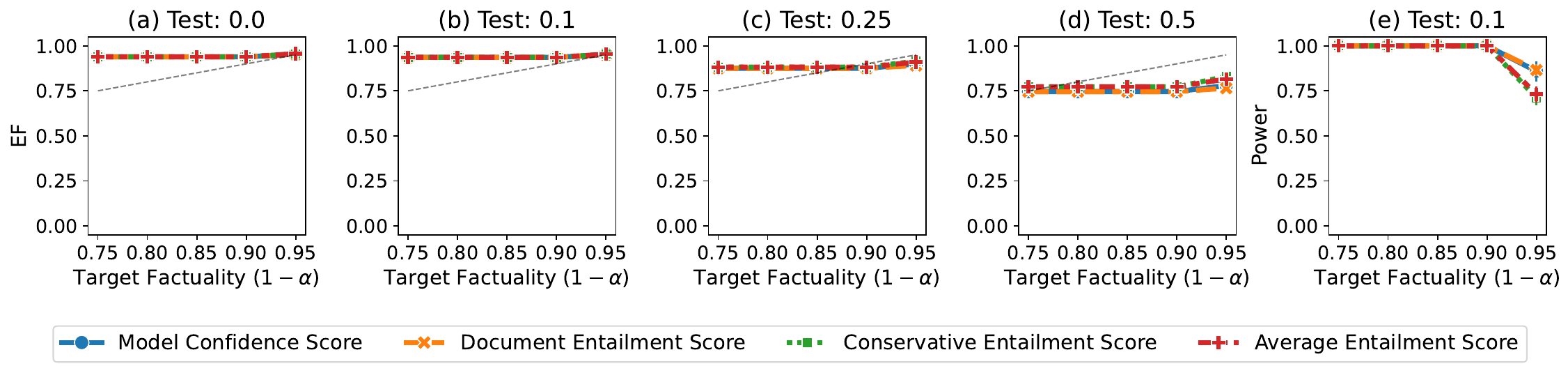}
  \caption{Empirical factuality under varying test distractor proportions on NQ-1K with \texttt{SmolLM2-1.7B-Instruct}.}
  \label{fig:nq_smol}
\end{figure}

\clearpage
\paragraph{NQ-1K dataset.}
Figures~\ref{fig:nq_qwen}--\ref{fig:nq_smol} complete the analysis on NQ-1K. The overall pattern is consistent across all three datasets: as distractors are injected, empirical factuality drops below the target level, and recovering factuality by raising the target threshold comes at a steep cost in power. This reinforces the main-paper conclusion that developing scoring functions robust to adversarial content is a critical direction for future work.
 
\clearpage
\subsubsection{Robustness with Adversarial Prepared Calibration} \label{app:adversarial_calibration}
 
Section~\ref{sec:distraction_aware} investigates whether including distractors in the calibration set can restore factuality guarantees when distractors are also present at test time. The main paper presents results for \texttt{Qwen3-4B} on FActScore. Here we provide the complete set of results across all three datasets and three scorer models.
 
The key question addressed by these experiments is: \emph{If we anticipate adversarial distractors during deployment, can we protect the factuality guarantee by injecting synthetic distractors into the calibration set?} The answer, as the following figures show, is nuanced: matching the distractor proportion in calibration to that in the test set does restore empirical factuality to the target level, but at a severe cost to the non-empty rate. This trade-off arises because the conformal threshold becomes more stringent when calibration data contain distractors, causing the filter to aggressively remove content---including correct claims.
 
\begin{figure}[htbp]
  \centering
  \includegraphics[width=\linewidth]{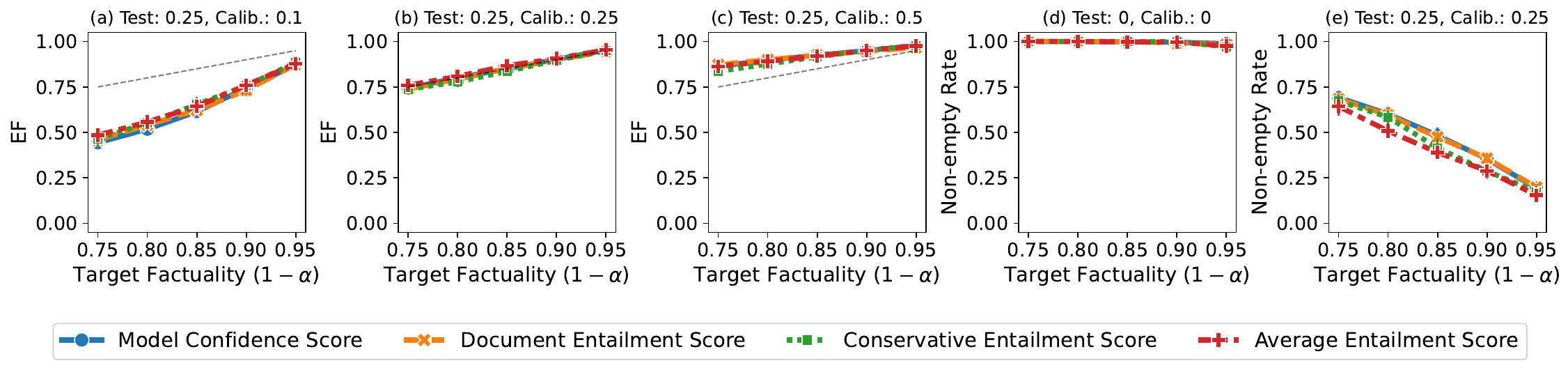}
  \caption{Empirical factuality and non-empty rates under varying calibration distractor proportions on FActScore with \texttt{Llama-3.2-3B-Instruct}. Matching calibration to test distractor levels restores factuality but reduces the non-empty rate.}
  \label{fig:r3_factscore_llama}
\end{figure}

\begin{figure}[htbp]
  \centering
  \includegraphics[width=\linewidth]{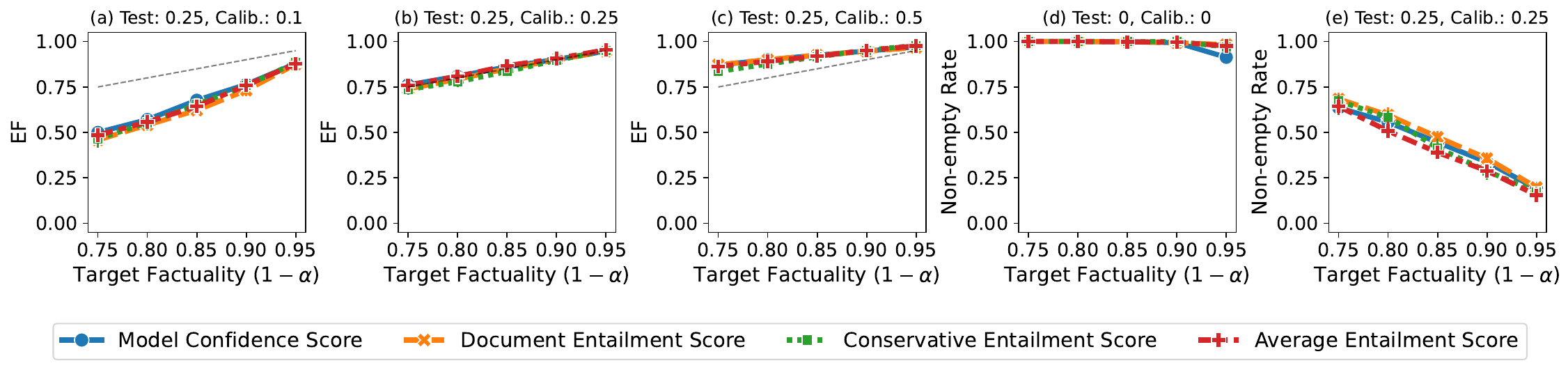}
  \caption{Empirical factuality and non-empty rates under varying calibration distractor proportions on FActScore with \texttt{SmolLM2-1.7B-Instruct}.}
  \label{fig:r3_factscore_smol}
\end{figure}

\paragraph{FActScore dataset.}
Figures~\ref{fig:r3_factscore_llama} and~\ref{fig:r3_factscore_smol} show the factuality--coverage trade-off for \texttt{Llama-3.2-3B-Instruct} and \texttt{SmolLM2-1.7B-Instruct} on FActScore. When calibration distractor proportion is underestimated (panels~a), empirical factuality remains below the target. Matching the proportion (panels~b) restores the guarantee, and overestimating it (panels~c) yields conservative but highly restrictive filtering. Panels~(d)--(e) illustrate the non-empty rate cost.
 
\begin{figure}[htbp]
  \centering
  \includegraphics[width=\linewidth]{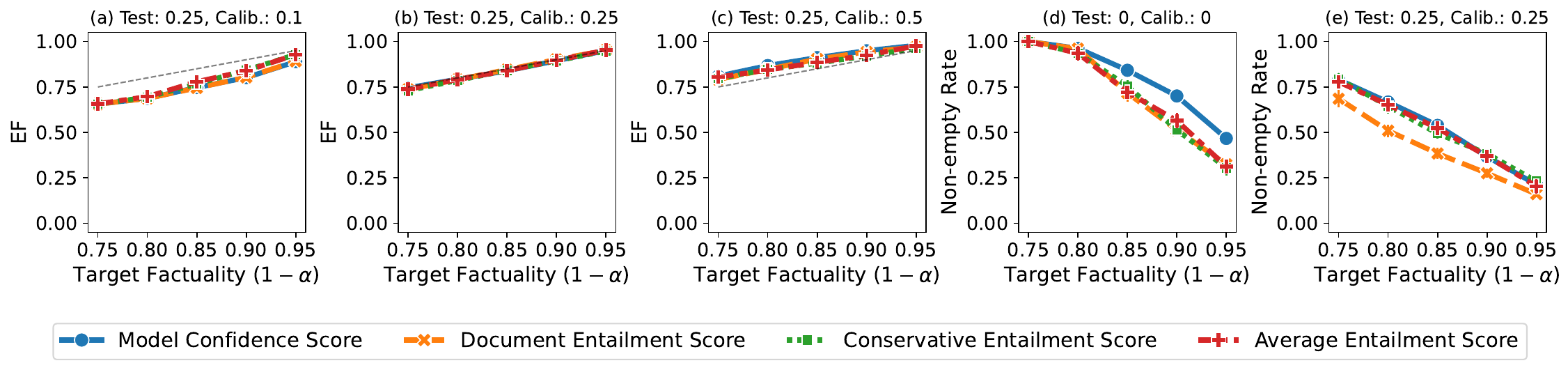}
  \caption{Empirical factuality and non-empty rates under varying calibration distractor proportions on MATH-1K with \texttt{Qwen3-4B}.}
  \label{fig:r3_math_qwen}
\end{figure}

\begin{figure}[htbp]
  \centering
  \includegraphics[width=\linewidth]{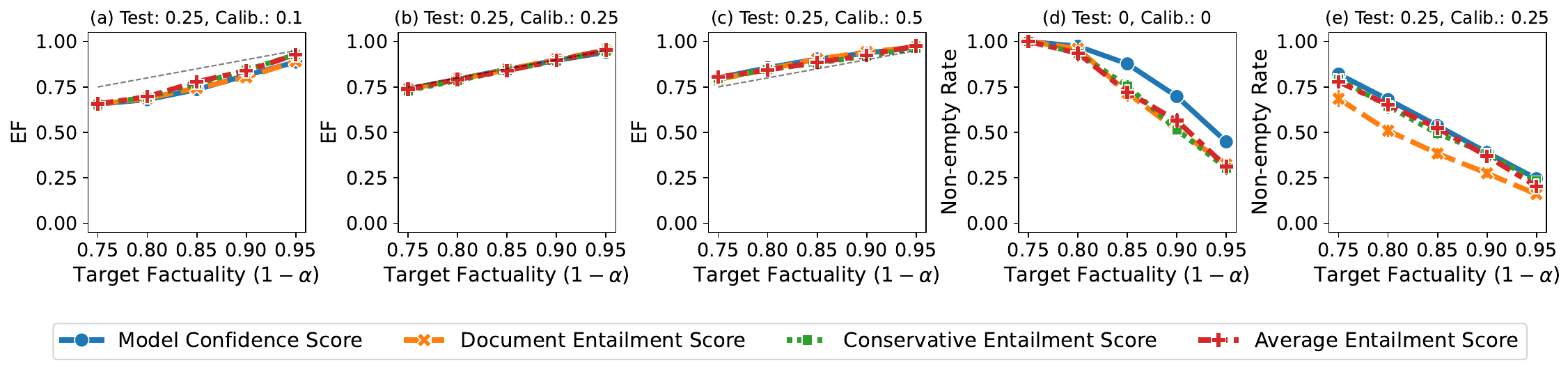}
  \caption{Empirical factuality and non-empty rates under varying calibration distractor proportions on MATH-1K with \texttt{Llama-3.2-3B-Instruct}.}
  \label{fig:r3_math_llama}
\end{figure}

\begin{figure}[htbp]
  \centering
  \includegraphics[width=\linewidth]{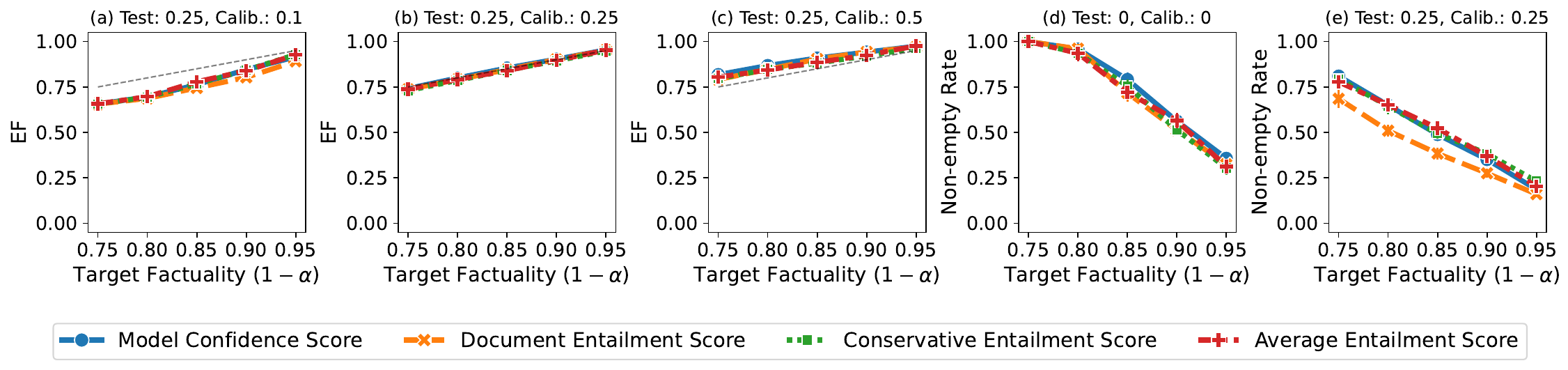}
  \caption{Empirical factuality and non-empty rates under varying calibration distractor proportions on MATH-1K with \texttt{SmolLM2-1.7B-Instruct}.}
  \label{fig:r3_math_smol}
\end{figure}

\clearpage
\paragraph{MATH-1K dataset.}
Figures~\ref{fig:r3_math_qwen}--\ref{fig:r3_math_smol} present the same analysis on MATH-1K. The trade-off is even starker here: the non-empty rate drops precipitously when distractors are introduced into calibration, reflecting the difficulty of distinguishing correct mathematical steps from plausible but incorrect ones.
 
\begin{figure}[htbp]
  \centering
  \includegraphics[width=\linewidth]{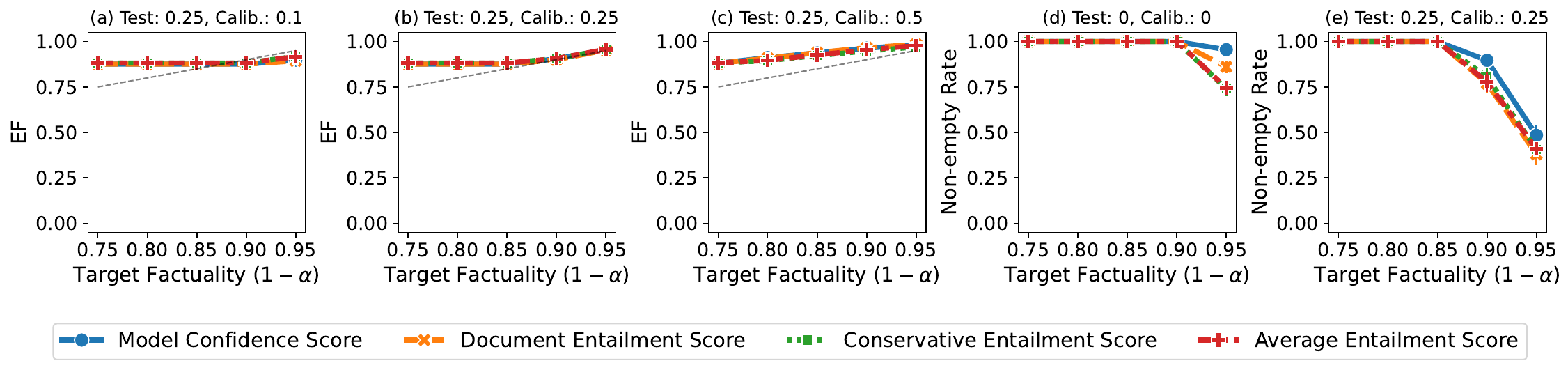}
  \caption{Empirical factuality and non-empty rates under varying calibration distractor proportions on NQ-1K with \texttt{Qwen3-4B}.}
  \label{fig:r3_nq_qwen}
\end{figure}

\begin{figure}[htbp]
  \centering
  \includegraphics[width=\linewidth]{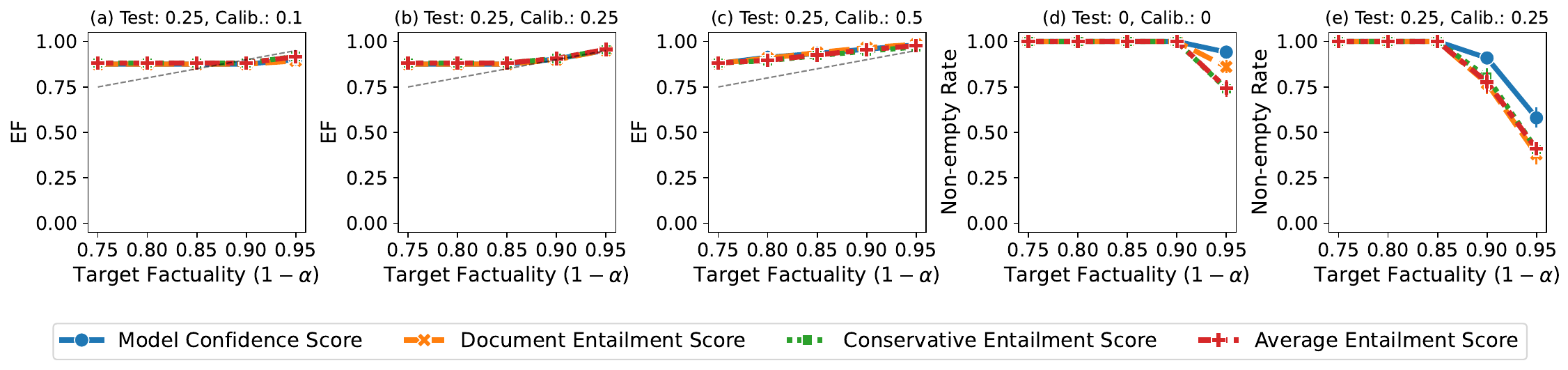}
  \caption{Empirical factuality and non-empty rates under varying calibration distractor proportions on NQ-1K with \texttt{Llama-3.2-3B-Instruct}.}
  \label{fig:r3_nq_llama}
\end{figure}

\begin{figure}[htbp]
  \centering
  \includegraphics[width=\linewidth]{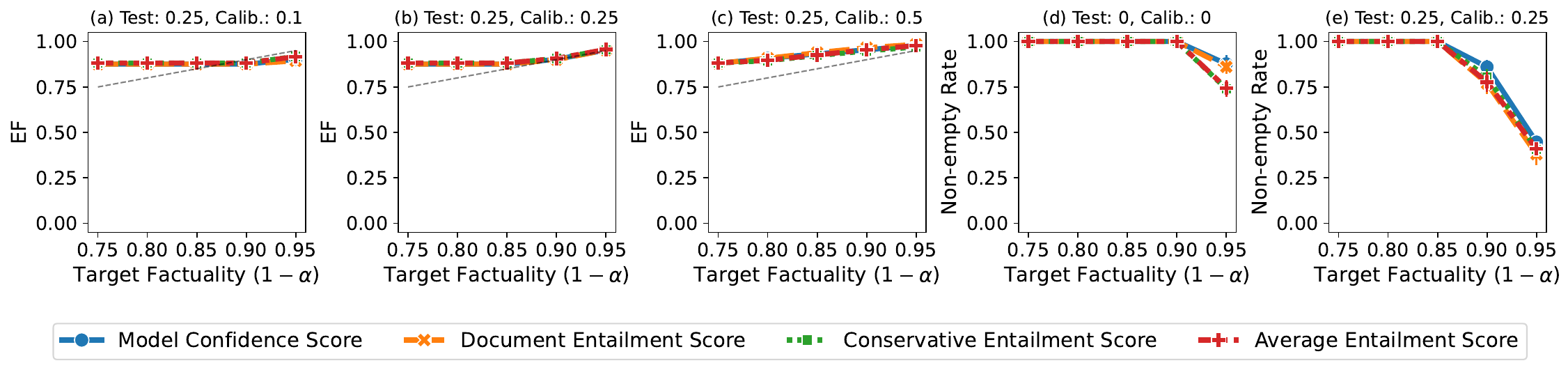}
  \caption{Empirical factuality and non-empty rates under varying calibration distractor proportions on NQ-1K with \texttt{SmolLM2-1.7B-Instruct}.}
  \label{fig:r3_nq_smol}
\end{figure}

\clearpage
\paragraph{NQ-1K dataset.}
Figures~\ref{fig:r3_nq_qwen}--\ref{fig:r3_nq_smol} complete the analysis on NQ-1K. The overall pattern is consistent: distraction-aware calibration can restore the statistical guarantee, but the resulting outputs are frequently empty, highlighting a fundamental limitation of threshold-based filtering when scoring functions cannot reliably distinguish correct claims from adversarial ones.
 
\clearpage
\subsection{Distractor Generation Protocol} \label{app:distractors}
 
A core assumption of the conformal factuality framework is that the test data are exchangeable with the calibration dataset. In practice, however, this assumption may be mildly violated---for example, when LLM outputs contain hallucinated claims that were not represented in calibration. To stress-test the robustness of scoring functions under such conditions (Section~\ref{sec:distractors}), we construct adversarial distractor claims as follows.
 
For each query $x_i$, the corresponding reference text $R(x)$, and the set of parsed claims $\{c_i\}$, we prompt the LLM to modify each $c_i$ (conditioned on $x$ and $R(x)$) so that the result resembles a plausible hallucination. The full prompt used for this generation step is provided in Appendix~\ref{generating-hallucination-claims}.
 
Crucially, we want these distractor claims to be sufficiently convincing that the model itself would believe it generated them. To enforce this, we apply a verification step: after generating a hallucinated claim, we prompt the LLM to judge whether the claim could plausibly be one of its own outputs given $x$ and $R(x)$ (prompt in Appendix~\ref{hallucination-verification}). If the model identifies the claim as a plausible self-generated hallucination, we retain it; otherwise, we regenerate until a sufficiently convincing distractor is found. This two-stage process ensures that the adversarial distractors are realistic and relevant to the failure modes of modern LLMs.
 
\clearpage
\subsection{Human Evaluation} \label{app:human_evaluation}
 
To validate the use of \texttt{gpt-5-nano} as a factuality judge (Section~\ref{sec:models}), we conducted a human evaluation study. We randomly sampled 200 claims from the FActScore dataset and had two graduate students (referred to as A and B) independently label the factuality of each claim. We also obtained labels from \texttt{gpt-5-nano} on the same set.
 
Table~\ref{tab:human} reports the pairwise agreement rates. The agreement between \texttt{gpt-5-nano} and each human annotator (76.5\% and 77.0\%) is comparable to---and slightly exceeds---the inter-annotator agreement between the two humans (73.0\%). This result supports the use of \texttt{gpt-5-nano} as a factuality judge in our experimental pipeline, as it performs on par with individual human annotators.
 
\begin{table}[htbp]
  \centering
  \begin{tabular}{@{}lll@{}}
    \toprule
    Pair               & Labelers      & Agreement Rate \\ \midrule
    Model--Human       & \texttt{gpt-5-nano} vs.\ Student A & 76.5\% \\
    Model--Human       & \texttt{gpt-5-nano} vs.\ Student B & 77.0\% \\
    Human--Human       & Student A vs.\ Student B            & 73.0\% \\
    \bottomrule
  \end{tabular}
  \caption{Agreement rates on factuality labels between \texttt{gpt-5-nano} and two human annotators, as well as between the two human annotators themselves. The model--human agreement is comparable to inter-human agreement.}
  \label{tab:human}
\end{table}
 
\clearpage
\subsection{Prompts} \label{app:prompts}
 
This section collects all prompts used throughout our experiments. We organize them by their role in the pipeline: generation (Sections~\ref{sec:prompt_gen_ref} and~\ref{sec:prompt_gen_noref}), claim parsing (Section~\ref{sec:prompt_parser}), factuality labeling (Sections~\ref{sec:prompt_labeler_nogt} and~\ref{sec:prompt_labeler_gt}), adversarial distractor generation and verification (Sections~\ref{generating-hallucination-claims} and~\ref{hallucination-verification}), model confidence scoring (Section~\ref{modelconfscore_prompt}), claim merging (Sections~\ref{sec:prompt_merger_factscore}--\ref{sec:prompt_merger_math}), correctness evaluation (Section~\ref{sec:prompt_correctness}), sufficient correctness evaluation (Section~\ref{sec:prompt_sc}), and MATH reference generation (Section~\ref{math_reference}). All prompts instruct the model to return structured JSON5 output to facilitate automated parsing.

\subsubsection{Generator (with reference)} \label{sec:prompt_gen_ref}

\begin{promptbox}
\begin{Verbatim}[breaklines=true, breakanywhere=true]
  You are a helpful assistant that answers queries strictly based on the provided reference text.

  Instructions:
  1. You will be given:
  - A reference text
  - A query
  2. Use only the information from the reference text to answer the query.
  3. Do not include any information not supported by the reference text.

  Output Requirements:
  - Output ONLY a single VALID JSON5 object with EXACTLY this schema:
  {
    "response": "...answer strictly based on the reference text..."
  }

  JSON5 Rules:
  - Use DOUBLE QUOTES (") for all keys and all string values.
  - Escape double quotes inside string values as \".
  - Escape backslashes as \\.
  - No trailing commas in objects or arrays.
  - Use the exact top-level container specified and close it properly.
  - Do not include comments, code fences, or any text outside the JSON5 output.
  - Follow the schema exactly; do not add or omit keys.

  Do NOT include:
  - Any text, explanations, comments, or formatting outside of the JSON5.
  - Any code block delimiters (e.g., ```json).

  Input:
  Reference Text: {reference}
  Query: {query}

  (Reiteration of the instruction)
  Answer the query strictly using only the reference text, and return a single JSON5 object with the key "response" only.

  Output:
\end{Verbatim}
\end{promptbox}

\subsubsection{Generator (without reference)} \label{sec:prompt_gen_noref}

\begin{promptbox}
\begin{Verbatim}[breaklines=true, breakanywhere=true]
  You are a helpful assistant that answers queries.

  Instructions:
  1. You will be given:
  - A query

  Output Requirements:
  - Output ONLY a single VALID JSON5 object with EXACTLY this schema:
  {
    "response": "...answer..."
  }

  JSON5 Rules:
  - Use DOUBLE QUOTES (") for all keys and all string values.
  - Escape double quotes inside string values as \".
  - Escape backslashes as \\.
  - No trailing commas in objects or arrays.
  - Use the exact top-level container specified and close it properly.
  - Do not include comments, code fences, or any text outside the JSON5 output.
  - Follow the schema exactly; do not add or omit keys.

  Do NOT include:
  - Any text, explanations, comments, or formatting outside of the JSON5.
  - Any code block delimiters (e.g., ```json).

  Input:
  Query: {query}

  (Reiteration of the instruction)
  Answer the query and return a single JSON5 object with the key "response" only.

  Output:
\end{Verbatim}
\end{promptbox}

\subsubsection{Parser} \label{sec:prompt_parser}

\begin{promptbox}
\begin{Verbatim}[breaklines=true, breakanywhere=true]
  You are an AI assistant tasked with breaking down input text into small, self-contained claims for easy human verification.

  Instructions:
  1. Parse the provided text into concise, independent, and non-overlapping subclaims.
  2. Ensure each subclaim is:
  - As small and specific as possible.
  - Independent and self-contained.
  - Do not use pronouns like he, she, his, her, it, its, etc.
  - Explicitly mention subjects.
  - Factually complete without relying on context from other subclaims.
  3. If the provided text is not a full sentence, use the provided text verbatim as the subclaim

  Output Requirements:
  1. The result must be a VALID and COMPLETE JSON list of dictionaries.
  2. Each dictionary must have the following structure:
  {
    "subclaim": "Subclaim text"
  }

  JSON Rules:
  - Ensure the JSON is STRICTLY VALID:
  - Use DOUBLE QUOTES ("") for all keys and string values.
  - DO NOT include trailing commas after the LAST item in arrays or objects.
  - Ensure ALL dictionaries are enclosed in curly braces {}.
  - Ensure the JSON list is ENCLOSED in square brackets [].
- CLOSE the JSON list properly with a closing square bracket ].
- DO NOT include any code block delimiters.
- DO NOT include any additional text, explanations, or comments in the output—return ONLY the JSON list.
Critical Requirement:
- Validate and ensure the JSON is complete, properly CLOSED, and parseable. The final JSON MUST pass validation without errors.

(Reiteration of the instruction)
1. Parse the provided text into concise, independent, and non-overlapping subclaims.
2. Ensure each subclaim is:
- As small and specific as possible.
- Independent and self-contained.
- Do not use pronouns like he, she, his, her, it, its, etc.
- Explicitly mention subjects.
- Factually complete without relying on context from other subclaims.
3. If the provided text is not a full sentence, use the provided text verbatim as the subclaim

Input:
{input}

Output:
\end{Verbatim}
\end{promptbox}

\subsubsection{Labeler (without ground-truth answer)} \label{sec:prompt_labeler_nogt}

\begin{promptbox}
\begin{Verbatim}[breaklines=true, breakanywhere=true]
You are an AI assistant tasked with assigning a label to a claim based on its factuality.

Instructions:
1. You are given:
- A reference text for the query.
- A query.
- A claim made in response to the query.
2. Rate the factuality of the claim as a boolean:
- true -> The claim is well-supported by the reference text.
- false -> The claim contradicts the reference text, or is not well-supported by it.
- If it is ambiguous, answer False.
3. Provide:
- The part(s) of the reference text that directly support your decision.
- A reasoning statement describing your rationale.
4. You must assign either true or false. Never return null or None.

Output Requirements:
- Output ONLY a single VALID JSON5 object with EXACTLY these keys:
{
  "highlighted_text": "Part(s) of the reference text that support the decision.",
  "reasoning": "A reasoning statement describing your rationale.",
  "answer": true
}
- "answer" must be a boolean (true/false).

JSON5 Rules:
- Use DOUBLE QUOTES (") for all keys and all string values.
- Escape double quotes inside string values as \".
- Escape backslashes as \\.
- No trailing commas in objects or arrays.
- Use the exact top-level container specified and close it properly.
- Do not include comments, code fences, or any text outside the JSON5 output.
- Follow the schema exactly; do not add or omit keys.

Do NOT include:
- Any text, explanations, comments, or formatting outside of the JSON5.
- Any code block delimiters (e.g., ```json).

Examples:
Example Input:
Reference Text: Michael Scott is a fictional character in the NBC sitcom The Office, portrayed by Steve Carell. Michael is the regional manager of the Scranton, Pennsylvania branch of Dunder Mifflin, a paper company, for the majority of the series. Like his counterpart in the original British version of the show, David Brent, he is characterized as a largely incompetent, unproductive, unprofessional boss, though he is depicted as kinder and occasionally shown to be effective at his job in key moments.
Query: Tell me a paragraph bio of Michael Scott.
Claim: The fictional character Michael Scott is the regional manager of a paper company.

Example Output:
{
  "highlighted_text": "Michael Scott is a fictional character in the NBC sitcom The Office, portrayed by Steve Carell. Michael is the regional manager of the Scranton, Pennsylvania branch of Dunder Mifflin, a paper company, for the majority of the series.",
  "reasoning": "Reference explicitly states Michael is regional manager at a paper company.",
  "answer": true
}

Example Input:
Reference Text: Michael Scott is a fictional character in the NBC sitcom The Office, portrayed by Steve Carell. Michael is the regional manager of the Scranton, Pennsylvania branch of Dunder Mifflin, a paper company, for the majority of the series. Like his counterpart in the original British version of the show, David Brent, he is characterized as a largely incompetent, unproductive, unprofessional boss, though he is depicted as kinder and occasionally shown to be effective at his job in key moments.
Query: Tell me a paragraph bio of Michael Scott.
Claim: The portrayal of Michael Scott in the NBC sitcom The Office is similar to that of David Brent, in the British version.

Example Output:
{
  "highlighted_text": "Like his counterpart in the original British version of the show, David Brent, he is characterized as a largely incompetent, unproductive, unprofessional boss, though he is depicted as kinder and occasionally shown to be effective at his job in key moments.",
  "reasoning": "Reference compares Michael Scott's characterization to David Brent's, indicating similarity.",
  "answer": true
}

Example Input:
Reference Text: Michael Scott is a fictional character in the NBC sitcom The Office, portrayed by Steve Carell. Michael is the regional manager of the Scranton, Pennsylvania branch of Dunder Mifflin, a paper company, for the majority of the series. Like his counterpart in the original British version of the show, David Brent, he is characterized as a largely incompetent, unproductive, unprofessional boss, though he is depicted as kinder and occasionally shown to be effective at his job in key moments.
Query: Tell me a paragraph bio of Michael Scott.
Claim: Michael Scott is the founder of The Michael Scott Paper Company.

Example Output:
{
  "highlighted_text": "Michael is the regional manager of the Scranton, Pennsylvania branch of Dunder Mifflin, a paper company, for the majority of the series.",
  "reasoning": "Reference mentions only Dunder Mifflin; founding another company is not supported.",
  "answer": false
}

Example Input:
Reference Text: Michael Scott is a fictional character in the NBC sitcom The Office, portrayed by Steve Carell. Michael is the regional manager of the Scranton, Pennsylvania branch of Dunder Mifflin, a paper company, for the majority of the series. Like his counterpart in the original British version of the show, David Brent, he is characterized as a largely incompetent, unproductive, unprofessional boss, though he is depicted as kinder and occasionally shown to be effective at his job in key moments.
Query: Tell me a paragraph bio of Michael Scott.
Claim: Michael Scott is the CEO of Dunder Mifflin.

Example Output:
{
  "highlighted_text": "Michael is the regional manager of the Scranton, Pennsylvania branch of Dunder Mifflin, a paper company, for the majority of the series.",
  "reasoning": "Reference states regional manager, not CEO.",
  "answer": false
}

Input:
Reference Text: {reference}
Query: {query}
Claim: {claim}

(Reiteration of the instruction)
Return a single JSON5 object with "highlighted_text", "reasoning", and "answer" (true if supported; false if contradicted or unsupported). Assign true or false—never null/None.

Output:
\end{Verbatim}
\end{promptbox}

\subsubsection{Labeler (with ground-truth answer)} \label{sec:prompt_labeler_gt}

\begin{promptbox}
\begin{Verbatim}[breaklines=true, breakanywhere=true]
You are an AI assistant tasked with assigning a label to a claim based on its factuality.

Instructions:
1. You are given:
- A reference text for the query.
- A query.
- A provided solution (final answer)
- A claim made in response to the query.
2. Rate the factuality of the claim as a boolean:
- true → The claim is well-supported by the reference text or match the given provided solution (final answer).
- false → The claim contradicts the reference text or is not well-supported by it and contradicts the provided solution (final answer).
3. Provide:
- The part(s) of the reference text or the provided solution that directly support your decision.
- A reasoning statement describing your rationale.
4. You must assign either true or false. Never return null or None.

Output Requirements:
- Output ONLY a single VALID JSON5 object with EXACTLY this schema:
{
  "highlighted_text": "Part(s) of the reference text or the provided solution that directly support the decision.",
  "reasoning": "A reasoning statement describing your rationale.",
  "answer": true
}
- "answer" must be a boolean (true/false).

JSON5 Rules:
- Use DOUBLE QUOTES (") for all keys and all string values.
- Escape double quotes inside string values as \".
- Escape backslashes as \\.
- No trailing commas in objects or arrays.
- Use the exact top-level container specified and close it properly.
- Do not include comments, code fences, or any text outside the JSON5 output.
- Follow the schema exactly; do not add or omit keys.

Do NOT include:
- Any text, explanations, comments, or formatting outside of the JSON5.
- Any code block delimiters (e.g., ```json).

Examples:
Example Input:
Reference Text: "Paris is the capital of France."
Answer: "Paris"
Query: "What is the capital of France?"
Claim: "Paris is the capital of France."

Example Output:
{
  "reasoning": "Claim matches reference and provided solution.",
  "answer": true
}

Example Input:
Reference Text: "Michael is the regional manager of Dunder Mifflin."
Answer: "Regional manager"
Query: "What is Michael's role at Dunder Mifflin?"
Claim: "Michael is the CEO of Dunder Mifflin."

Example Output:
{
  "reasoning": "Claim contradicts the reference and solution.",
  "answer": false
}

Example Input:
Reference Text: "Compute 12 × 3 + 4."
Answer: "40"
Query: "Evaluate 12 × 3 + 4."
Claim: "The value is 40."

Example Output:
{
  "reasoning": "Computation equals 40; claim matches the solution.",
  "answer": true
}

Example Input:
Reference Text: "Compute 12 × 3 + 4."
Answer: "40"
Query: "Evaluate 12 × 3 + 4."
Claim: "The value is 42."

Example Output:
{
  "reasoning": "The solution gives 40, not 42.",
  "answer": false
}

Example Input:
Reference Text: "The sum of the interior angles of a triangle is 180 degrees."
Answer: "180 degrees"
Query: "What is the sum of the interior angles of a triangle?"
Claim: "The sum is 180 degrees."

Example Output:
{
  "reasoning": "Claim agrees with the stated geometric fact and the solution.",
  "answer": true
}

Input:
Reference Text: {reference}
Answer: {answer}
Query: {query}
Claim: {claim}

(Reiteration of the instruction)
Return a single JSON5 object with "highlighted_text", "reasoning", and "answer" (true if supported; false if contradicted or unsupported). Assign true or false—never null/None.

Output:
\end{Verbatim}
\end{promptbox}

\subsubsection{Attacker} \label{generating-hallucination-claims}

\begin{promptbox}
\begin{Verbatim}[breaklines=true, breakanywhere=true]
You are a helpful assistant that modifies a correct claim into a false but plausible claim ("hallucination") given a reference text and a query.

Instructions:
1. Read the reference text, the query, the original claim, and the two lists of prior modified claims (that do not work / that do work).
2. Generate exactly one false claim that:
- Directly answers the query.
- Contradicts or is not supported by the reference text.
- Appears plausible given the reference and query (e.g., near-miss number/date, closely related entity, role swap).
- Does not match or paraphrase any claim in either provided list.
3. Provide a reasoning statement describing why the false claim might look plausible.

Output Requirements:
- Output ONLY a single VALID JSON5 object with EXACTLY this schema:
{
  "reasoning": "A reasoning statement describing why the false claim might look plausible.",
  "subclaim": "One-sentence false but plausible answer to the query."
}
- The "subclaim" must be a standalone sentence that answers the query and is false with respect to the reference.

JSON5 Rules:
- Use DOUBLE QUOTES (") for all keys and all string values.
- Escape double quotes inside string values as \".
- Escape backslashes as \\.
- No trailing commas in objects or arrays.
- Use the exact top-level container specified and close it properly.
- Do not include comments, code fences, or any text outside the JSON5 output.
- Follow the schema exactly; do not add or omit keys.

Do NOT include:
- Any text, explanations, comments, or formatting outside of the JSON5.
- Any code block delimiters (e.g., ```json).

Examples:
Example Input:
Modified claims that do not work: []
Modified claims that do work: []
Reference Text: "Gustave Eiffel oversaw the construction of the Eiffel Tower, which opened in 1889. The tower's design is credited to Gustave Eiffel."
Query: "Who designed the Eiffel Tower?"
Claim: "Gustave Eiffel designed the Eiffel Tower."

Example Output:
{
  "reasoning": "Reference credits Eiffel; another associated architect name may look credible.",
  "subclaim": "Stephen Sauvestre designed the Eiffel Tower."
}

Example Input:
Modified claims that do not work: []
Modified claims that do work: []
Reference Text: "The capital of Australia is Canberra."
Query: "What is the capital of Australia?"
Claim: "Canberra."

Example Output:
{
  "reasoning": "Larger city misconception is common.",
  "subclaim": "Sydney is the capital of Australia."
}

Input:
Modified claims that do not work: {claims_not_work}
Modified claims that do work: {claims_work}
Reference Text: {reference}
Query: {query}
Claim: {claim}

(Reiteration of the instruction)
Generate exactly one false but plausible claim answering the query, unsupported by or contradicting the reference, not overlapping with provided lists; then return a JSON5 object with "reasoning" and "subclaim".

Output:
\end{Verbatim}
\end{promptbox}

\subsubsection{Confusee} \label{hallucination-verification}

\begin{promptbox}
\begin{Verbatim}[breaklines=true, breakanywhere=true]
You are a helpful assistant that evaluates whether a claim sounds like something you (an LLM) would plausibly say—and potentially hallucinate—based on the provided reference text and query.

Instructions:
1. Read the reference text, the query, and the claim.
2. Decide whether the claim is a plausible LLM-style statement that could arise from the reference and query even if it is not supported or is false.
3. Return true if the claim is plausibly hallucinated; otherwise return false.
4. Provide a reasoning statement describing your rationale.

Output Requirements:
- Output ONLY a single VALID JSON5 object with EXACTLY this schema:
{
  "reasoning": "A reasoning statement describing your rationale.",
  "answer": (true|false)
}
- "answer" must be a boolean (true/false).

JSON5 Rules:
- Use DOUBLE QUOTES (") for all keys and all string values.
- Escape double quotes inside string values as \".
- Escape backslashes as \\.
- No trailing commas in objects or arrays.
- Use the exact top-level container specified and close it properly.
- Do not include comments, code fences, or any text outside the JSON5 output.
- Follow the schema exactly; do not add or omit keys.

Do NOT include:
- Any text, explanations, comments, or formatting outside of the JSON5.
- Any code block delimiters (e.g., ```json).

Examples:
Example Input:
Reference Text: "The capital of Australia is Canberra."
Query: "What is the capital of Australia?"
Claim: "Sydney is the capital of Australia."

Example Output:
{
  "reasoning": "Common misconception favors the largest city over the capital.",
  "answer": true
}

Example Input:
Reference Text: "Paris is the capital of France."
Query: "What is the capital of France?"
Claim: "Paris is the capital of France."

Example Output:
{
  "reasoning": "Claim is directly supported by the reference; not a hallucination.",
  "answer": false
}

Input:
Reference Text: {reference}
Query: {query}
Claim: {claim}

(Reiteration of the instruction)
Return a JSON5 object with "reasoning" and "answer" (true if plausibly hallucinated; else false).

Output:
\end{Verbatim}
\end{promptbox}

\subsubsection{Model Confidence Score} \label{modelconfscore_prompt}

\begin{promptbox}
\begin{Verbatim}[breaklines=true, breakanywhere=true]
You are an AI assistant tasked with assigning a confidence score to a claim based on its factuality.

Instructions:
1. You are given:
- A reference text for the query.
- A query.
- A claim made in response to the query.
2. Rate the factuality of the claim with a numeric score in [0.0, 1.0]:
- 0.0 -> The claim is false, contradicts the reference, or is not well-supported.
- 1.0 -> The claim is factual and well-supported by the reference, or can be logically deduced from the reference and query.
3. Provide:
- A score.
- The parts of the reference text that directly support your scoring decision.
- A reasoning statement describing your rationale.
4. You must assign a numeric score. Never return null, None, or a non-numeric value.

Important: A claim should be considered true (score 1.0) if it is either:
- Directly stated in the reference text, OR
- Can be logically deduced or calculated from the information provided in the reference text and query.
For mathematical claims, perform the necessary calculations based on the given data.
Output Requirements:
- Output ONLY a single VALID JSON5 object with EXACTLY these keys:
{
  "highlighted_text": "Part(s) of the reference text that support the decision.",
  "reasoning": "A reasoning statement describing your rationale.",
  "score": 0.0-1.0
}

JSON5 Rules:
- Use DOUBLE QUOTES (") for all keys and all string values.
- Escape double quotes inside string values as \".
- Escape backslashes as \\\.
- No trailing commas in objects or arrays.
- Follow the schema exactly.

Do NOT include:
- Any text, explanations, comments, or formatting outside of the JSON5.

Input:
Reference Text: {reference}
Query: {query}
Claim: {claim}

Reiteration of Instructions:
Return a single JSON5 object with "highlighted_text", "reasoning", "score". Assign a numeric score—never null/None.

Output:
\end{Verbatim}
\end{promptbox}

\subsubsection{Merger (FActScore)} \label{sec:prompt_merger_factscore}

\begin{promptbox}
\begin{Verbatim}[breaklines=true, breakanywhere=true]
You will get an instruction and a set of facts that are true. Construct an answer using ONLY the facts provided, and try to use all facts as long as its possible. If the input facts are empty, output the empty string. Do not repeat the instruction.
Input:
The facts: {claims}
The instruction: {query}
Remember, If the input facts are empty, output the empty string. Do not repeat the instruction.
Output:
\end{Verbatim}
\end{promptbox}

\subsubsection{Merger (Natural Questions)} \label{sec:prompt_merger_nq}

\begin{promptbox}
\begin{Verbatim}[breaklines=true, breakanywhere=true]
You will get a natural question and parts of an answer, which you are to merge into coherent prose. Make sure to include all the parts in the answer. There may be parts that are seemingly unrelated to the others, but DO NOT add additional information or reasoning to merge them. If the input parts are empty, output the empty string. Do not repeat the question.
Input:
The parts: {claims}
The question: {query}
Remember, DO NOT add any additional information or commentary, just combine the parts. If the input parts are empty, output the empty string. Do not repeat the question.
Output:
\end{Verbatim}
\end{promptbox}

\subsubsection{Merger (MATH)} \label{sec:prompt_merger_math}

\begin{promptbox}
\begin{Verbatim}[breaklines=true, breakanywhere=true]
You will get a math problem and a set of steps that are true. Construct an answer using ONLY the steps provided. Make sure to include all the steps in the answer, and do not add any additional steps or reasoning. These steps may not fully solve the problem, but merging them could assist someone in solving the problem. If the input steps are empty, output the empty string. Do not repeat the math problem.
Input:
The steps: {claims}
The math problem: {query}
Remember, do not do any additional reasoning, just combine the given steps. If the input steps are empty, output the empty string. Do not repeat the math problem.
Output:
\end{Verbatim}
\end{promptbox}

\subsubsection{Correctness} \label{sec:prompt_correctness}

\begin{promptbox}
\begin{Verbatim}[breaklines=true, breakanywhere=true]
I need your help in evaluating an answer provided by an LLM against ground truth answers. Your task is to determine if the LLM's response matches the ground truth answers. Please analyze the provided data and make a decision.

Instructions:
1. Carefully compare the "Predicted Answer" with the "Ground Truth Answers".
2. Consider the substance of the answers – look for equivalent information or correct answers. Do not focus on exact wording unless the exact wording is crucial to the meaning.
3. Your final decision should be based on whether the meaning and the vital facts of the "Ground Truth Answers" are present in the "Predicted Answer."
4. Categorize the answer as one of the following:
- "perfect": The answer is completely correct and matches the ground truth.
- "acceptable": The answer is partially correct or contains the main idea of the ground truth.
- "incorrect": The answer is wrong or contradicts the ground truth.
- "missing": The answer is "I don't know", "invalid question", or similar responses indicating lack of knowledge.

Output Requirements:
- Output ONLY a single VALID JSON5 object with EXACTLY this schema:
{
  "reasoning": "A reasoning statement describing your rationale.",
  "answer": "One of perfect, acceptable, incorrect, or missing"
}

JSON5 Rules:
- Use DOUBLE QUOTES (") for all keys and all string values.
- Escape double quotes inside string values as \".
- Escape backslashes as \\.
- No trailing commas in objects or arrays.
- Use the exact top-level container specified and close it properly.
- Do not include comments, code fences, or any text outside the JSON5 output.
- Follow the schema exactly; do not add or omit keys.

Do NOT include:
- Any text, explanations, comments, or formatting outside of the JSON5.
- Any code block delimiters (e.g., ```json).

Input:
Query: {query}
Predicted Answer: {merged_string}
Ground Truth Answer: {answer}

Reiteration of Instructions (before output):
1. Carefully compare the "Predicted Answer" with the "Ground Truth Answers".
2. Consider the substance of the answers – look for equivalent information or correct answers. Do not focus on exact wording unless the exact wording is crucial to the meaning.
3. Your final decision should be based on whether the meaning and the vital facts of the "Ground Truth Answers" are present in the "Predicted Answer."
4. Categorize the answer as one of the following:
- "perfect": The answer is completely correct and matches the ground truth.
- "acceptable": The answer is partially correct or contains the main idea of the ground truth.
- "incorrect": The answer is wrong or contradicts the ground truth.
- "missing": The answer is "I don't know", "invalid question", or similar responses indicating lack of knowledge.

Output:
\end{Verbatim}
\end{promptbox}

\subsubsection{Sufficient Correctness} \label{sec:prompt_sc}

\begin{promptbox}
\begin{Verbatim}[breaklines=true, breakanywhere=true]
You are an expert LLM evaluator that excels at evaluating a RESPONSE with respect to a QUERY given a REFERENCE.

Consider the following criteria:
Sufficient Correctness:
1 IF the RESPONSE contains a sufficient amount of CORRECT information (verified against the REFERENCE) to infer the answer to the QUERY.
0 IF the RESPONSE does not contain a sufficient amount of CORRECT information to infer the answer to the QUERY.
Important:
Judge only the correctness of the RESPONSE content according to the REFERENCE.
Do not infer correctness from external knowledge.

Output ONLY a single VALID JSON5 object with EXACTLY these keys:
{
  "explanation": An explanation describing your rationale, with with you will make your decision on sufficient_correctness,
  "sufficient_correctness": 1 IF the RESPONSE contains a sufficient amount of correct information (verified against the REFERENCE) to infer the answer to the QUERY, 0 IF the RESPONSE does not contain a sufficient amount of correct information to infer the answer to the QUERY.
}
- "sufficient_correctness" must be an integer (0/1).

JSON5 Rules:
- Use DOUBLE QUOTES (") for all keys and all string values.
- Escape double quotes inside string values as \".
- Escape backslashes as \\.
- No trailing commas in objects or arrays.
- Use the exact top-level container specified and close it properly.
- Do not include comments, code fences, or any text outside the JSON5 output.
- Follow the schema exactly; do not add or omit keys.

Do NOT include:
- Any text, explanations, comments, or formatting outside of the JSON5.
- Any code block delimiters (e.g., ```json).

Input:
### QUERY
{query}
### REFERENCE
{reference}
### RESPONSE
{response}
Output:
\end{Verbatim}
\end{promptbox}

\subsubsection{MATH reference} \label{math_reference}

\begin{promptbox}
\begin{Verbatim}[breaklines=true, breakanywhere=true]
You are a helpful assistant that extracts the prerequisite mathematics knowledge needed to answer a given question—without solving it.

Instructions:
1. Read the question.
2. Identify only the minimal prerequisite items across concepts, definitions, theorems/properties, formulas, techniques, notation, assumptions/conditions, and common pitfalls required to answer the question.
3. When you state a theorem, a definition, or a formula, write out its full, standard statement (not just the name). For theorems, include hypotheses and conclusions; for definitions, give the precise meaning; for formulas, write the exact equation(s) in standard notation.
4. Do NOT provide the answer or partial solution steps.
5. Do NOT use examples that arise directly from the given question; keep statements general and problem-agnostic.

Output Requirements:
- Produce PLAIN TEXT ONLY.
- Write free-form prose (no headings, lists, or numbering).
- Mention required topics, definitions, fully written theorems/properties, formulas, techniques, notation conventions, assumptions/conditions, and common pitfalls in sentences.
- The output may be long; include complete statements where needed.

Do NOT include:
- Any answer, hints, or step-by-step solution.
- Any examples derived from the given question.

Reiteration of the instructions:
1. Read the question.
2. Identify only the minimal prerequisite items across concepts, definitions, theorems/properties, formulas, techniques, notation, assumptions/conditions, and common pitfalls required to answer the question.
3. When you state a theorem, a definition, or a formula, write out its full, standard statement (not just the name). For theorems, include hypotheses and conclusions; for definitions, give the precise meaning; for formulas, write the exact equation(s) in standard notation.
4. Do NOT provide the answer or partial solution steps.
5. Do NOT use examples that arise directly from the given question; keep statements general and problem-agnostic.

Input:
{query}

Output:

\end{Verbatim}
\end{promptbox}

\end{document}